\documentclass[10pt,twocolumn,letterpaper]{article}

\usepackage[pagenumbers]{iccv} %

\usepackage{amsthm}
\usepackage{mathtools}
\usepackage[percent]{overpic}
\usepackage{pifont}
\usepackage{subcaption}
\usepackage{lipsum}
\usepackage{gensymb}
\usepackage{bm}
\usepackage{upgreek}
\usepackage{cuted}
\usepackage{wrapfig}

\usepackage{booktabs}
\usepackage{tabularray}
\UseTblrLibrary{booktabs}
\usepackage[export]{adjustbox}
\usepackage{rotating}
\usepackage{makecell}

\usepackage{xspace}

\definecolor{iccvblue}{rgb}{0.21,0.49,0.74}
\usepackage[pagebackref,breaklinks,colorlinks,allcolors=iccvblue]{hyperref}

\def\paperID{10103} %
\def\confName{ICCV}
\def\confYear{2025}

\title{AnyCalib: \\ On-Manifold Learning for Model-Agnostic Single-View Camera Calibration}

\author{Javier Tirado-Garín \qquad Javier Civera\\
I3A, University of Zaragoza\\
{\tt\small \{j.tiradog, jcivera\}@unizar.es}
}

\newcommand{\PAR}[1]{\vskip4pt \noindent{\bf #1~}}

\newcommand{\bgcolor}[2]{\setlength{\fboxsep}{0pt}\colorbox{#1}{\strut #2}}

\newcommand{\bc}{\mathbf{c}}

\newcommand{\bp}{\mathbf{p}}

\newcommand{\bv}{\mathbf{v}}

\newcommand{\bx}{\mathbf{x}}

\newcommand{\bz}{\mathbf{z}}

\newcommand{\btheta}{\boldsymbol{\theta}}

\newcommand{\Stwo}{\mathcal{S}^2}

\newcommand{\bff}{\mathbf{f}}

\newcommand{\ceqq}{\coloneqq}

\newcommand{\tangent}{T_{\mathbf{z}_1}\mathcal{S}^2}

\newcommand{\twodots}{\mathinner{\ldotp\ldotp}}

\DeclareMathOperator{\atantwo}{atan2}
\DeclareMathOperator{\logz}{Log_{\bz_{1}}}
\DeclareMathOperator{\expz}{Exp_{\bz_{1}}}

\definecolor{pastelred}{RGB}{235,119,119}
\definecolor{pastelorange}{RGB}{255,202,126}

\makeatletter
\newcommand{\setword}[2]{%
  \phantomsection
  #1\def\@currentlabel{\unexpanded{#1}}\label{#2}%
}
\makeatother

\providetoggle{foobool}

\newcommand{\datp}{OP$_{\mathrm{p}}$\xspace}
\newcommand{\datr}{OP$_{\mathrm{r}}$\xspace}
\newcommand{\datd}{OP$_{\mathrm{d}}$\xspace}
\newcommand{\datg}{OP$_{\mathrm{g}}$\xspace}

\begin{document}
\crefname{appendix}{Supp.}{Supps.}
\Crefname{appendix}{Supplementary}{Supplementaries}
\maketitle

\begin{strip}
    \centering
    \includegraphics[width=0.99\textwidth]{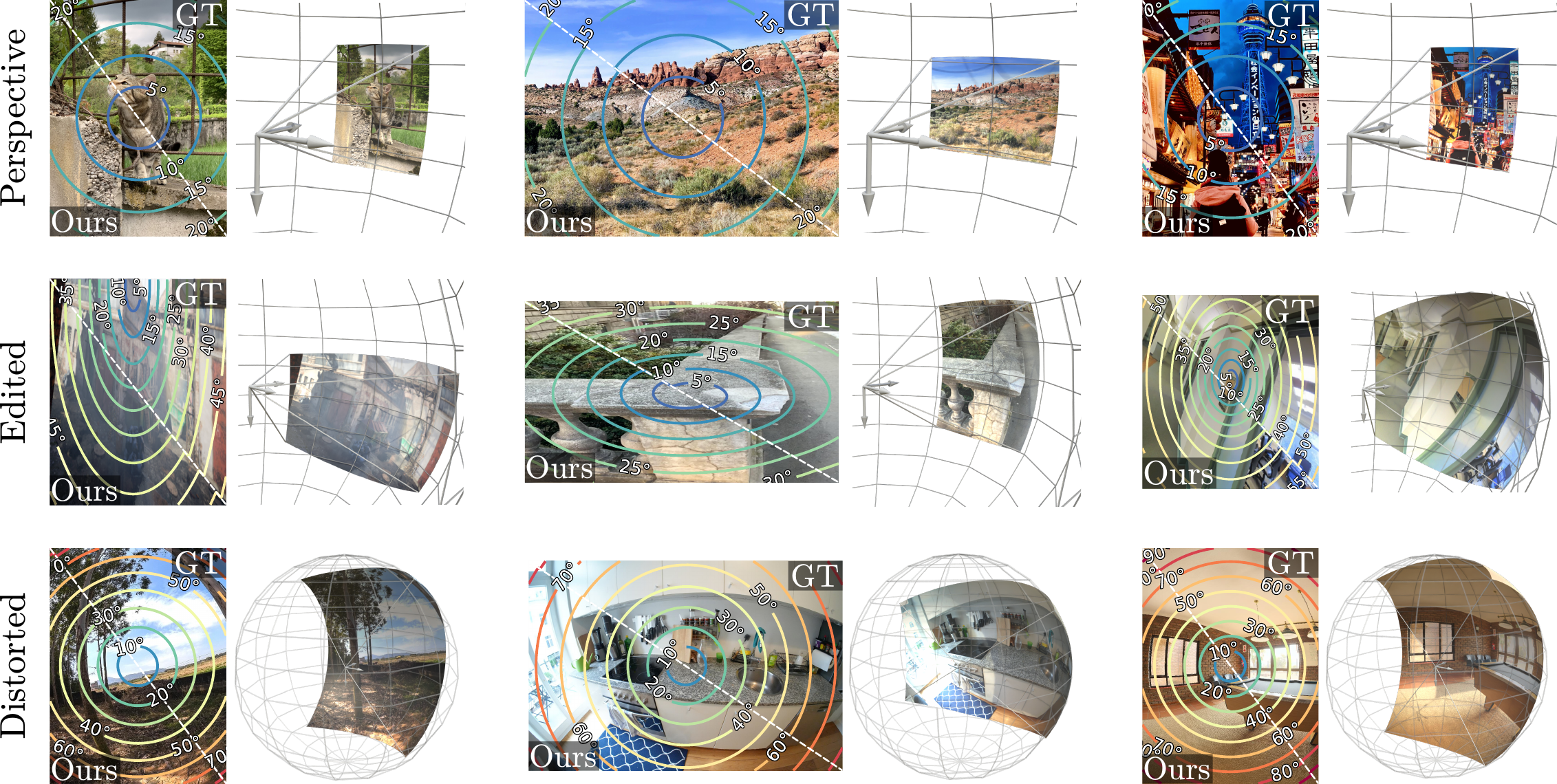}
    \captionof{figure}{\textbf{We introduce AnyCalib}, the first model-agnostic single-view camera calibration method. AnyCalib works with perspective, edited and distorted images and has flexibility in selecting the camera model at runtime.
    The left subfigures compare ground-truth and estimated \emph{polar angles} (angle between the optical axis and the ray direction of each pixel) on \emph{in-the-wild} images. Despite the wide range of field-of-view and the absence of perspective cues in natural and close-up photos, AnyCalib accurately calibrates each camera. It does so by framing the calibration process as the regression of the rays corresponding to each pixel, which we represent on the right subfigures.}
    \label{fig:teaser}
    \vspace{-5pt}
\end{strip}

\begin{abstract}
We present AnyCalib, a method for calibrating the intrinsic parameters of a camera from a single in-the-wild image, that is agnostic to the camera model. Current methods are predominantly tailored to specific camera models and/or require extrinsic cues, such as the direction of gravity, to be visible in the image. In contrast, we argue that the perspective and distortion cues inherent in images are sufficient for model-agnostic camera calibration. 
To demonstrate this, we frame the calibration process as the regression of the rays corresponding to each pixel. 
\newline
\newline
\newline
We show, for the first time, that this intermediate representation allows for a closed-form recovery of the intrinsics for a wide range of camera models, including but not limited to: pinhole, Brown-Conrady and Kannala-Brandt. Our approach also applies to edited---cropped and stretched---images. Experimentally, we demonstrate that AnyCalib consistently outperforms alternative methods, including 3D foundation models, despite being trained on orders of magnitude less data. 
Code is available at \url{https://github.com/javrtg/AnyCalib}.
\newpage
\end{abstract}

\vspace{-13mm}
\section{Introduction}
\label{sec:intro}

Camera calibration is the task of estimating the mapping from image points to their corresponding \emph{ray} directions~\cite{Hartley2004mvg}. 
Although alternatives exist \cite{Schops2020why, Grossberg2001generic}, this mapping is typically modeled using a set of \emph{intrinsic parameters} (or \emph{intrinsics} for short), which depend on the chosen camera model \cite{sturm2011camera, usenko2018double, lochman2021babel}.
Accurate intrinsics are crucial for many computer vision tasks that aim to recover the 3D geometry of an observed scene, as \eg, in SLAM \cite{mur2015orb, mur2017orb2, campos2021orb3}, Structure-from-Motion (SfM) \cite{agarwal2011building, pan2024global, schonberger2016colmap} and novel-view synthesis \cite{mildenhall2020nerf, Kerbl2023splatting}. 

The literature on calibration from multiple images is vast and mature. The most accurate methods assume \emph{controlled} scenes, where a calibration target with known geometry is observed \cite{zhang2000calibration, bouguet2004calibration, Furgale2013kalibr, scaramuzza2006calibration, urban2015calibration, lochman2021babel}. Intrinsics can also be estimated in non-controlled scenes as part of SfM \cite{agarwal2011building, schonberger2016colmap, pan2024global} and visual SLAM \cite{civera2009calibration, Hagemann2023droidcalib, murai2024mast3r}, provided that sufficient geometric constraints are present across images/views.

However, some applications require known intrinsics for \emph{in-the-wild} \emph{single-view} tasks, like depth \cite{facil2019camconvs, Berenguel-Baeta2023convkb, Yin2023metric3d, hu2024metric3dv2, Lichy2024fovadepth} and normal \cite{bae2024dsine} map estimations.
Moreover, in non-controlled scenes, multi-view geometric methods fail when the intrinsics are insufficiently constrained, due to, \eg, limited visual overlap across views.
In such cases, single-view calibration methods are an attractive alternative. Rather than multi-view constraints, they use alternative geometric \cite{Lochman2021sva,Pautrat2023uvp, Antunes2017unsupervised, wildenauer2013closed} and learned \cite{Jin2023perspective, veicht2024geocalib, HoldGeoffroy2023perceptual, lopez2019deepcalib} cues present in a single input image. 

They, however, have limitations. Purely geometric approaches excel in structured scenes, where their assumptions, \eg the presence of parallel lines, are satisfied, but catastrophically fail otherwise \cite{veicht2024geocalib}. On the other hand, methods trained to directly predict intrinsics \cite{Jin2023perspective, HoldGeoffroy2018perceptual, lopez2019deepcalib} only learn a subset of image projections, which impacts their accuracy in out-of-domain ones. 
Additionally, they often condition the calibration with extrinsic cues, such as the direction of gravity and the image location of the horizon \cite{Jin2023perspective, veicht2024geocalib, lopez2019deepcalib}. Thus, their accuracy decreases when these cues are not visible. Finally, current single-view calibration methods, geometric and learned, are tailored to specific camera models, either by design \cite{lochman2021babel, Pautrat2023uvp, wildenauer2013closed, HoldGeoffroy2023perceptual, lopez2019deepcalib} and/or during training \cite{Jin2023perspective, veicht2024geocalib}.

To address these limitations, we propose AnyCalib, a novel single-view camera calibration method. Our main contributions are:
\begin{itemize}
    \item[$\circ$] AnyCalib is the first \emph{model-agnostic} single-view camera calibration model. We frame the calibration process as the regression of the rays corresponding to each pixel. We show, that this representation is model-agnostic since it allows for a closed-form recovery of the intrinsics for a wide range of camera models, without conditioning its training nor design on specific ones. 
    Thus, our method also applies to \emph{edited} (stretched and cropped) images. 
    Some qualitative results are shown in \cref{fig:teaser}.
    \item[$\circ$] We introduce \emph{Field of View (FoV) fields}, a novel intermediate representation that is bijective to the rays of each pixel. In contrast to ray (or incidence \cite{zhu2023wildcam}) fields, FoV fields are a minimal Euclidean representation that is directly related to the image content. We demonstrate that this representation leads to improved accuracy.
    \item[$\circ$] Since our approach is not coupled with extrinsic cues, we extend the OpenPano dataset \cite{veicht2024geocalib} with panoramas that do not need to be aligned with the gravity direction. This also allows us to show that AnyCalib is scalable, as this extended dataset leads to improved accuracy.
    \item[$\circ$] We propose an alternative ``light'' DPT decoder \cite{Ranftl2021dpt} that does not use expensive transposed convolutions to upsample the predictions. This improves the model efficiency.
    \item[$\circ$] Finally, AnyCalib sets a new state-of-the-art across perspective, edited and distorted images, while having flexibility in selecting camera models at runtime.
\end{itemize}

\section{Related Work}
\label{sec:rw}

\PAR{Purely geometric approaches} calibrate intrinsics using constraints derived from low-level visual cues. Vanishing points and lines \cite{Hartley2004mvg} are the most common \cite{Lochman2021sva, Pautrat2023uvp, kovsecka2002compass, wildenauer2013closed, Pritts2018radially, Antunes2017unsupervised, chen2008full, deutscher2002automatic} and are typically fitted to line detections \cite{grompone2010lsd, Pautrat2023deeplsd} that are parallel and/or coplanar,
using \eg RANSAC-based methods \cite{Kluger2020consac, Pautrat2023uvp, Wildenauer2012robust, Lochman2021sva}. 
These approaches, thus, have strict requirements: parallel lines must be detected and vanishing points may need to form a Manhattan frame \cite{kovsecka2002compass, lochman2021babel, deutscher2002automatic, Pautrat2023uvp}. Geometric approaches are thus accurate in structured scenes but fail otherwise, even in urban imagery \cite{veicht2024geocalib, Li2018megadepth}. 
Moreover, they are restricted to the pinhole \cite{Pautrat2023uvp, kovsecka2002compass, chen2008full, deutscher2002automatic} or division \cite{Lochman2021sva, Antunes2017unsupervised, wildenauer2013closed} camera models.
In contrast, AnyCalib is model-agnostic and more robust since it does not rely only on the previous low-level cues. Our proposed intermediate representation and training dataset are appropriate for scenes lacking them, as shown in
\cref{fig:teaser}.

\PAR{End-to-end learned approaches} train deep neural networks to directly predict the intrinsics of a certain camera model, such as pinhole \cite{Jin2023perspective, Song2024mscc, Lee2021ctrlc, Janampa2024sofi},  radial \cite{lopez2019deepcalib}, UCM \cite{HoldGeoffroy2018perceptual, HoldGeoffroy2023perceptual, bogdan2018deepcalib} or alternative proposed ones \cite{wakai2022rethinking, Wakai2024heatmap}.
They are generally more robust than geometric approaches \cite{veicht2024geocalib}, but at the expense of \emph{accuracy}, as they do not impose geometric constraints, and \emph{flexibility}, since adapting to other cameras would require retraining and architectural changes.
Additionally, they often rely on extrinsic cues, such as the direction of gravity \cite{Jin2023perspective}, the image location of the horizon \cite{lopez2019deepcalib} or lines converging at the horizon and zenith \cite{Lee2021ctrlc}. Thus, they lose accuracy if these cues are not visible. 
In contrast, AnyCalib is not tied to a specific camera model or extrinsic cues and its accuracy is comparable or better than geometric approaches thanks to imposing geometric constraints.

\begin{figure*}[t]
    \centering
    \includegraphics[width=\linewidth]{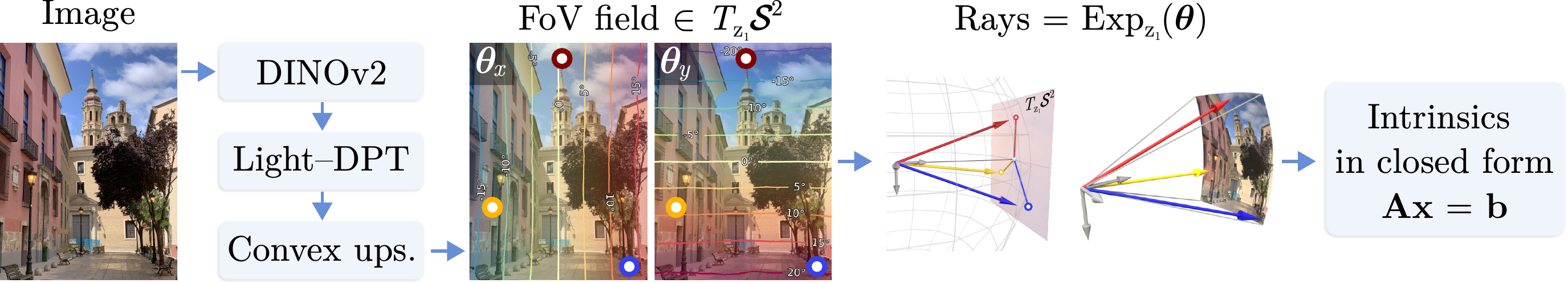}
    \caption{\textbf{Method.} AnyCalib predicts dense FoV fields (\cref{sec:fov}) using a transformer backbone and a light CNN decoder (\cref{sec:imp}). FoV fields are supervised on the unit sphere, in the tangent plane at the optical axis of the camera. This representation is bijective to rays, which along their corresponding image coordinates, allow a closed-form model-agnostic calibration of a wide range of camera models (\cref{sec:calib}).}
    \label{fig:method}
\end{figure*}

\PAR{Hybrid approaches} train an intermediate representation that is used to fit the intrinsics by imposing geometric constraints \cite{zhu2023wildcam, he2025diffcalib, veicht2024geocalib, dalcin2024revisiting}, in a similar spirit to previous works on distortion \cite{Li2019blind} and rotation \cite{Xian2019upright} estimation. WildCam \cite{zhu2023wildcam} and DiffCalib \cite{he2025diffcalib} directly estimate the ray directions for each pixel, restricting their implementation to perspective (pinhole) and edited images. 
\citet{dalcin2024revisiting} also propose regressing rays, using a polar input representation that restricts their method to non-edited images. It requires heuristics \cite{storn1997differential} to fit the intrinsics and its training uses a highly non-linear reprojection error \cite{usenko2018double} as loss function, which can difficult its convergence. GeoCalib \cite{veicht2024geocalib} iteratively optimizes the intrinsics from learned perspective fields \cite{Jin2023perspective}, that are weighted by uncertainties trained by supervising intrinsics. 
This generally leads to more accurate local minimums but increases the risk of overfitting. 
Additionally, the accuracy of perspective fields decreases when the horizon and gravity direction are not perceivable \cite{veicht2024geocalib}.

In contrast to all previous methods, AnyCalib works with \emph{perspective}, \emph{edited} and \emph{distorted} images and is not tied to extrinsic cues. Different to methods that also predict rays, AnyCalib uses a minimal Euclidean intermediate representation that leads to improved accuracy and that also applies to edited images. Additionally, we show, for the first time, that rays allow for a \emph{closed-form solution} of the intrinsics for a wide range of camera models.

\begin{figure*}[t]
    \centering
    \includegraphics[width=1.0\linewidth]{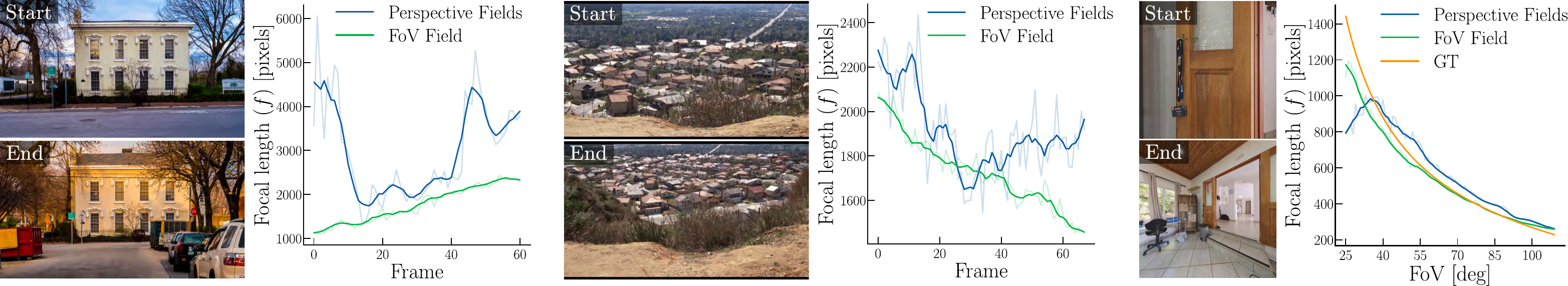}
    \caption{\textbf{Perspective vs FoV fields.} 
    We show focal length ($f$) estimates by GeoCalib \cite{veicht2024geocalib} and AnyCalib.
    Sequences depict dolly-zooms, where $f$ continuously increases (left) or decreases (middle) while the camera moves to keep a focused object visually unchanged, and a zoom-out (right) where $f$ decreases. 
    AnyCalib correctly predicts these tendencies, while GeoCalib struggles since the horizon is occluded (left), not visible (middle) or there is insufficient image information (right).
    FoV fields are more robust as they rely on image content rather than extrinsic cues. 
    None of the images were seen during training. Media sources, from left to right: \cite{dollytuto}, \cite{dollyet} and \cite{polyhaven}.}
    \label{fig:zoom_dolly}
\end{figure*}

\section{Method}

Given an $H\times W$ input image, AnyCalib densely predicts FoV fields (\cref{sec:fov}), which are bijective to the ray directions corresponding to each pixel. From these rays and image coordinates, we obtain, in closed-form, the globally-optimal intrinsics of the camera model of choice (\cref{sec:calib}).

\subsection{FoV field as intermediate representation}
\label{sec:fov}
\PAR{Current target representations.} The unit rays corresponding to each pixel, lie on the $\mathcal{S}^2$ manifold or unit sphere, which is not closed under linear combinations of its elements. Thus, common operations, such as the upsampling done in WildCam \cite{zhu2023wildcam} to densely regress rays, are not well-defined \cite{Teed2021raft3d}. DiffCalib \cite{he2025diffcalib} and \citet{dalcin2024revisiting} use better-defined representations, but they are either limited by design to moderate Field-of-Views \cite{he2025diffcalib}---pinhole unprojections, or to non-edited images \cite{dalcin2024revisiting}---perfect square pixels and a centered principal point.

\PAR{FoV cue.} On the other hand, instead of directly regressing rays, multiple works \cite{lopez2019deepcalib,patel2024camerahmr,Kocabas2021spec,Lee2021ctrlc,HoldGeoffroy2023perceptual} identify the Field-of-View (FoV) as an image cue that 
 is inherent and directly perceivable from the image content.
However, these works limit the FoV prediction to just a single value---the maximum angular extent of the image. As such, they cannot densely impose geometric constraints.
 
\PAR{FoV fields.} To address these limitations, we propose \emph{FoV fields} as intermediate representation. As depicted in \cref{fig:method}, FoV fields correspond to elements in the tangent space $\tangent$ \cite{boumal2023intromanifolds} at the optical axis $\mathbf{z}_1\ceqq[~0~,~0~,~1~]^\top\in\mathcal{S}^2$:
\begin{equation}\label{eq:tangentv0}
    \tangent = \{\bv\in\mathbb{R}^3~\mid~\mathbf{z}_1^\top\bv=0\}~.
\end{equation}
Any ray $\bp\in\mathcal{S}^2$ can be uniquely mapped to $\tangent$ via the \emph{logarithm map} $\logz~:~\mathcal{S}^2\to\tangent$ \cite{boumal2023intromanifolds}:
\begin{equation}
    \bv = \logz(\bp) = \arccos(\bz_1^\top\bp)\frac{\bp - (\bz_1^\top\bp)\bz_1}{\lVert\bp - (\bz_1^\top\bp)\bz_1\rVert}~,
\end{equation}
which implies that 
\begin{equation}
\theta\ceqq\lVert \logz(\bp) \rVert = \arccos(\bz_1^\top\bp)~,    
\end{equation}
which is precisely the \emph{polar angle} $\theta$ corresponding to the unit ray $\bp$, \ie, the angle between $\bp$ and the optical axis $\bz_1$. In other words, FoV fields encode the FoV\footnote{The term \emph{FoV} commonly refers to the \emph{maximum} angular extent of an image. In this paper, we slightly extend this definition for referring to the angular extent \emph{up to any pixel}.} (double of $\theta$) corresponding to each pixel. 

However, $\bv\in\tangent$ are \emph{non-minimal}, \emph{constrained} vectors (\cref{eq:tangentv0}). Thus, to obtain a \emph{minimal}, \emph{unconstrained} target representation $\btheta\in\mathbb{R}^2$ to learn by our network, we use  local coordinates \cite{factor_graphs2017dellaert}, \ie we express them in a 2D coordinate system located in the tangent plane. We use axes parallel to the camera's $x$- and $y$-axes. This change of basis\footnote{Note that $\theta=\lVert\btheta\rVert$ since $\bv$ has a null z-component.} simply takes the first two elements of $\bv$:
\begin{equation}
    \btheta \ceqq \mathbf{B}_{xy} \bv~,\quad
    \mathbf{B}_{xy} = \begin{bmatrix}
        1 & 0 & 0 \\
        0 & 1 & 0 
    \end{bmatrix}~.
\end{equation}
Ground-truth values of $\btheta\in\mathbb{R}^2$ are used for training. They are obtained from ground-truth rays as
\begin{equation}\label{eq:theta_log}
    \btheta = \mathbf{B}_{xy}\logz(\bp) = 
    \frac{\theta}{\sin\theta}  \mathbf{B}_{xy}\bp~,
\end{equation}
since $\mathbf{B}_{xy}\bz_1=\mathbf{0}$ and\footnote{
Since $\bz_1,~\bp\in\mathcal{S}^2$, this implies that $\lVert\bp - (\bz_1^\top\bp)\bz_1\rVert=(1 - (\bz_1^\top\bp)^2)^{1/2}=(1 - \cos^2\theta)^{1/2}=\sin\theta.$} 
$\lVert\bp - (\bz_1^\top\bp)\bz_1\rVert=\sin\theta$.

Given dense $\btheta$ predictions by our network, we map them to unit rays using the \emph{exponential map}. It is defined as \cite{boumal2023intromanifolds}:
\begin{equation}
    \bp = \expz(\bv) = \cos(\theta)\bz_1 + \frac{\sin\theta}{\theta}\bv~,
\end{equation}
which under our representation, simplifies to
\begin{equation}\label{eq:exp_simp}
    \bp = 
    \expz(\mathbf{B}_{xy}^\top\btheta) =
    \begin{bmatrix}
        \frac{\sin\theta}{\theta} \btheta \\
        \cos\theta
    \end{bmatrix}~.
\end{equation}

\PAR{Importance of image content.}
State-of-the-art representations tied to extrinsic cues \cite{Jin2023perspective, veicht2024geocalib} often struggle when these are not clearly visible. In contrast, FoV fields are more robust, as they rely more on the image content and relative position of the elements within a scene, as shown in \cref{fig:zoom_dolly}.

\begin{table*}[t]
\footnotesize

\definecolor{unkc}{RGB}{178,255,178} %

\newcommand{\unkfmt}[1]{#1} %

    \centering
    \begin{tblr}{
        colspec={lrlll},
        cell{1}{2}={c=2}{l},
        cell{2}{2}={r=5}{l},
        vline{2,4,5}={solid, gray9},
    }
\toprule
Model & Model-specific function & - & Linear constraints & Unknowns \\
\midrule
Pinhole & $\phi(R, Z)=$ & $1/Z$ & $R_a \unkfmt{f} = Z r_c$ & $f$ \\
Brown-Conrady \cite{duane1971close} & - &
$(1/Z) (1 + \sum_{n=1}^{N} k_n (R/Z)^{2n})$ & 
$r_c Z \unkfmt{g} - R_a\sum_{n=1}^{N} \unkfmt{k_n} (R/Z)^{2n} = R_a$ & 
$g\ceqq1/f~,~\{k_1\twodots k_N\}$ \\
Kannala-Brandt \cite{kannala2006generic} & - & 
$(\theta + \sum_{n=1}^N k_n\theta^{2n+1})/R$ & 
$Rr_c g - R_a\sum_{n=1}^N k_n\theta^{2n+1} = R_a \theta$ & 
$g\ceqq1/f~,~\{k_1\twodots k_N\}$ \\
UCM \cite{mei2007ucm, geyer2000ucm} & - & 
$(\xi d + Z)^{-1}$ &
$R_af - r_cd\xi = r_cZ$ &
$f,~\xi$ \\
EUCM$^\dagger$ \cite{Khomutenko2016eucm} & - & 
$(\alpha \sqrt{\beta R^2 + Z^2} + (1-\alpha)Z)^{-1}$ & 
$r^2R^2 \gamma +2rZ (rZ-R)\alpha = (R-rZ)^2$ &
$\gamma\ceqq\alpha^2\beta~,~\alpha$\\
\midrule[gray9]
Division \cite{Fitzgibbon2001division, Larsson2019revisiting} &
$\psi(r)=$ & 
$1 + \sum_{n=1}^N k_nr^{2n}$ & 
$R_a (f + \sum_{n=1}^N k'_n r_{ca}^{2n}) = Zr_c$ &
\scalebox{0.73}{$f,~\{k'_1\twodots k'_N\},~k'_n\ceqq k_n/f^{2n-1}$}\\
\bottomrule
    \end{tblr}
    \caption{\textbf{Implemented camera models.} Once the principal point $\mathbf{c}$ and pixel aspect-ratio $a$ are known via \cref{eq:ppoint_ar}, the \cref{eq:proj,eq:unproj} become linear w.r.t. the remaining intrinsics---using the reparameterizations of the rightmost column. 
    Our implementation allows a variable number of distortion coefficients, $k$, in models that allow so \cite{kannala2006generic, duane1971close, Fitzgibbon2001division, Larsson2019revisiting}.
    $^\dagger$For EUCM \cite{Khomutenko2016eucm} we operate slightly differently since linearity in \cref{eq:proj} is lost when $f$ is unknown. Instead, to estimate $f$, we use a proxy camera model \cite{kannala2006generic} that leads to practically the same focal length value \cite{usenko2018double, lochman2021babel}.
    \emph{Auxiliary definitions} for expressing the linear constraints (see \cref{sec:supp_constraints}): 
    $R_a \ceqq \sqrt{X^2 + a^2Y^2}$, $r_c \ceqq\lVert\bx-\bc\rVert$, $\theta\ceqq\atantwo(R, Z)$, 
    $d\ceqq\sqrt{R^2 + Z^2}$ and $r_{ca}^2 \ceqq (u-c_x)^2+(v-c_y)^2/a^2$. 
    }
    \label{tab:cam_models}
\end{table*}

\subsection{Model-agnostic calibration from rays}
\label{sec:calib}

\PAR{Projection models.}
Differently to current single-view calibration methods, we altogether consider: 1) projection models with radial distortion \cite{lochman2021babel, usenko2018double}, 2) non-square image pixels, \ie, different focal lengths, $\bff \ceqq [f_x, f_y]^\top,~a\ceqq f_y/f_x\neq1$, for each image axis, 3) a non-centered principal point, $\bc \ceqq [c_x, c_y]^\top$, \ie we do not fix $\bc$ to $[W/2, H/2]^\top$, and 4) different, forward and backward \cite{lochman2021babel}, camera models that can be freely chosen at runtime.

Under these characteristics, a general \emph{forward} projection function, $\pi~:~\mathcal{S}^2~\to~\Omega \subset \mathbb{R}^2$, mapping rays in the unit sphere $\bp\ceqq[X, Y, Z]^\top\in\mathcal{S}^2$ to image coordinates $\bx\ceqq[u, v]^\top\in\Omega$ in the image domain $\Omega$, is given by\footnote{To alleviate the notation, we define $f\ceqq f_x$, so $f_y=a f = a f_x$.}
\begin{equation}\label{eq:proj}
    \pi(\bp) = 
    \begin{bmatrix} u \\ v \end{bmatrix} =
    f~\phi(R, Z)
    \begin{bmatrix} X \\ aY \end{bmatrix} +
    \begin{bmatrix} c_x \\ c_y \end{bmatrix}~,
\end{equation}
where $\phi(R, Z)$ is a camera-model specific function of the ray radius $R=\sqrt{X^2 + Y^2}$ and its $Z$ component. 
On the other hand, a general \emph{backward} \cite{lochman2021babel} unprojection function, $\pi^{-1}~:~\Omega~\to~\mathcal{S}^2$, is given by
\begin{equation}\label{eq:unproj}
    \pi^{-1}(\bx) = 
    \begin{bmatrix} X \\ Y \\ Z \end{bmatrix} = 
    \lambda
    \begin{bmatrix} m_x \\ m_y \\ \psi(r) \end{bmatrix}~, 
\end{equation}
where $\lambda$ is a unit-norm normalization constant, $\psi(r)$ is a function that depends on the chosen backward model and:
\begin{equation}\label{eq:proj_aux}
    \begin{bmatrix} m_x \\ m_y \end{bmatrix} \ceqq 
    \begin{bmatrix} (u-c_x)/f \\ (v-c_y)/(af) \end{bmatrix}~, 
    \quad 
    r \ceqq \sqrt{m_x^2 + m_y^2}~.
\end{equation}

\PAR{Principal point and pixel aspect ratio.}
Given the image coordinates $\bx$ and their predicted ray coordinates, we can directly recover $\bc$ and $a$ in closed-form. Either by subtracting $\bc$ from $\bx$ in \cref{eq:proj} and then dividing their entries, or by dividing the first two entries of \cref{eq:unproj}, the model-specific functions $\phi$ and $\psi$, and focal length $f$ disappear, leading to the simplified constraint:
\begin{equation}\label{eq:ppoint_ar}
    uYa - Yac_x + Xc_y = vX~,
\end{equation}
which is well defined for all $\bp\in\mathcal{S}^2$, $\bx\in\Omega$, and depends linearly
on $a$, $ac_x$ and $c_y$. Thus we can recover $\bc$ and $a$ in closed-form solving the corresponding overconstrained linear system and undoing the reparameterization in $ac_x$.

\PAR{Remaining intrinsics.} Given estimations%
\footnote{
Estimating first $a$ and $\bc$ leads to a unified formulation for obtaining intrinsics in closed-form across the implemented models. However, \emph{all} intrinsics of the pinhole, BC \cite{duane1971close} and KB \cite{kannala2006generic} models can be obtained in one single step since manipulations of \cref{eq:proj} are linear on them.}
for $a$ and $\bc$, the \cref{eq:proj,eq:unproj} become linear with respect to the remaining intrinsics of a wide range of commonly used camera models \cite{usenko2018double, lochman2021babel}. This is shown in \cref{tab:cam_models}, where we also indicate the currently implemented models, but our formulation can also be applied to others, \eg the WoodScape model \cite{Yogamani2019woodscape} or the ones proposed by \citet{Scaramuzza2006flexible} and \citet{urban2015calibration}. 
Among them, UCM \cite{mei2007ucm} and EUCM \cite{Khomutenko2016eucm} require special care, as their parameters are bounded ($\xi\geq0~,~\alpha\in[0, 1]~,~\beta>0$). Since the number of bounds is limited, they can be efficiently enforced using a simplified active set method \cite{stark1995bounded}, that checks, when needed, which bounds need to be active. 
Experimentally, we did not found it necessary to enforce $f>0$ as it is well constrained from the dense 2D-3D correspondences between rays and image points.

\PAR{Final refinement.}
The previous closed-form solutions are globally optimal in an \emph{algebraic} parameter space that allows a linear recovery of the intrinsics, but the optimized quantity is not interpretable. Thereby, as the final stage of our method, we iteratively optimize a nonlinear \emph{geometric} quantity, starting from the previous global optimum.  
Specifically, we minimize the angular distance between the rays predicted by the network and those corresponding to the intrinsics, using five Gauss-Newton \cite{triggs2000bundle} iterations.

\subsection{Implementation}
\label{sec:imp}

\begin{figure}
    \centering
    \begin{tblr}{
        width=1.0\linewidth,
        colspec={lc},
        colsep=8pt,
        rowsep=1pt,
        cell{1}{1}={r=3}{l},
        stretch=0,
    }
        \includegraphics[width=4cm, valign=b]{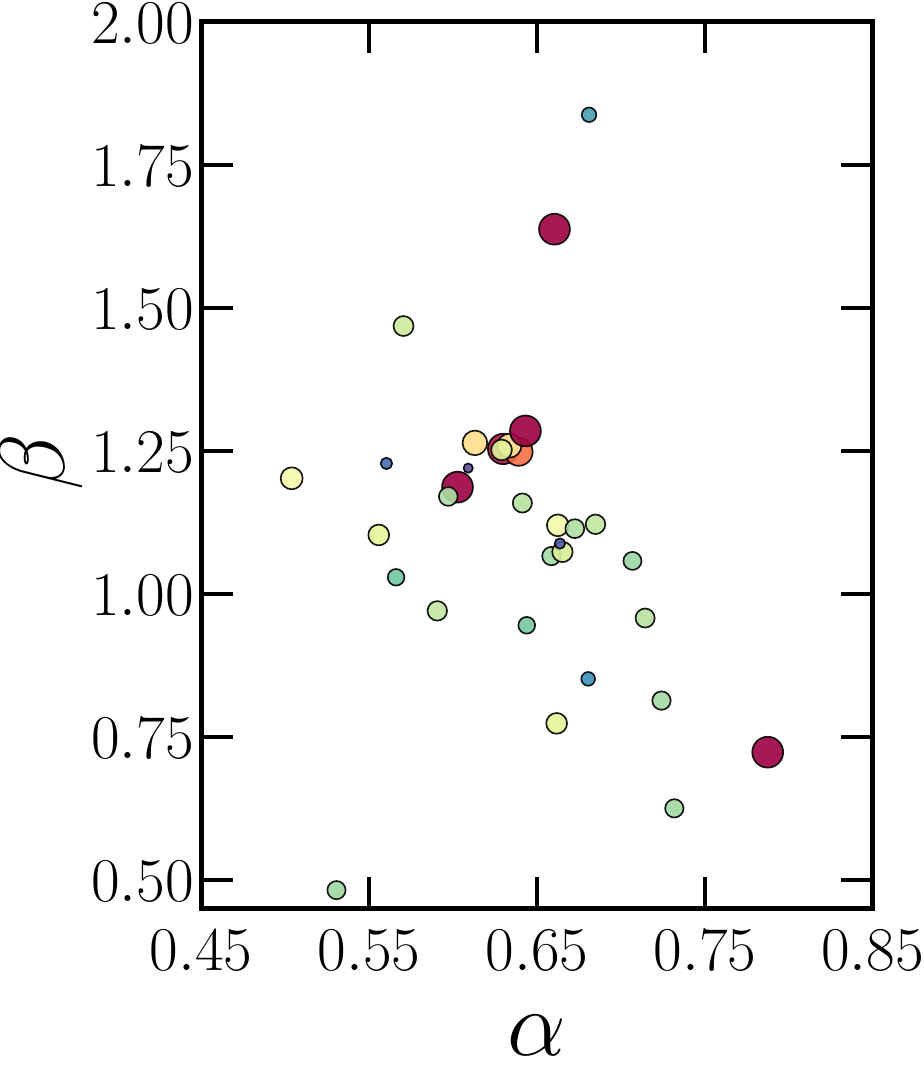} & \includegraphics[width=2.6cm, valign=b]{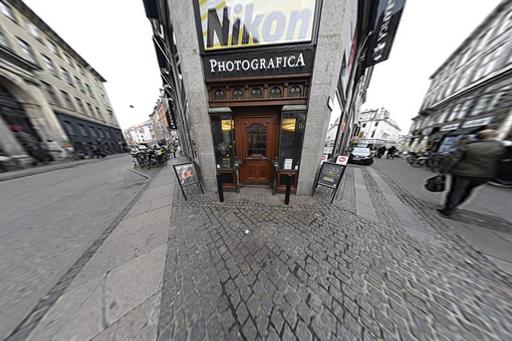} \\
        & \includegraphics[width=3.2cm, valign=t]{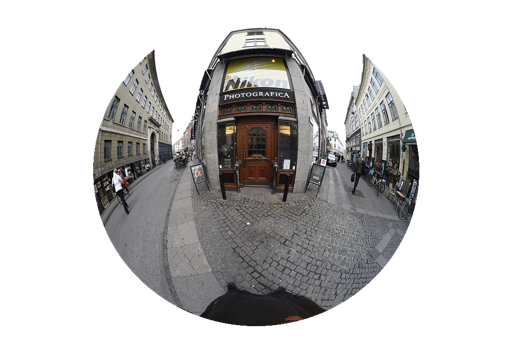} \\
        & $\alpha=0.60$, $\beta=1.19$
    \end{tblr}
    \caption{\textbf{Real-world sample of EUCM intrinsics.} The LensFun database \cite{lensfun} contains a wide range of calibrated lenses. We consider its \emph{fisheye} lenses and map (see \cref{sec:supp_data_details}) their parameters to the EUCM's $\alpha$ and $\beta$ \cite{Khomutenko2016eucm} to define its training bounds. (left) Resulting distribution, colored by vFoV, which ranges from $50^\circ$ to $>180^\circ$. (right) Undistortion using the mapped $\alpha$ and $\beta$ on an image (from \cite{nikon_im}) captured with a Nikon AI-S Fisheye-Nikkor lens.}
    \label{fig:eucm_lensfun}
\end{figure}

\PAR{Architecture and training.} We use a ViT-L backbone \cite{dosovitskiy2021vit} pretrained by DINOv2 \cite{oquab2024dinov2}. Our CNN decoder is based on DPT \cite{Ranftl2021dpt} which we modify to not use expensive transposed convolutions. Instead, we regress FoV fields (\cref{sec:fov}) at $1/7$ of the input resolution and finally upsample them to input resolution using convex upsampling \cite{teed2020raft}.
The backbone and decoder use an initial learning rate of $6\times10^{-6}$ and $6\times10^{-5}$, respectively. We use the AdamW \cite{loshchilov2018adamw} optimizer, with a learning rate that is linearly warmed up during the first $10^3$ steps and decayed with a factor of $0.3$ at $10^4$ and $3\times10^{4}$ steps. We use the same augmentations of \cite{veicht2024geocalib} and a batch size of 24 images with resolution of $~320^2$ pixels and aspect ratios uniformly sampled $\in[0.5,~2.0]$. We train for 40 epochs during 1 day with 3 NVIDIA V100 GPUs. Since estimating the principal point, arbitrarily placed in the image is an ill-posed problem \cite{schonberger2016colmap,agapito1998selfcalib}, for edited (stretched and cropped) images, we train a different model: we do the same as before, except that, following \cite{zhu2023wildcam,he2025diffcalib}, the geometric transformations correspond to uniformly sampling the pixel aspect-ratio $\in[0.5,~2]$ and an image crop of at most half its size.

We supervise the elements, $\btheta$, of the FoV field predictions using a L1 loss function:
\begin{equation}
    \mathcal{L} = \frac{1}{HW} \sum_{i}^{HW}~\lVert \btheta_{\mathrm{GT},~i} - \btheta_i \rVert_1 ~.
\end{equation}
Values of $\btheta_{\mathrm{GT}}$ are obtained with \cref{eq:theta_log} using ray directions/unprojections obtained with ground-truth intrinsics. 

\PAR{Inference.} AnyCalib calibrates a camera in $\sim25$ms in a RTX 4090. At inference, we resize and center-crop the image to the closest resolution and aspect seen during training.

\PAR{Training data.} Our dataset extends OpenPano \cite{veicht2024geocalib}, which consists of 2557 panoramas used to synthetically generate images with fine-grained control over the intrinsics and camera rotation. However, since GeoCalib \cite{veicht2024geocalib} supervises extrinsic parameters, the panoramas need to be aligned with the gravity direction. AnyCalib does not have this restriction, thus we extend OpenPano to 4055 unconstrained and freely-available panoramas from the Laval Indoor dataset
\cite{Bolduc2023lavalindoor,gardner2017lavalindoor}, PolyHaven \cite{polyhaven}, HDRmaps \cite{hdrmaps}, AmbientCG \cite{ambientcg} and BlenderKit \cite{blenderkit}.

As detailed in \cref{sec:supp_data_details}, we create four datasets: \datp, \datr, \datd and \datg, each used to separately train AnyCalib. They differ on the type of images sampled from panoramas:
\begin{itemize}
    \item \datp: Perspective (pinhole), following \cite{veicht2024geocalib},
    \item \datr: Distorted (using \cite{duane1971close}), also following \cite{veicht2024geocalib},
    \item \datd: Distorted \cite{duane1971close} and strongly distorted \cite{Khomutenko2016eucm},
    \item \datg: General \ie perspective, distorted and strongly distorted.
\end{itemize}
Both \datd and \datg are motivated by the limited distortion allowed by the Brown-Conrady camera model \cite{Leotta2022maxradius, usenko2018double}. In contrast, EUCM accurately models strong distortions \cite{usenko2018double,lochman2021babel,Khomutenko2016eucm}. \datg allows us to demonstrate that AnyCalib, trained on general projections, outperforms models trained on specific ones. The sampling of EUCM intrinsics is based on real-lens values extracted from the public LensFun database \cite{lensfun} (\cref{fig:eucm_lensfun}). Each training dataset consists of $54$k images. See \cref{sec:supp_data_details} for more details.

\section{Experiments}
\label{sec:exp}

We evaluate the calibration accuracy of AnyCalib on perspective, edited and distorted images across a wide range of benchmarks, using images not seen during training. We ablate our design choices in \cref{sec:ablation} and present additional qualitative results in \cref{sec:supp_qual}.

\subsection{Perspective images}
\label{sec:exp_p}

\PAR{Metrics} for measuring calibration accuracy should be \emph{agnostic} to the chosen camera model. Otherwise, methods predicting different camera models cannot be fairly compared. A notable example is Perceptual \cite{HoldGeoffroy2023perceptual}, which predicts UCM \cite{mei2007ucm} intrinsics. This model, as well as the majority of models, has a focal length, $f$, parameter. However, its values, while correct, can be an order of magnitude different than its counterpart in other models. We show this in \cref{sec:supp_agnostic_eval}.
Because of this, we use model-agnostic metrics. As it is standard \cite{Jin2023perspective,veicht2024geocalib}, we report median errors and the Area Under the recall Curve (AUC) up to 1/5/10$^\circ$ for the hFoV and vFoV, which we compute in a model-agnostic way (\cref{sec:supp_agnostic_eval}). 
We also report the median of the mean reprojection (RE) and angular unprojection (AE) errors whose mean is computed within a uniform $H\times W$ image grid.

\PAR{Datasets}
We follow GeoCalib \cite{veicht2024geocalib} and evaluate on: i) Stanford2D3D \cite{armeni2017stanford2d3d} which consists of image samples from panoramic images captured inside university buildings, ii) TartanAir's \cite{wang2020tartanair} photo-realistic renderings of scenes with changing light and various weather conditions, iii) MegaDepth's \cite{Li2018megadepth} crowd-sourced images of popular landmarks, with cameras calibrated with SfM, and iv) LaMAR \cite{sarlin2022lamar}, which is an AR dataset for localization, captured over multiple years in university buildings and a city center.

\PAR{Baselines} We consider state-of-the-art methods that calibrate a camera from a single image: GeoCalib \cite{veicht2024geocalib}, Perceptual \cite{HoldGeoffroy2023perceptual}, WildCam \cite{zhu2023wildcam} and DiffCalib \cite{he2025diffcalib}. We also compare AnyCalib with the 3D foundation models DUSt3R \cite{Wang2024dust3r}, UniDepth \cite{Piccinelli2024unidepth} and MoGe \cite{wang2025moge}, which either directly predict intrinsics \cite{Piccinelli2024unidepth}, or predict 3D pointmaps that can be used to calibrate a camera \cite{wang2025moge, Wang2024dust3r}. These foundation models are trained on $>3$ million images. 
Among the baselines, WildCam and DiffCalib also regress rays as in our method. \citet{dalcin2024revisiting} also predict rays, however, we could not evaluate it since it does not have public weights and we could not successfully 
retrain it.

\PAR{Results} in \cref{tab:perspective_results} show that AnyCalib$_\mathrm{pinhole}$ and AnyCalib$_\mathrm{gen}$, trained on \datp and \datg, respectively, outperform alternative single-view calibration methods. 
It also performs similar or better than 3D foundation models \cite{wang2025moge,Wang2024dust3r}, despite AnyCalib is trained on only 54k images and is not tied to perspective projections. 
We show results for distorted images in \cref{sec:exp_d}, that \cite{wang2025moge,Wang2024dust3r} cannot handle.

\begin{table*}[t]
\small

\newcommand{\moge}{MoGe$_{\text{CVPR}'25}$ \cite{wang2025moge}}
\newcommand{\dust}{DUSt3R$_{\text{CVPR}'24}$ \cite{Wang2024dust3r}}
\newcommand{\perc}{Perceptual$_{\text{TPAMI}'23}$ \cite{HoldGeoffroy2023perceptual}}
\newcommand{\geoc}{GeoCalib$_{\text{ECCV}'24}$ \cite{veicht2024geocalib}}
\newcommand{\wild}{WildCam$_{\text{NeurIPS}'23}$ \cite{zhu2023wildcam}}
\newcommand{\difc}{DiffCalib$_{\text{AAAI}'25}$ \cite{he2025diffcalib}}
\newcommand{\unid}{UniDepth$_{\text{CVPR}'24}$ \cite{Piccinelli2024unidepth}}
\newcommand{\anyp}{\textbf{AnyCalib$_{\text{pinhole}}$ (Ours)}}
\newcommand{\anyg}{\textbf{AnyCalib$_{\text{gen}}$ (Ours)}}

\newcommand{\stan}{Stanford2D3D \cite{armeni2017stanford2d3d}}
\newcommand{\tart}{TartanAir \cite{wang2020tartanair}}
\newcommand{\mega}{MegaDepth \cite{Li2018megadepth}}
\newcommand{\lama}{LaMAR \cite{sarlin2022lamar}}

\definecolor{onec}{RGB}{178,255,178}
\definecolor{twoc}{RGB}{255,235,158}

    \centering
    \begin{tblr}{
        colspec={ll*{10}c},
        rowsep=0.3pt,
        cell{1}{5,9}={c=4}{c},
        cell{2}{6,10}={c=3}{c},
        cell{3,12,21,30}{1}={r=9}{c, cmd=\rotatebox{90}},
        cell{10}{3-Z}={bg=onec}, %
        cell{11}{3-Z}={bg=twoc},
        cell{19}{3-Z}={bg=onec}, %
        cell{16}{6}={bg=twoc},
        cell{20}{3-5,7-Z}={bg=twoc},
        cell{28}{3-Z}={bg=onec}, %
        cell{29}{3-Z}={bg=twoc},
        cell{36}{3-Z}={bg=onec}, %
        cell{37}{3-Z}={bg=twoc},
    }
\toprule
& Method & AE \textdownarrow  & RE \textdownarrow & vFoV [$^\circ$] & - & - & - & hFoV [$^\circ$] & - & - & -  \\
\cmidrule[lr]{5-8} \cmidrule[lr]{9-12}
& & [$^\circ$] & [pix] & error\textdownarrow & AUC@1/5/10 \textuparrow & - & - & error\textdownarrow & AUC@1/5/10 \textuparrow & - & - \\
\midrule
\stan & \perc   &   4.67   &  48.81   & 10.52  &    5.60   &   15.20   &  27.10    &   11.35  &    4.90   &  13.70   &  24.90 \\
-     & \wild   &   5.46   &  58.00   & 10.26  &    5.20   &   13.60   &  25.70    &   13.97  &    3.30   &   8.80   &  18.00 \\
-     & \dust   &   2.39   &  27.09   &  5.34  &   10.70   &   26.10   &  43.80    &    6.07  &    9.60   &  23.50   &  40.30 \\
-     & \unid   &   6.55   &  73.60   & 14.65  &    3.20   &    8.50   &  17.20    &   16.72  &    2.80   &   7.70   &  15.00 \\
-     & \geoc   &   1.45   &  15.16   &  3.23  &   17.40   &   39.90   &  59.40    &    3.53  &   15.60   &  37.20   &  56.50 \\
-     & \difc   &   6.35   &  82.25   & 13.63  &    4.30   &   10.80   &  20.30    &   16.49  &    3.70   &   8.70   &  17.00 \\
-     & \moge   &   1.92   &  21.60   &  4.33  &   16.10   &   33.70   &  49.50    &    4.99  &   14.00   &  30.30   &  46.00 \\
-     & \anyp   &   1.13   &  12.11   &  2.55  &   21.20   &   46.80   &  64.60    &    2.88  &   19.50   &  43.30   &  61.80 \\
-     & \anyg   &   1.30   &  13.90   &  2.95  &   20.80   &   43.60   &  61.60    &    3.24  &   18.60   &  40.30   &  58.90 \\
\midrule[gray8]
\tart                 & \perc   &   3.54   &  28.47   &  8.01  &    4.80   &   28.00   &  39.40    &    8.37  &    4.90   &  28.60   &  38.80 \\
-                     & \wild   &   7.40   &  83.53   & 19.25  &    0.90   &    2.30   &   6.30    &   16.07  &    0.60   &   2.20   &   7.30 \\
-                     & \dust   &   5.49   &  56.45   & 12.37  &    1.40   &    3.80   &   9.20    &   13.27  &    1.40   &   3.70   &   8.10 \\
-                     & \unid   &  13.99   & 204.17   & 32.16  &    0.80   &    1.60   &   3.00    &   35.10  &    0.60   &   1.50   &   3.10 \\
-                     & \geoc   &   2.18   &  19.36   &  4.91  &   13.80   &   30.50   &  47.80    &    5.11  &   13.60   &  29.60   &  46.50 \\
-                     & \difc   &  12.96   & 178.98   & 30.28  &    0.00   &    0.00   &   0.00    &   31.49  &    0.00   &   0.00   &   0.00 \\
\SetRow{gray9}        & \moge   &   0.73   &   6.24   &  1.61  &   26.60   &   67.00   &  83.20    &    1.68  &   25.40   &  65.70   &  82.50 \\
-                     & \anyp   &   1.63   &  15.02   &  3.62  &   15.50   &   36.40   &  55.10    &    3.93  &   14.70   &  34.70   &  53.00 \\
-                     & \anyg   &   1.76   &  15.98   &  4.06  &   13.60   &   33.40   &  52.80    &    4.09  &   13.90   &  33.40   &  52.20 \\
\midrule[gray8]
\mega                 & \perc   &   2.54   &  81.69   &  6.21  &    8.80   &   22.60   &  39.80    &    6.52  &    8.00   &  20.80   &  37.60 \\
\SetRow{gray9}        & \wild   &   1.65   &  57.24   &  2.90  &   18.60   &   42.90   &  63.10    &    3.42  &   15.30   &  37.80   &  59.30 \\
\SetRow{gray9}        & \dust   &   0.77   &  30.75   &  1.82  &   31.70   &   56.70   &  72.30    &    1.96  &   30.00   &  54.40   &  70.60 \\
                      & \unid   &   4.51   & 128.07   & 10.82  &    6.80   &   16.20   &  27.40    &   11.10  &    5.80   &  14.90   &  25.30 \\
-                     & \geoc   &   1.88   &  57.77   &  4.56  &   13.80   &   31.60   &  48.10    &    4.93  &   13.00   &  30.20   &  46.40 \\
-                     & \difc   &   3.26   & 113.65   &  4.26  &   14.20   &   31.80   &  51.30    &    5.56  &    9.70   &  24.40   &  43.80 \\
\SetRow{gray9}        & \moge   &   1.12   &  28.58   &  2.16  &   25.40   &   53.50   &  72.20    &    2.31  &   24.50   &  50.80   &  69.90 \\
-                     & \anyp   &   1.31   &  39.51   &  3.14  &   19.40   &   40.80   &  59.10    &    3.36  &   18.00   &  39.30   &  57.20 \\
-                     & \anyg   &   1.48   &  47.11   &  3.57  &   14.80   &   36.60   &  55.70    &    3.84  &   14.90   &  34.50   &  53.20 \\
\midrule[gray8]
\lama  & \perc   &   2.51   &  91.78   &  6.70  &    7.00   &   13.90   &  31.60    &    5.68  &    7.00   &  15.90   &  36.20 \\
-      & \wild   &   1.79   &  52.30   &  2.85  &   18.80   &   43.90   &  66.60    &    2.72  &   18.80   &  44.00   &  64.00 \\
-      & \dust   &   2.25   &  84.22   &  5.99  &    5.40   &   17.20   &  42.00    &    5.07  &    6.30   &  21.40   &  50.20 \\
-      & \unid   &   1.14   &  36.99   &  2.46  &    5.10   &   37.10   &  49.10    &    2.89  &    1.50   &  32.40   &  47.90 \\
-      & \geoc   &   1.17   &  39.99   &  3.09  &   19.00   &   41.20   &  59.80    &    2.65  &   22.00   &  45.80   &  64.00 \\
-      & \difc   &   4.18   & 163.83   & 11.26  &    0.00   &    1.50   &  15.50    &    5.34  &    8.20   &  22.00   &  47.40 \\
-      & \moge   &   0.69   &  25.64   &  1.97  &   26.30   &   55.30   &  73.50    &    1.69  &   31.70   &  60.60   &  77.50 \\
-      & \anyp   &   0.85   &  28.61   &  2.25  &   24.60   &   51.60   &  70.50    &    1.92  &   28.60   &  57.10   &  74.60 \\
-      & \anyg   &   1.08   &  35.88   &  2.81  &   19.30   &   44.10   &  65.00    &    2.49  &   21.00   &  48.30   &  68.70 \\
\bottomrule
    \end{tblr}
    \caption{\textbf{Results on perspective images.} Best and second-best results are highlighted in
    \bgcolor{onec}{green}
    and 
    \bgcolor{twoc}{yellow}%
    , respectively. Methods trained on evaluated datasets are highlighted in 
    \bgcolor{gray9}{gray}
    and excluded from the ranking to ensure a fair comparison. Among all evaluated methods, only Perceptual \cite{HoldGeoffroy2023perceptual} and AnyCalib$_{\text{gen}}$ are not specifically trained on perspective (pinhole) images. AnyCalib, either trained on perspective-only (AnyCalib$_{\text{pin}}$) or general (AnyCalib$_{\text{gen}}$) projections, performs similarly or better than alternatives, including the 3D foundation models DUSt3R \cite{Wang2024dust3r}, UniDepth \cite{Piccinelli2024unidepth} and MoGe \cite{wang2025moge}, despite AnyCalib is trained on orders of magnitude less data.} 
    \label{tab:perspective_results}
\end{table*}

\begin{table*}[t]
\small

\newcommand{\perc}{Perceptual$_{\text{TPAMI}'23}$ \cite{HoldGeoffroy2023perceptual}}
\newcommand{\geop}{GeoCalib-pin$_{\text{ECCV}'24}$ \cite{veicht2024geocalib}}
\newcommand{\geor}{GeoCalib-rad$_{\text{ECCV}'24}$ \cite{veicht2024geocalib}}
\newcommand{\svaa}{SVA$_{\text{WACV}'21}$ \cite{Lochman2021sva}}
\newcommand{\anyr}{\textbf{AnyCalib$_{\text{radial}}$ (Ours)}}
\newcommand{\anyg}{\textbf{AnyCalib$_{\text{gen}}$ (Ours)}}
\newcommand{\anyd}{\textbf{AnyCalib$_{\text{dist}}$ (Ours)}}

\newcommand{\mega}{MD \cite{Li2018megadepth}}
\newcommand{\scan}{SN++ \cite{Yeshwanth2023scannetpp}}
\newcommand{\mono}{Mono \cite{engel2016monodataset}}

\definecolor{onec}{RGB}{178,255,178}
\definecolor{twoc}{RGB}{255,235,158}

    \centering
    \begin{tblr}{
        colspec={ll*{10}c},
        rowsep=0.1pt,
        cell{3}{1}={r=5}{c, cmd=\rotatebox{90}},
        cell{8,14}{1}={r=6}{c, cmd=\rotatebox{90}},
        cell{1}{5,9}={c=4}{c},
        cell{2}{6,10}={c=3}{c},
        cell{6}{3-Z}={bg=onec},
        cell{7}{3-Z}={bg=twoc},
        cell{6}{5,7,9}={bg=twoc},
        cell{7}{8,5,7,9}={bg=onec},
        cell{8}{6,7,10}={bg=onec},
        cell{8}{11}={bg=twoc},
        cell{12}{3,4,5,8,9,11,12}={bg=onec},
        cell{12}{6,7,10}={bg=twoc},
        cell{13}{3,4,5,8,9,12}={bg=twoc},
        cell{18}{3-Z}={bg=onec},
        cell{19}{3-Z}={bg=twoc},
        cell{19}{6,11}={bg=onec},
    }
\toprule
& Method & AE \textdownarrow  & RE \textdownarrow & vFoV [$^\circ$] & - & - & - & hFoV [$^\circ$] & - & - & -  \\
\cmidrule[lr]{5-8} \cmidrule[lr]{9-12}
& & [$^\circ$] & [pix] & error\textdownarrow & AUC@1/5/10 \textuparrow & - & - & error\textdownarrow & AUC@1/5/10 \textuparrow & - & - \\
\midrule
\mega & \perc                        &    2.63  &   82.97  &   6.22  &   8.40  &  22.20  &  39.50  &  6.88   &   7.50  &  20.20  &  36.60  \\
-     & \geop                        &    1.97  &   55.91  &   4.67  &  15.00  &  31.70  &  47.70  &  5.03   &  13.80  &  30.40  &  46.10  \\
-     & \geor                        &    1.93  &   56.14  &   4.55  &  14.20  &  31.70  &  47.70  &  5.01   &  13.50  &  30.40  &  46.20  \\
-     & \anyr                        &    1.48  &   45.89  &   3.63  &  16.40  &  36.70  &  55.90  &  3.81   &  14.70  &  35.00  &  54.40  \\
-     & \anyg                        &    1.45  &   44.63  &   3.55  &  15.20  &  37.10  &  55.90  &  3.79   &  14.00  &  34.60  &  53.80  \\
\midrule[gray8]
\scan & \svaa                        &    2.92  &   37.64  &   6.39  &  21.00  &  33.80  &  42.30  &   6.80  &  17.60  &  31.50  &  41.20  \\
-     & \perc                        &    3.10  &   44.08  &   7.65  &   0.80  &   1.70  &  22.10  &   3.96  &   2.60  &  24.20  &  39.10  \\
-     & \geop                        &    7.00  &  104.72  &  14.61  &   1.70  &   5.40  &  12.80  &  20.41  &   2.00  &   4.60  &  10.10  \\
-     & \geor                        &    4.97  &   71.07  &  10.11  &   3.80  &  11.30  &  23.70  &  14.20  &   3.40  &   9.90  &  18.50  \\
-     & \anyd                        &    1.59  &   20.61  &   3.90  &  11.20  &  31.80  &  57.70  &   3.38  &  15.80  &  39.10  &  62.60  \\ %
-     & \anyg                        &    1.88  &   24.75  &   4.26  &  10.50  &  29.50  &  55.10  &   5.05  &   6.90  &  23.70  &  47.50  \\ %
\midrule[gray8]
\mono & \svaa                        &    8.29  &  107.61  &  19.09  &  12.20  &  22.70  &  31.00  &  19.64  &  11.00  &  21.30  &  29.80  \\
-     & \perc                        &    4.06  &   53.22  &   9.48  &   2.10  &  15.90  &  26.70  &   9.09  &  10.20  &  14.60  &  26.60  \\
-     & \geop                        &    4.38  &   63.63  &  10.10  &   5.00  &  13.90  &  26.40  &  11.73  &   4.40  &  12.50  &  23.40  \\
-     & \geor                        &    3.94  &   56.49  &   7.90  &   8.40  &  19.50  &  33.70  &   9.47  &   6.60  &  16.20  &  28.80  \\
-     & \anyd                        &    1.60  &   21.70  &   3.60  &  16.20  &  37.40  &  57.30  &   3.94  &  15.00  &  34.20  &  53.80  \\ %
-     & \anyg                        &    1.64  &   22.52  &   3.66  &  16.20  &  36.80  &  56.20  &   3.95  &  14.50  &  34.20  &  53.00  \\ %
\bottomrule
    \end{tblr}
    \caption{\textbf{Results on distorted images.} Best and second-best results are highlighted in \bgcolor{onec}{green} and \bgcolor{twoc}{yellow}, respectively. 
    AnyCalib, either trained on distorted-only images or general image projections is generally more accurate than alternative methods. Since AnyCalib is model-agnostic, different camera models can be fitted without retraining. To demonstrate this, results in MegaDepth \cite{Li2018megadepth}, ScanNet++ \cite{Yeshwanth2023scannetpp} and Mono \cite{engel2016monodataset} use, respectively: 
    the radial (Brown-Conrady) model \cite{duane1971close} with one distortion parameter, the Kannala-Brandt model \cite{kannala2006generic} with four distortion parameters and UCM \cite{mei2007ucm,geyer2000ucm} (the model to which Perceptual \cite{HoldGeoffroy2023perceptual} is tailored).
    Changing the camera model yields practically the same accuracy, thus providing flexibility in selecting the camera model that best suits a certain application.} 
    \label{tab:distorted_results}
\end{table*}

\begin{table*}[t]
\small

\newcommand{\wild}{WildCam$_{\text{NeurIPS}'23}$ \cite{zhu2023wildcam}}
\newcommand{\difc}{DiffCalib$_{\text{AAAI}'25}$ \cite{he2025diffcalib}}
\newcommand{\anyc}{\textbf{AnyCalib (Ours)}}

\newcommand{\stan}{Stanford2D3D \cite{armeni2017stanford2d3d}}
\newcommand{\tart}{TartanAir \cite{wang2020tartanair}}
\newcommand{\lama}{LaMAR \cite{sarlin2022lamar}}

\definecolor{onec}{RGB}{178,255,178}
\definecolor{twoc}{RGB}{255,235,158}

    \centering
    \begin{tblr}{
        colspec={l*{9}c},
        rowsep=0.1pt,
        cell{1}{2,5,8}={c=3}{c},
        cell{5}{2-Z}={bg=onec},
        vline{2,5,8}={solid, gray9}
    }
    \toprule
    Method & \stan & - & - & \tart & - & - & \lama & -  & - \\
           & $e_f$ \textdownarrow & $e_c$ \textdownarrow & AE [$^\circ$] \textdownarrow 
           & $e_f$ \textdownarrow & $e_c$ \textdownarrow & AE [$^\circ$] \textdownarrow 
           & $e_f$ \textdownarrow & $e_c$ \textdownarrow & AE [$^\circ$] \textdownarrow \\
    \midrule
    \wild  & 0.32 & 0.47 & 12.74     & 0.83 & 0.38 & 15.60     & 0.25 & 0.51 & 10.38 \\ 
    \difc  & 0.40 & 0.55 & 14.43     & 0.62 & 0.55 & 18.01     & 0.24 & 0.55 & 11.35 \\
    \anyc  & 0.15 & 0.26 &  7.46     & 0.19 & 0.28 &  9.88     & 0.22 & 0.41 & 9.46  \\
    \bottomrule
    \end{tblr}
    \caption{\textbf{Results on edited (stretched and cropped) images.} Best results are highlighted in \bgcolor{onec}{green}. All evaluated methods estimate the same camera model (pinhole with four degrees of freedom).  Compared with the recent WildCam \cite{zhu2023wildcam} and DiffCalib \cite{he2025diffcalib}, AnyCalib consistently estimates the focal lengths $f_x, f_y$, and the arbitrarily placed principal point $\bc$ more accurately.} 
    \label{tab:edited_results}
\end{table*}

\subsection{Distorted images}
\label{sec:exp_d}

\PAR{Metrics.} 
We consider the same metrics as in \cref{sec:exp_p}.

\PAR{Datasets.}
We follow \cite{veicht2024geocalib} and evaluate on distorted images from MegaDepth \cite{Li2018megadepth}. Since their distortion is limited, we also evaluate on subsets of 2000 randomly sampled images from the indoor DSLR sequences of ScanNet++ \cite{Yeshwanth2023scannetpp}, and from the last 16 outdoor and indoor sequences\footnote{Which are the ones without black borders caused by the absence of light hitting the image sensor. No method has been trained with this effect.} of the SLAM dataset from \cite{engel2016monodataset}, which we refer to as Mono.

\PAR{Baselines.}
We consider methods evaluated in \cref{sec:exp_p} that work with distorted images: Perceptual \cite{HoldGeoffroy2023perceptual} and GeoCalib \cite{veicht2024geocalib}. We also consider SVA \cite{Lochman2021sva}, which is a geometric method.
For \cite{veicht2024geocalib}, we consider its two models, trained with perspective (pin) and radially-distorted (rad) images. We use its implementation of the radial model in MegaDepth \cite{Li2018megadepth}, and, for the other datasets, we use the division model, since they present strong radial distortions and wider field of views, not appropriate for the radial model \cite{Leotta2022maxradius, usenko2018double}. 
We only evaluate SVA \cite{Lochman2021sva} in datasets with strong distortions, as this is an assumption of this method. Note that SVA leads to crashes (lack of estimations) in approximately 38\% of images in both ScanNet++ \cite{Yeshwanth2023scannetpp} and Mono \cite{engel2016monodataset}, as it fails to detect either circular arcs or repeated patterns in the input images, as similarly reported in \cite{veicht2024geocalib, HoldGeoffroy2023perceptual}.

\PAR{Results} on \cref{tab:distorted_results} show that AnyCalib, either trained on \datr (AnyCalib$_{\mathrm{radial}}$), \datd (AnyCalib$_{\mathrm{dist}}$) or \datg (AnyCalib$_{\mathrm{gen}}$), is generally more accurate than the alternatives. On ScanNet++ \cite{Yeshwanth2023scannetpp}, the accuracy of AnyCalib and SVA \cite{lochman2021babel} is comparable. However, AnyCalib is more robust, as SVA leads to crashes in 38\% of the images.

\subsection{Edited images}

\PAR{Baselines.}
We consider WildCam and DiffCalib \cite{zhu2023wildcam,he2025diffcalib} as they are the only baselines that can deal with edited images.

\PAR{Metrics.} Since both \cite{zhu2023wildcam,he2025diffcalib} estimate pinhole intrinsics, we follow their evaluation and report median values of relative errors for focal and principal point, defined as: $e_f=\lVert (\bff_{\mathrm{GT}} - \bff)/\bff_{\mathrm{GT}} \rVert_\infty$ and $e_c=2\lVert (\bc_{\mathrm{GT}} - \bc)/ [W,~H] \rVert_\infty$, with the division being done elementwise. To compare with previous results we also report the angular error.

\PAR{Datasets.}
We consider the same datasets as in \cref{sec:exp_p}, excluding MegaDepth, as it already contains crops and \cite{zhu2023wildcam} is trained on it. Following \cite{zhu2023wildcam,he2025diffcalib}, we randomly resize the images to have pixel-aspect ratios $\in[0.5,~2]$ and randomly crop them to, at most, half of its size.

\PAR{Results} on \cref{tab:edited_results} confirm that AnyCalib trained on \datp with a training strategy based on \cite{zhu2023wildcam,he2025diffcalib} (\cref{sec:imp}), also improves over alternative methods on stretched and cropped images. Note that estimating the principal point, arbitrarily placed in the image, is challenging and recognized as an ill-posed problem \cite{schonberger2016colmap, agapito1998selfcalib}. This explains the increase in AE.

\section{Conclusion}
\label{sec:conclusion}

In this paper, we introduced AnyCalib, a novel single-view calibration method that, for the first time, works with edited, perspective and distorted images. We frame the calibration of a camera as the regression of FoV fields, a novel, robust, intermediate representation not tied to extrinsic cues and that is bijective to the pixelwise ray directions of the camera. AnyCalib is model-agnostic, as it calibrates, in closed-form, a wide range of camera models freely chosen at runtime. Experimentally, AnyCalib sets a new state-of-the-art across multiple indoor and outdoor benchmarks.

\section*{Acknowledgements}
Thanks to Jean-François Lalonde for granting us permission to release our models trained on the Laval Photometric Indoor HDR Dataset.
The top row of \cref{fig:teaser} uses images from \href{https://explorecams.com/photos/model/iphone-xs}{ExploreCams}, taken by \href{https://explorecams.com/photos/pDWj91ExFb?lens=iphone-xs-back-dual-camera-6mm-f-2-4}{Michele\_Sacchet}, \href{https://explorecams.com/photos/WkEuJktL9z?lens=iphone-xs-back-dual-camera-6mm-f-2-4}{srkcalifano} and \href{https://explorecams.com/photos/2xjsF8XuiV?lens=iphone-xs-back-dual-camera-6mm-f-2-4}{donchc}.
The input image in \cref{fig:method} has a \href{https://creativecommons.org/licenses/by/2.0/}{CC BY 2.0 license} and its author is \href{https://www.flickr.com/photos/sergeigussev/49521010167/}{Sergei Gussev}.
This work was supported by the Ministerio de Universidades Scholarship FPU21/04468, the Spanish Government (projects PID2021-127685NB-I00 and TED2021-131150B-I00) and the Aragón Government
(project T45\_23R).

{
    \small
    \bibliographystyle{ieeenat_fullname}
    \bibliography{main}
}

\clearpage
\newpage
\maketitlesupplementary

\appendix

\section{Ablations}
\label{sec:ablation}
We report ablation results in \cref{tab:ablation}. Experiments 1-4. are conducted by training AnyCalib on \datp and averaging errors across the benchmarks of \cref{sec:exp_p}. The fifth, RANSAC, ablation, is performed on ScanNet++, following \cref{sec:exp_d}. MACs are computed for a $280\times364$ input image, which results from resizing an image with a 3:4 ($H$:$W$) aspect ratio to the training resolution of $320^2$ pixels.

\PAR{1-2. Intermediate representation.} We test the performance of AnyCalib when learning rays instead of our proposed FoV fields (\cref{sec:fov}). As first baseline, we use the target representation (rays) and loss function of WildCam \cite{zhu2023wildcam}, which is a cosine similarity loss. As a stronger baseline, we evaluate also the training strategy of DSINE \cite{bae2024dsine} for learning rays \ie using an angular loss. Compared to these baselines, FoV fields lead to more accurate calibrations.

\PAR{3. Decoder architecture.} Our proposed light DPT decoder, when compared to the original \cite{Ranftl2021dpt}, decreases $\sim$20\% the computation and leads to slight accuracy improvements.

\PAR{4. Dataset extension.} Our extended version of OpenPano \cite{veicht2024geocalib} leads to improved accuracy. This experiment shows that AnyCalib is scalable.

\PAR{5. RANSAC \cite{fischler1981ransac}}  can also be applied to our derivations in \cref{sec:calib} by using minimal samples from the set of 2D-3D correspondences between the regressed rays and image points. However, minimal samples lead to inaccurate intrinsics when fitting high-complexity camera models such as Kannala-Brandt (KB) \cite{kannala2006generic}. This motivates our non-minimal estimation of the intrinsics.

In conclusion, FoV fields are an appropriate intermediate representation for calibration and key to the performance of AnyCalib: their supervision, when compared to alternatives, leads to learning patterns that are more useful. Moreover, since FoV fields are not tied to extrinsic cues, this is what has allowed us to extend the training dataset with panoramas not aligned with the gravity direction.

\begin{table}[t]
\small

\newcommand{\anyc}{\textbf{AnyCalib}}
\newcommand{\rayw}{1. Learning rays as \cite{zhu2023wildcam}}
\newcommand{\rayd}{2. Learning rays as \cite{bae2024dsine}}
\newcommand{\dptd}{3. Original DPT \cite{Ranftl2021dpt}}
\newcommand{\data}{4. Orig. OpenPano \cite{veicht2024geocalib}}
\newcommand{\rsac}{5. RANSAC in KB \cite{kannala2006generic}}

\definecolor{onec}{RGB}{178,255,178}

    \centering
    \begin{tblr}{
        colspec={l*{4}c},
        rowsep=0.5pt,
        cell{1}{3}={c=2}{c},
        cell{2}{2-Z}={bg=onec},
        cell{3,4,6}{5}={bg=onec},
        cell{7}{2-4}={bg=onec},
    }
\toprule
Experiment  & RE  & \{v, h\}FoV  & - & MACs \\
\midrule
\anyc    &  23.81  &  2.89  &  3.02  & 187.8G \\
\rayw    &  26.05  &  3.13  &  3.24  & 187.8G \\
\rayd    &  24.95  &  3.02  &  3.13  & 187.8G \\
\dptd    &  24.99  &  3.00  &  3.12  & 243.2G \\
\data    &  26.62  &  3.23  &  3.35  & 187.8G \\
\midrule[gray8]
\anyc    &   20.61  &  3.90  &  3.38  & n/a \\
\rsac    &    1019  &  16.1  &  16.2  & n/a \\
\bottomrule
    \end{tblr}
    \caption{\textbf{Ablation study} over representation, decoder design, dataset and intrinsics fitting method. See \cref{sec:ablation} for details.} 
    \label{tab:ablation}
\end{table}

\begin{table*}[t]
\small

    \centering
    \begin{tblr}{
        colspec={ll*{4}c},
    }
\toprule
Data & Models & FoV [$^\circ$] & $\hat{k}=kH/f$ & $\alpha$ & $\beta$ \\
\midrule
\datp & 100\% pinhole                        & $\mathcal{U}(20, 105)$ & - & - & - \\
\midrule[gray9]
\datr & 100\% BC \cite{duane1971close}       & $\mathcal{U}(20, 105)$ & $\mathcal{N}_t(0,~0.07,~[-0.3,~0.3])$ & - & - \\
\midrule[gray9]
\datd & 50\% BC \cite{duane1971close}        & $\mathcal{U}(20, 105)$ & $\mathcal{N}_t(0,~0.07,~[-0.3,~0.3])$ & - & - \\
      & 50\% EUCM \cite{Khomutenko2016eucm}  & $\mathcal{U}(50, 180)$ & - & $\mathcal{U}(0.5, 0.8)$ & $\mathcal{U}(0.5, 2)$ \\
\midrule[gray9]
\datg & 34\% pinhole                         & $\mathcal{U}(20, 105)$ & - & - & - \\
      & 33\% BC \cite{duane1971close}        & $\mathcal{U}(20, 105)$ & $\mathcal{N}_t(0,~0.07,~[-0.3,~0.3])$ & - & - \\
      & 33\% EUCM \cite{Khomutenko2016eucm}  & $\mathcal{U}(50, 180)$ & - & $\mathcal{U}(0.5, 0.8)$ & $\mathcal{U}(0.5, 2)$ \\
\bottomrule
    \end{tblr}
    \caption{\textbf{Sampling distributions within the datasets.} $\mathcal{U}(a,~b)$ denotes a uniform distribution $\in[a,~b]$. $\mathcal{N}_t(\mu,~\sigma,~[a,~b])$ denotes a normal distribution $\mathcal{N}(\mu,~\sigma)$ truncated at $[a,~b]$. \textbf{\datp} and \textbf{\datr} follow the setup of GeoCalib \cite{veicht2024geocalib}. Since the distortions allowed by the Brown-Conrady (radial) model \cite{duane1971close} are limited \cite{Leotta2022maxradius,usenko2018double}, for creating \textbf{\datd} and \textbf{\datg}, we use EUCM \cite{Khomutenko2016eucm} to generate strongly distorted images. The limits for sampling its parameters $\alpha$ and $\beta$ are based on real-lens values from the public LensFun database \cite{lensfun} (\cref{fig:eucm_lensfun}).} 
    \label{tab:data}
\end{table*}

\section{Datasets details}
\label{sec:supp_data_details}

As mentioned in \cref{sec:imp}, we create four datasets. We separately train AnyCalib in each of them to study its accuracy according to the trained projection models. The intrinsics used to create the datasets are detailed in \cref{tab:data}. For the camera rotations\footnote{For panoramas not aligned with the gravity direction, these rotations are only approximate.} we follow GeoCalib \cite{veicht2024geocalib} and uniformly sample the roll and pitch angles within $\pm45^\circ$. All datasets are formed by sampling 16 square images in each of the 3651/202/202 training/val/test panoramas, which yields an approximate distribution of 54k/3k/3k training/val/test images per dataset.

\PAR{Obtaining the focal length.} As shown in \cref{tab:data}, we do not directly sample the focal length $f$. Instead, to ensure a uniform distribution of image FoVs, we indirectly sample it from the rest of the parameters. For pinhole images, we use the well-known conversion $f=(H/2)/\tan(\mathrm{FoV}/2)$.
For BC \cite{duane1971close} and EUCM \cite{Khomutenko2016eucm}, we note that, from \cref{eq:proj}:
\begin{equation}
    f = \frac{H/2}{R~\phi(R,~Z)}~,
    \quad
    \begin{array}{l}
        R = \sin(\mathrm{FoV}/2)~, \\
        Z = \cos(\mathrm{FoV}/2)~,
    \end{array}
\end{equation}
since we form the datasets with square images ($H=W$), unit aspect ratio and centered principal point. During training, images are geometrically transformed on-the-fly to match the training resolution and sampled aspect-ratio.

\begin{figure*}[t]
    \centering
    \includegraphics[width=0.99\linewidth]{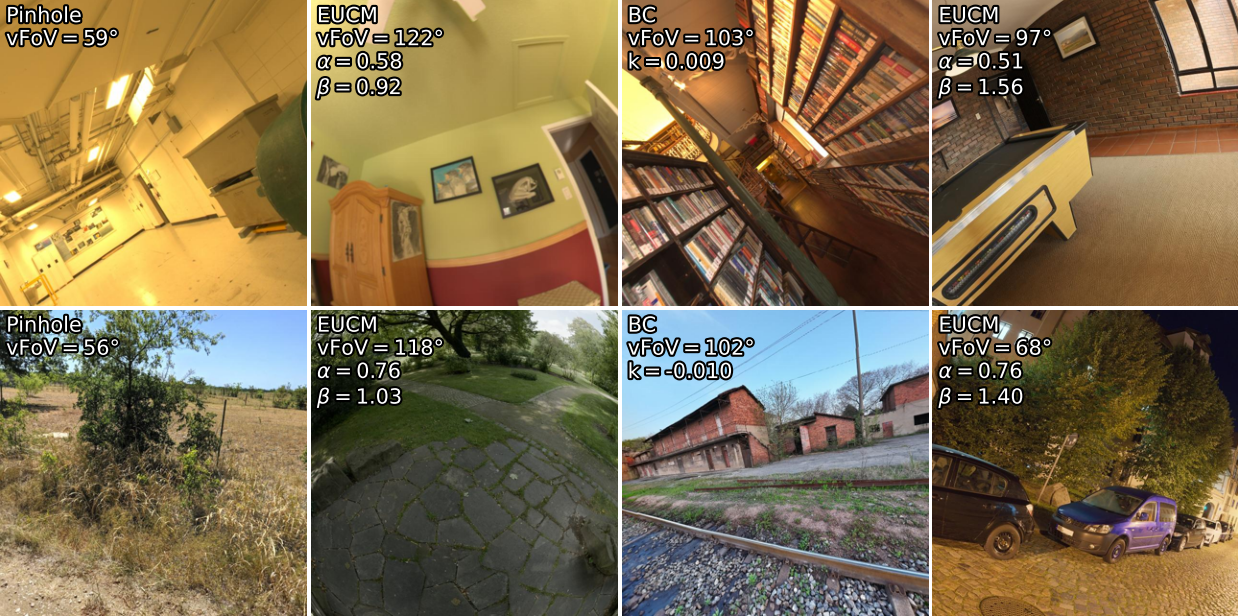}
    \caption{\textbf{Sample} images and intrinsics from the dataset \datg.}
    \label{fig:supp_data_samples}
\end{figure*}

\PAR{Ensuring valid intrinsics.} 
Independently sampling intrinsics of BC \cite{duane1971close} and EUCM \cite{Khomutenko2016eucm} can lead to projection models that project different, distant, rays to the same image coordinates \cite{Leotta2022maxradius}, which is not physically valid. We guard for this by clamping $f$ according to its limits \cite{Leotta2022maxradius, usenko2018double}:
\begin{align}
    \mathrm{BC}  & \to ~ f \geq \begin{cases}
        0 & \text{if } k \geq 0~, \\
        \frac{r_{\mathrm{im}}}{r_{\mathrm{max}}(1+kr_\mathrm{max}^2)} & \text{if } k < 0~,
    \end{cases} \\
    \mathrm{EUCM} & \to ~  f \geq \begin{cases}
        0 & \text{if } \alpha \leq 0.5~, \\
        r_{\mathrm{im}}\sqrt{\beta(2\alpha-1)} & \text{if } \alpha > 0.5~,
    \end{cases}
\end{align}
where $r_{\mathrm{im}}=0.5(H^2+W^2)^{0.5}$ and $r_{\mathrm{max}} = 1/\sqrt{-3k}$ .

\PAR{Mapping LensFun coefficients.} As explained in \cref{sec:imp,fig:eucm_lensfun}, we use the LensFun database \cite{lensfun} for defining the sampling bounds of EUCM's $\alpha$ and $\beta$. LensFun uses its own polynomial distortion models\footnote{Explained in {\scriptsize\url{https://lensfun.github.io/manual/v0.3.2/group__Lens.html\#gaa505e04666a189274ba66316697e308e}}}, so we need to map them. 
Conveniently, our formulation in \cref{sec:calib} is applicable: given normalized image coordinates (obtained with the lens focal and image sensor size) and their unprojected rays, we can linearly recover $\alpha$ and $\beta$. For getting these unprojections, we first undistort a uniform grid of image/sensor coordinates using Newton's root finding algorithm and finally invert the ideal (equisolid, equidistant, orthographic or stereographic \cite{lensfun}) fisheye projection model of the lens.

\PAR{Sample datapoints} of \datg are shown in \cref{fig:supp_data_samples}.

\begin{figure*}[t]
    \centering
    \begin{tblr}{
        width=\linewidth,
        colspec={ccc},
        colsep=1pt,
    }
        original image & w/o fixing $f$ & fixing $f$ \\ 
        \includegraphics[width=5.7cm]{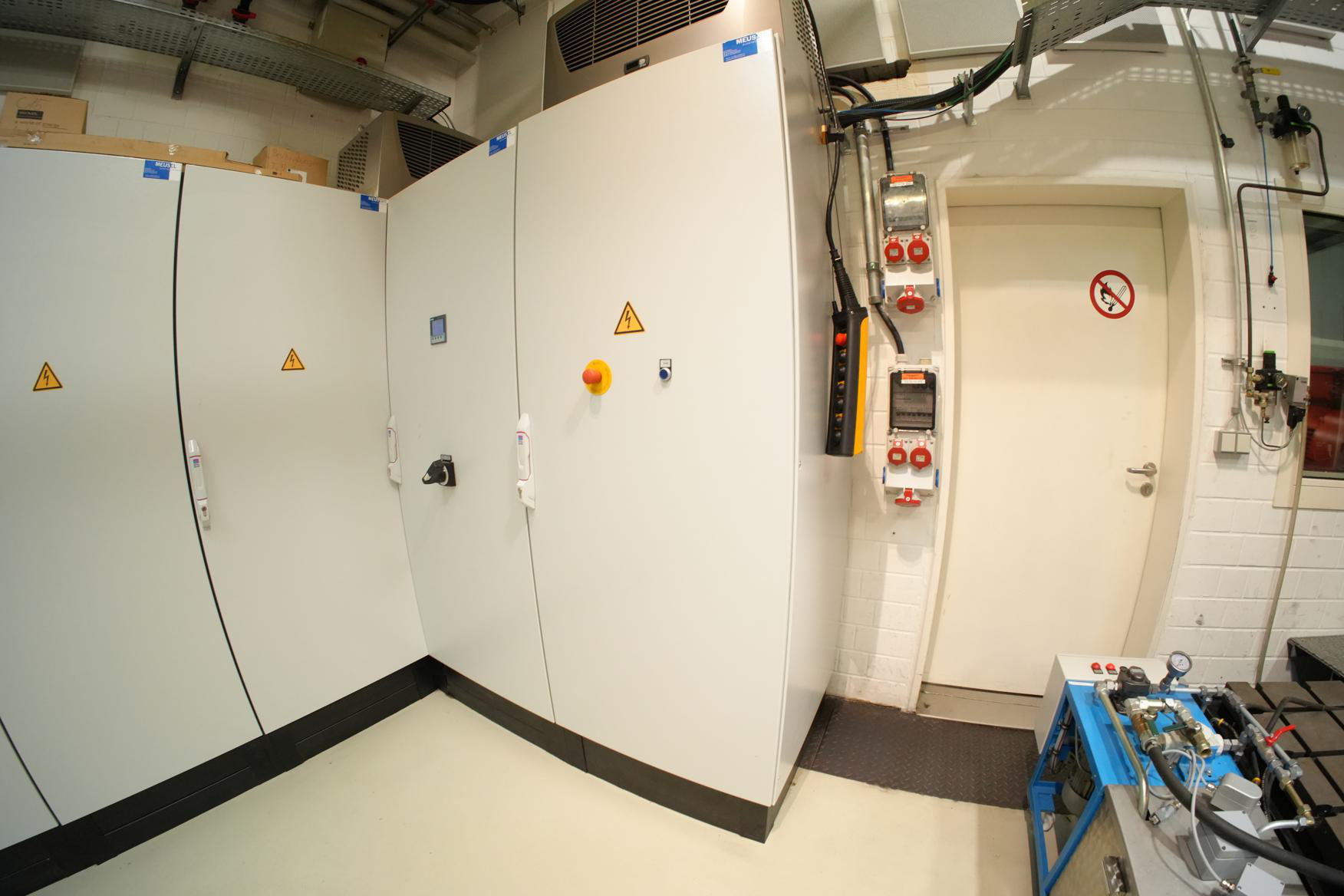} & 
        \includegraphics[width=5.7cm]{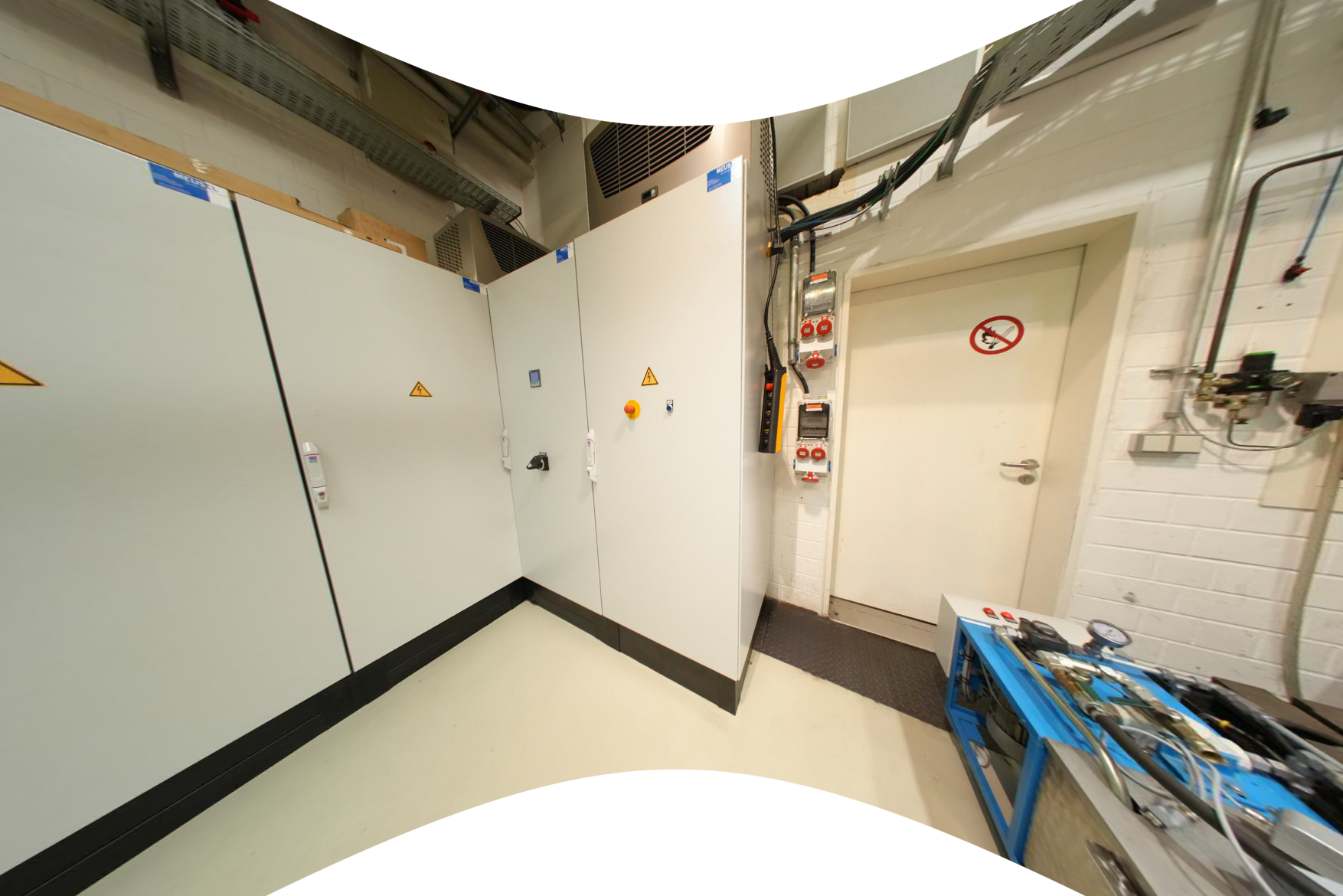} &   
        \includegraphics[width=5.7cm]{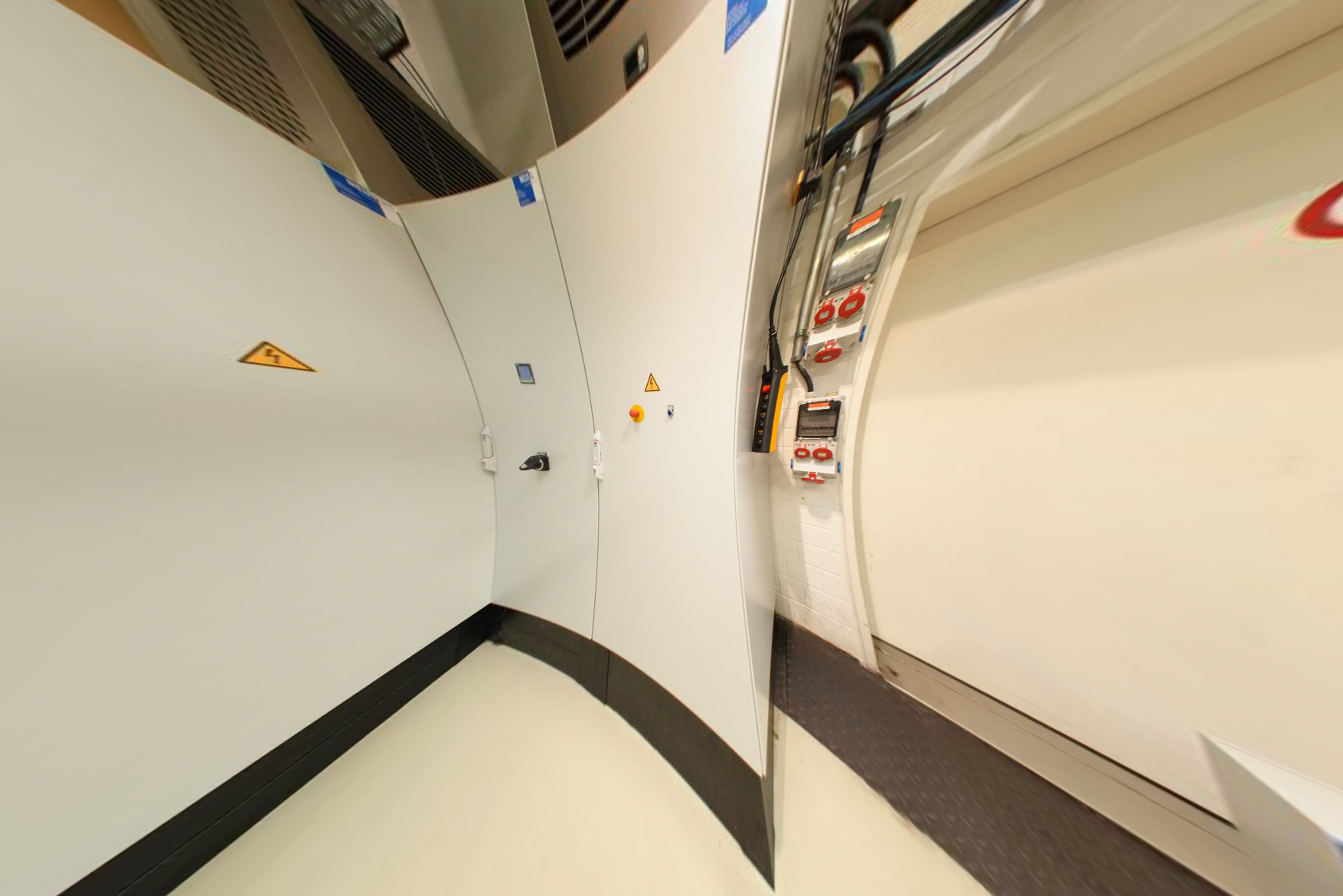} \\  
        $\mathbf{f}=[616.1,~616.1]$ &
        $\mathbf{f}=[1331.9,~1331.9]$ & 
        $\mathbf{f}=[616.1,~616.1]$  \\
        $\mathbf{k}=10^{-2}~[6.0,~0.61,~0.06,-0.03]$ &
        $\xi=1.17$ &
        $\xi=0.88$
    \end{tblr}
    \caption{\textbf{The focal length ($f$) in different camera models} can take significantly different values. We show this for UCM \cite{mei2007ucm}. We map the KB \cite{kannala2006generic} intrinsics corresponding to an image from ScanNet++ \cite{Yeshwanth2023scannetpp} (left) to UCM following \cref{sec:calib}. 
    We do this without fixing $f$, \ie, also mapping it to UCM (middle) and fixing it (right).
    The resulting intrinsics are used to undistort the image.
    The same KB focal for UCM leads to a model that does not truthfully model the lens, leading to a failed undistortion. The converse occurs when also mapping $f$.}
    \label{fig:ucm_focal}
\end{figure*}

\section{Model-agnostic evaluation}
\label{sec:supp_agnostic_eval}

\PAR{Intrinsics in different models.} Different camera models, can have an order of magnitude difference in their focal length $f$ values \cite[Tab. 3]{usenko2018double}. We visualize this behavior in 
\cref{fig:ucm_focal}
by mapping the ground-truth Kannala-Brandt (KB) \cite{kannala2006generic} intrinsics from ScanNet++ \cite{Yeshwanth2023scannetpp} to UCM intrinsics using our formulation from \cref{sec:calib}. As shown, if we fix the ground-truth KB focal length and only map the distortion coefficients, the resulting UCM intrinsics fail to accurately model the camera lens projection, leading to an undistortion failure. The converse occurs when we also map the focal length.

\PAR{Model-agnostic FoV.} The horizontal (hFoV) and vertical 
\begin{wrapfigure}{r}{3cm}
    \includegraphics[width=3cm]{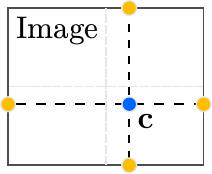}
\end{wrapfigure} 
(vFoV) angular extents of an image can be computed independently of the camera model. To compute the hFoV, we unproject the rays located at the left and right borders, based on the location of the principal point, $\bc$ (yellow points on the  schematic on the right), and sum the angles between them and the optical axis. The vFoV is computed similarly, but using the top and bottom borders instead.

\section{Linear constraints}\label{sec:supp_constraints}
\citet{lochman2021babel} show that the distortion parameters of a wide range of camera models can be estimated linearly from 1D-1D correspondences between the radii on the retinal plane, $\lVert(\bx-\bc)/f\rVert$, and the ray radii, $\sqrt{X^2+Y^2}$. Building on this, we show in this section that, together with \cref{eq:ppoint_ar}, \emph{all the intrinsics} of a wide range of standard camera models can be linearly recovered from 2D-3D correspondences between image coordinates $\bx\in\Omega$ and ray directions in $\Stwo$. 

To obtain the remaining linear constraints, presented in \cref{tab:cam_models}, we first define auxiliary variables according to the notation in \cref{sec:calib}:
\begin{align}
    & R_a \ceqq \sqrt{X^2 + a^2Y^2}~, \quad 
      r_c \ceqq\lVert\bx-\bc\rVert~, \\
    & \theta \ceqq \atantwo(R, Z)~, \quad 
      r \ceqq \sqrt{m_x^2 + m_y^2}~, \\
    & d \ceqq \sqrt{R^2+Z^2}~, \quad
    r_{ca}^2 \ceqq (u-c_x)^2+(v-c_y)^2/a^2~.
\end{align}
For forward camera models (\cref{eq:proj}), the linear constraints derive from \cref{eq:deriv_aux}:
\begin{align}
    & \pi(\bp) = \bx =
    f~\phi(R, Z) \begin{bmatrix}
        X \\
        a Y
    \end{bmatrix} + \bc~, \\
    \implies 
    & \lVert \bx - \bc \rVert = 
    f~\phi(R, Z)
    \left\lVert  
    \begin{bmatrix}
        X \\
        a Y
    \end{bmatrix}
    \right\lVert~, \\
    \implies 
    & r_c = f~\phi(R, Z)~R_a~. \label{eq:deriv_aux}
\end{align}
by substituting the corresponding model-specific function $\phi(R, Z)$ from \cref{tab:cam_models}.

\PAR{Pinhole:} $r_c = \frac{f}{Z} R_a \implies R_a f = Z r_c~.$

\PAR{Brown-Conrady:}
\begin{align}
    &r_c = f \frac{R_a}{Z}
    \left(1 + \sum_{n=1}^N k_n (R/Z)^{2n} \right)~,\\
    \implies &
    r_cZ / f - R_a \sum_{n=1}^N k_n (R/Z)^{2n} = R_a~.
\end{align}

\PAR{Kannala-Brandt:}
\begin{align}
    &r_c = f \frac{R_a}{R}(\theta + \sum_{n=1}^N k_n \theta^{2n+1})~, \\
    \implies &
    Rr_c/f - R_a\sum_{n=1}^N k_n\theta^{2n+1} = R_a \theta~.
\end{align}

\PAR{UCM:}
\begin{equation}
    r_c = f\frac{R_a}{\xi d + Z} \implies R_af - r_cd\xi = r_cZ~.
\end{equation}
    
\PAR{EUCM} For this camera model, linearity in \cref{eq:deriv_aux} is lost when $f$ is unknown. Instead, to estimate $f$, we use a proxy camera model \cite{kannala2006generic} that leads to practically the same focal length value \cite{usenko2018double, lochman2021babel}. Thus, instead of \cref{eq:proj_aux} we start from $r = \phi(R, Z) R$, which for the EUCM model, leads to:
\begin{align}
    r &= \frac{R}{\alpha \sqrt{\beta R^2 + Z^2} + (1-\alpha)Z}~, \\
    \implies &
    r\alpha \sqrt{\beta R^2 + Z^2} = R - (1-\alpha)Zr~, \\
    \implies &
    r^2\alpha^2(\beta R^2+Z^2) = (R^2 - (1-\alpha)Zr)^2~, \\
    \implies &
    r^2R^2 \alpha^2\beta +2rZ (rZ-R)\alpha = (R-rZ)^2~.
\end{align}

\PAR{Division} For this backward model, from \cref{eq:unproj} we know that
\begin{equation}\label{eq:supp_div}
    f \begin{bmatrix} X \\ Y \\ Z \end{bmatrix} = 
    \lambda
    \begin{bmatrix} (u-c_x) \\ (v-c_y)/a \\ f \psi(r) \end{bmatrix}~.
\end{equation}
To remove the nonlinearity stemming from $\lambda$, we use an approach similar to DLT \cite{Hartley2004mvg}. Since both sides must be parallel, their cross product is the null vector, which leads to the following constraints:
\begin{equation}
    (f + \sum_{n=1}^N k'_n r_{ca}^{2n}) 
    \begin{bmatrix}
    X \\
    aY
    \end{bmatrix} =
    Z (\bx - \bc)~,
\end{equation}
with $k'_n\ceqq k_n/f^{2n-1}$. As inferred from \cref{eq:supp_div}, these two equations are linearly dependent. Thus, we consider only the norm of both sides, which results in:
\begin{equation}
    R_a(f +\sum_{n=1}^N k'_n r_{ca}^{2n}) = Zr_c~.
\end{equation}

\section{Additional qualitative results}\label{sec:supp_qual}
\begin{figure*}[t]

    \newlength{\figh}
    \setlength{\figh}{3cm}

    \definecolor{codecolor}{RGB}{246, 248, 250}
    \newcommand{\code}[1]{\colorbox{codecolor}{\texttt{#1}}}

    \newcommand{\lfnamebase}{figures/supp_qualitative_perspective/amusement-amusement-Easy-P003-image_left-001402_left}
    \newcommand{\rfnamebase}{figures/supp_qualitative_perspective/endofworld-endofworld-Easy-P003-image_left-000317_left}

    \centering
    \begin{tblr}{
        width=1.0\linewidth,
        vspan=even,
        colspec={*{6}c},
        colsep=1pt,
        rowsep=1pt,
        cell{2}{1}={r=2}{c, cmd=\rotatebox{90}},
        cell{4}{1}={c, cmd={\rotatebox[origin=c]{90}}},
        cell{5,6,7}{2}={c, cmd={\rotatebox[origin=c]{90}}},
        cell{5}{1}={r=3}{c, cmd=\rotatebox{90}},
    }
        && ground-truth & prediction & ground-truth & prediction \\
        FoV field & $\theta_x$ & 
        \includegraphics[height=\figh,valign=m]{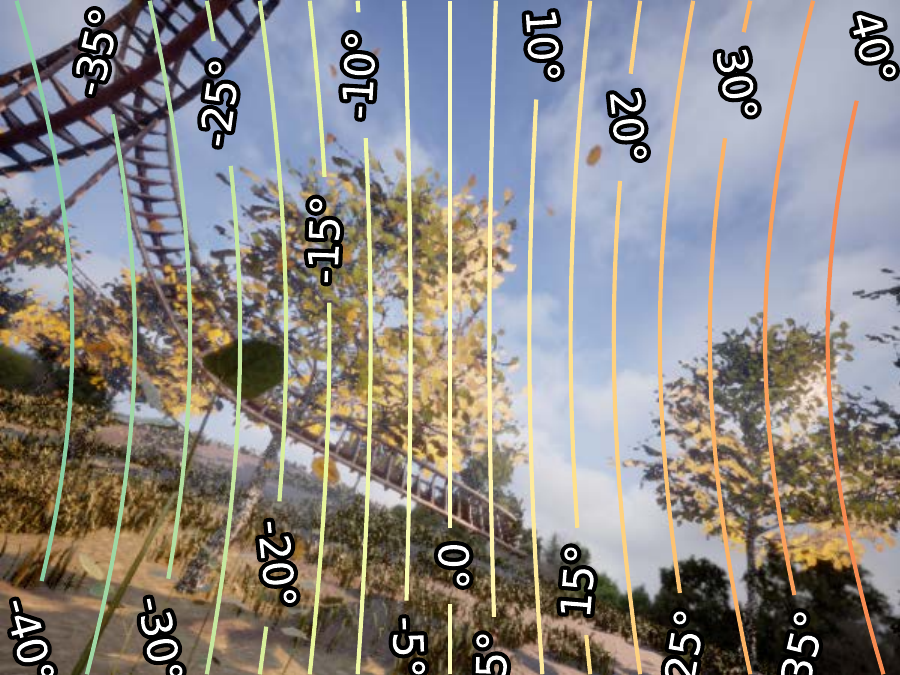} & 
        \includegraphics[height=\figh,valign=m]{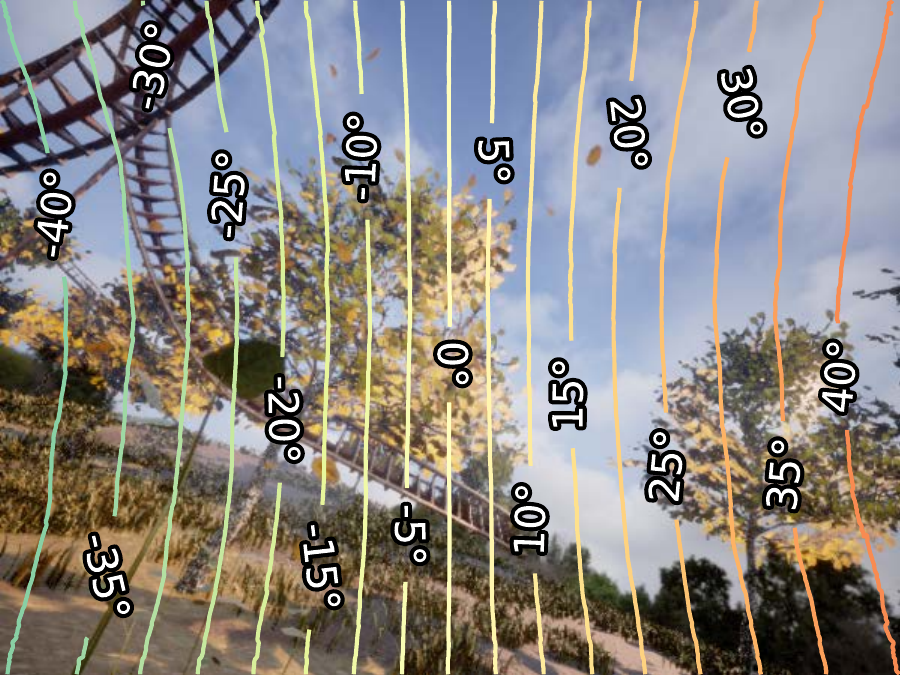} & 
        \includegraphics[height=\figh,valign=m]{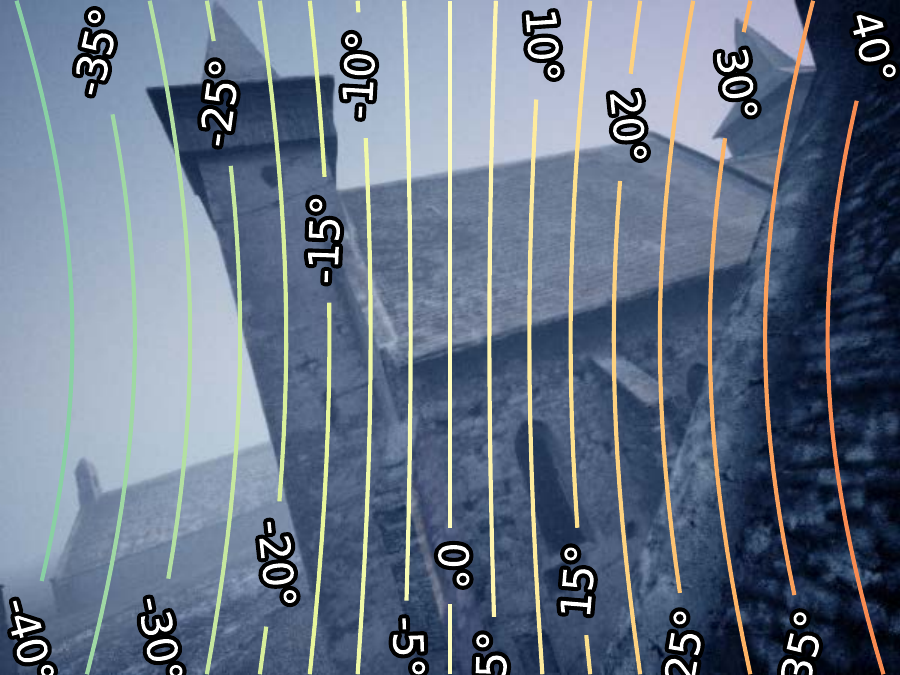} & 
        \includegraphics[height=\figh,valign=m]{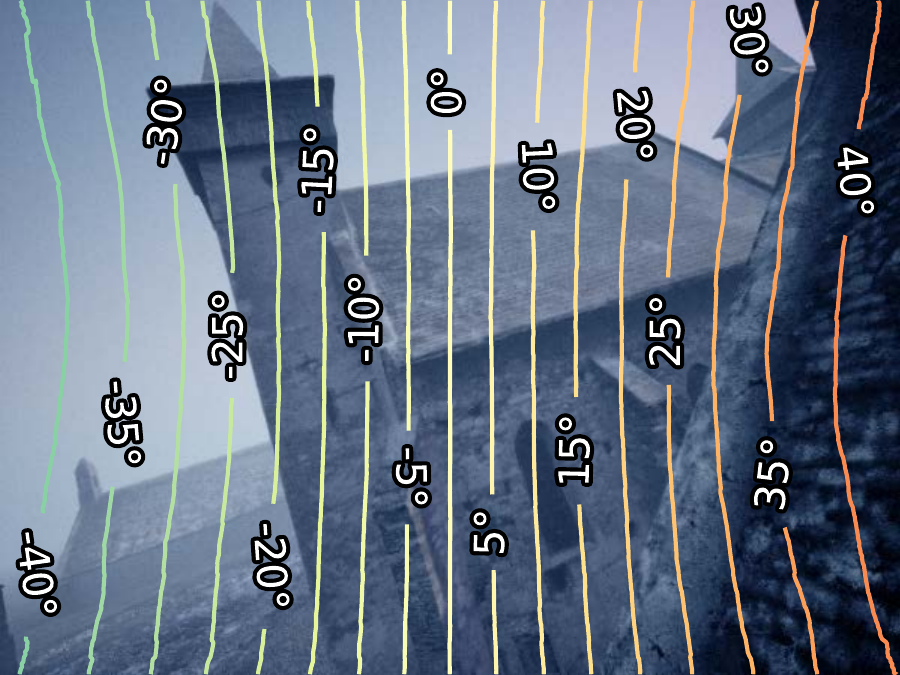} \\
        -         & $\theta_y$ & 
        \includegraphics[height=\figh,valign=m]{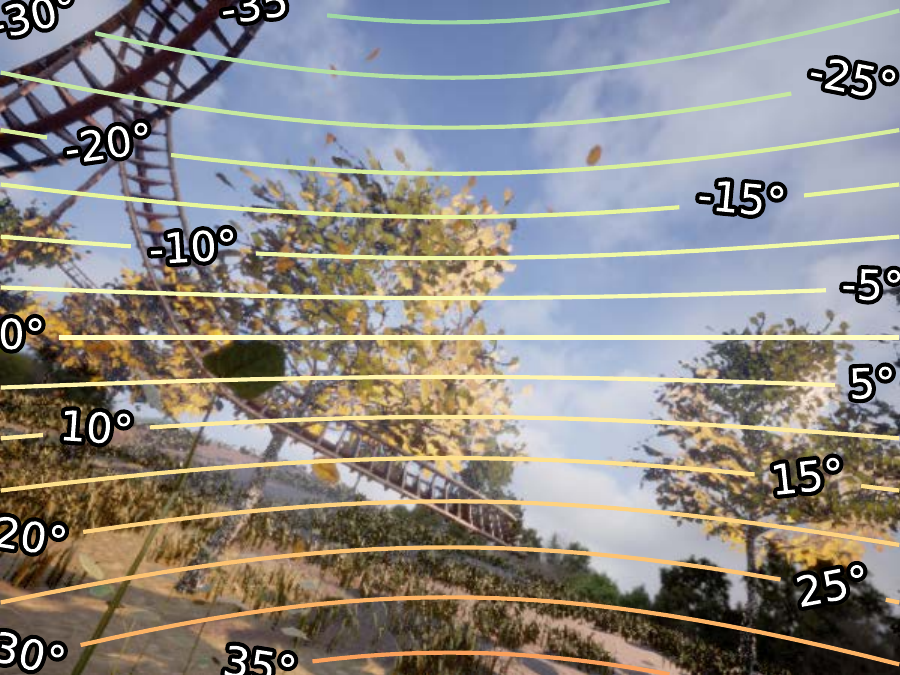} & 
        \includegraphics[height=\figh,valign=m]{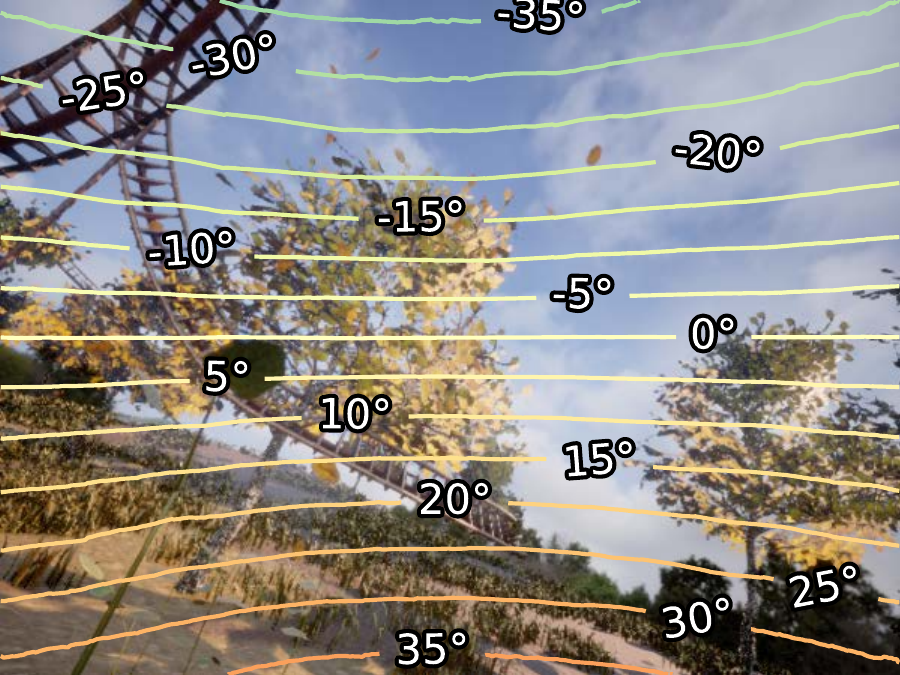} &
        \includegraphics[height=\figh,valign=m]{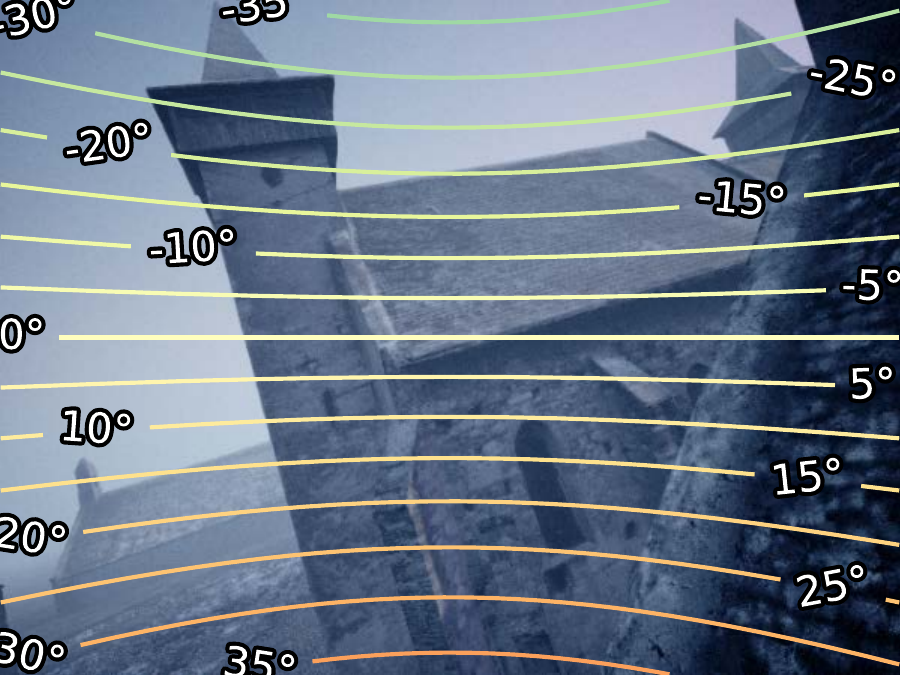} & 
        \includegraphics[height=\figh,valign=m]{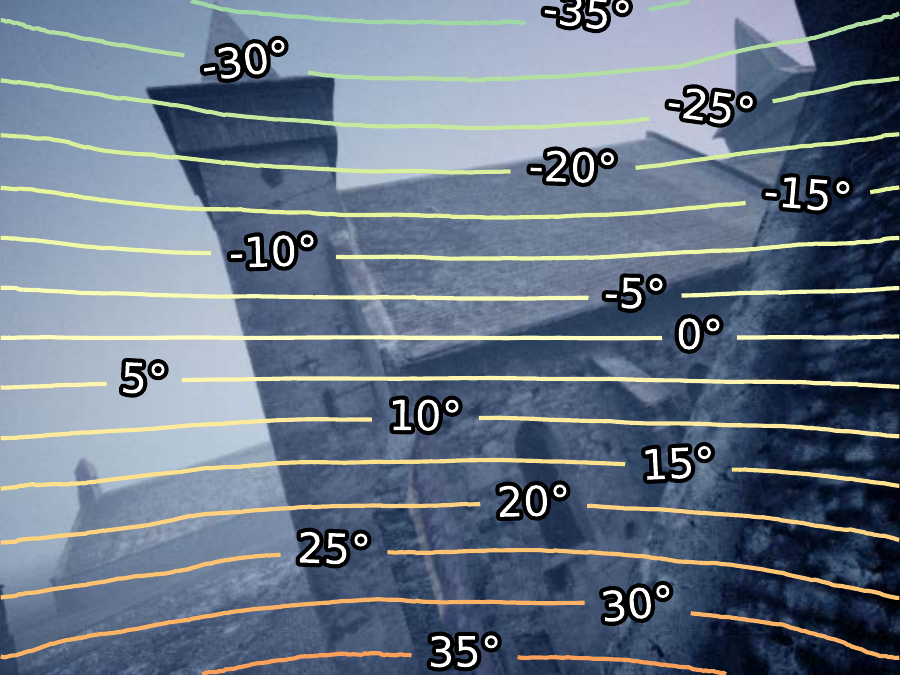} \\
        polar angles $\lVert\boldsymbol{\theta}\rVert$ & & 
        \includegraphics[height=\figh,valign=m]{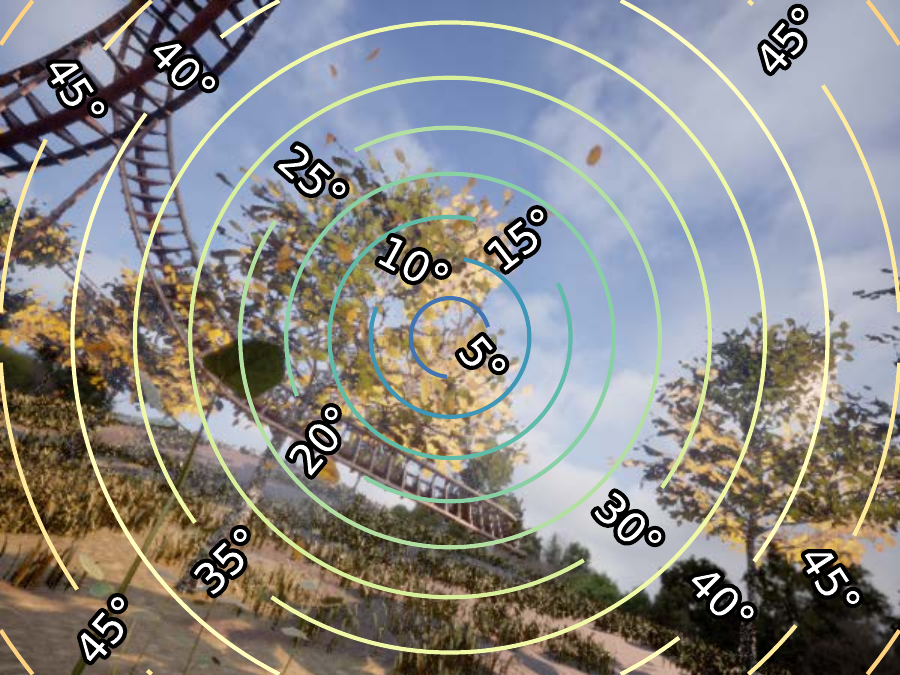} & 
        \includegraphics[height=\figh,valign=m]{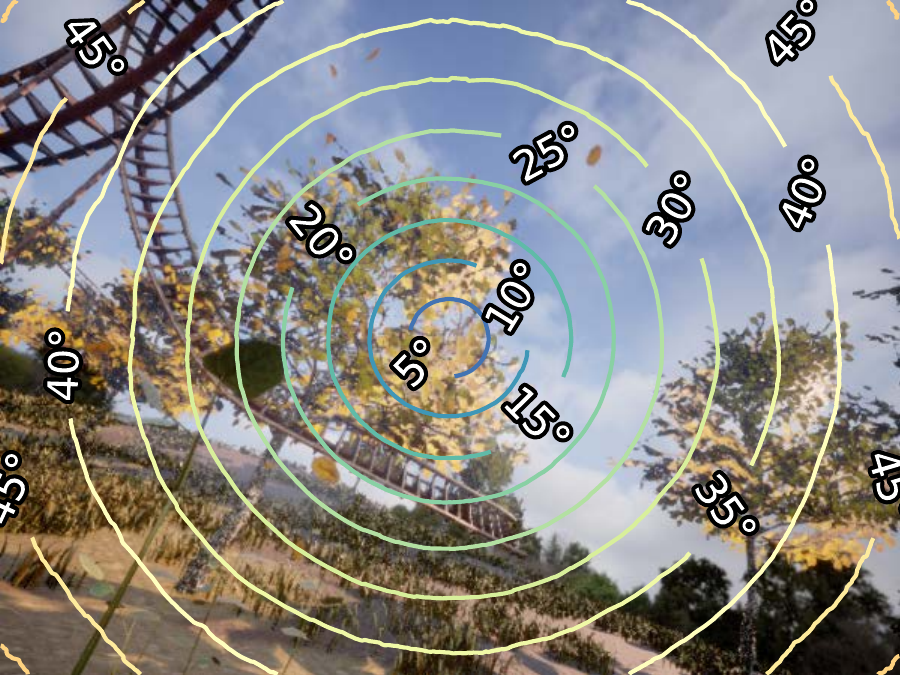} & 
        \includegraphics[height=\figh,valign=m]{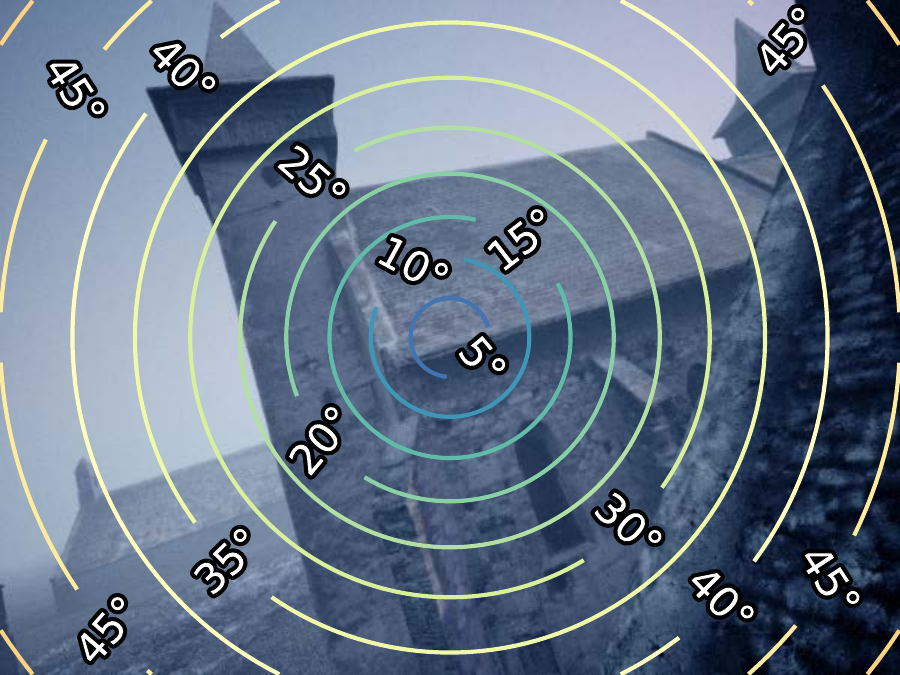} & 
        \includegraphics[height=\figh,valign=m]{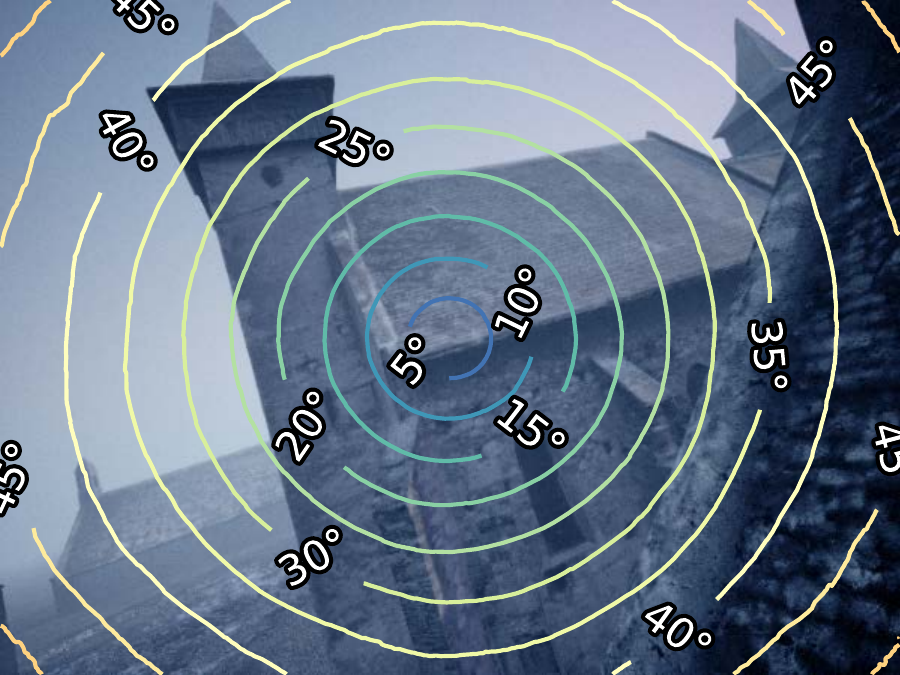} \\
        fits & \code{pinhole} & 
        \includegraphics[height=\figh,valign=m]{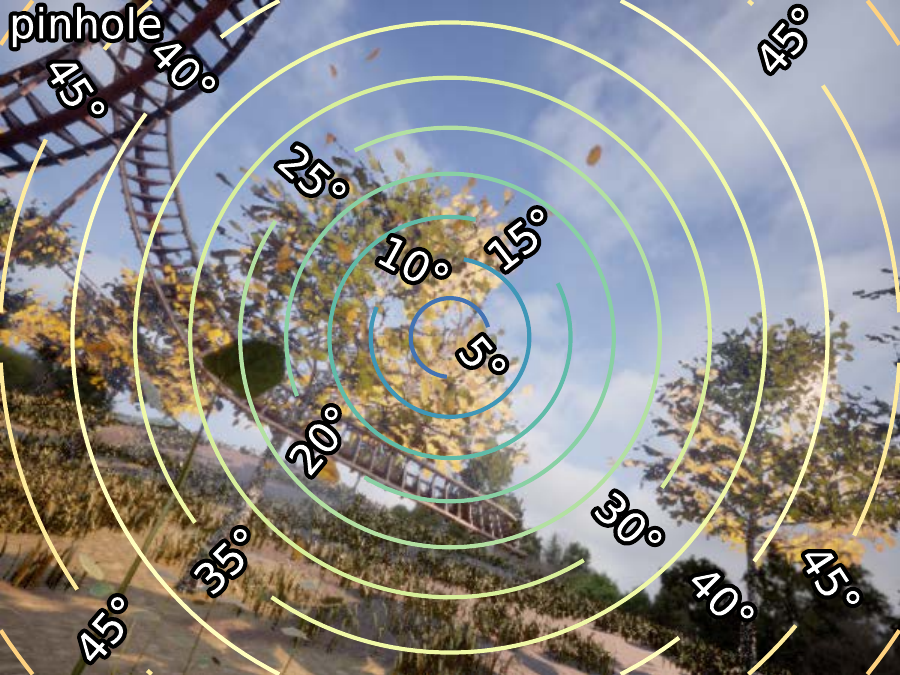} & 
        \includegraphics[height=\figh,valign=m]{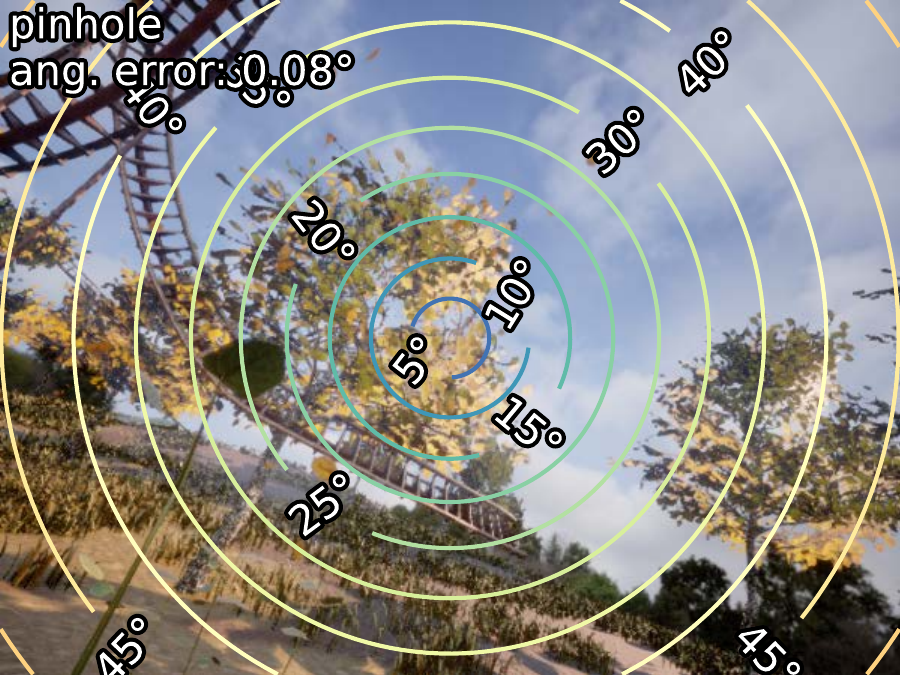} &
        \includegraphics[height=\figh,valign=m]{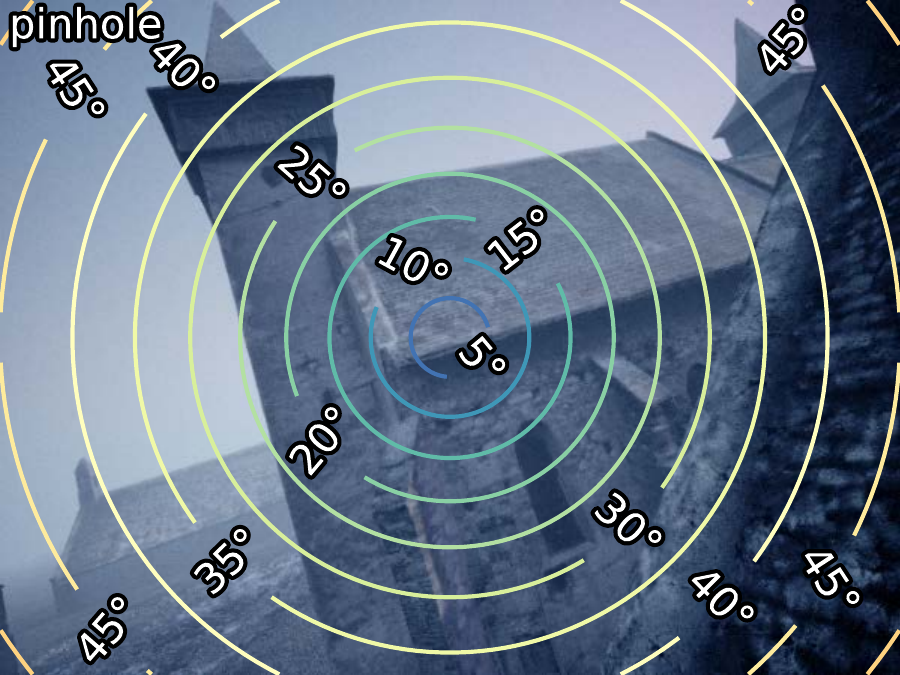} & 
        \includegraphics[height=\figh,valign=m]{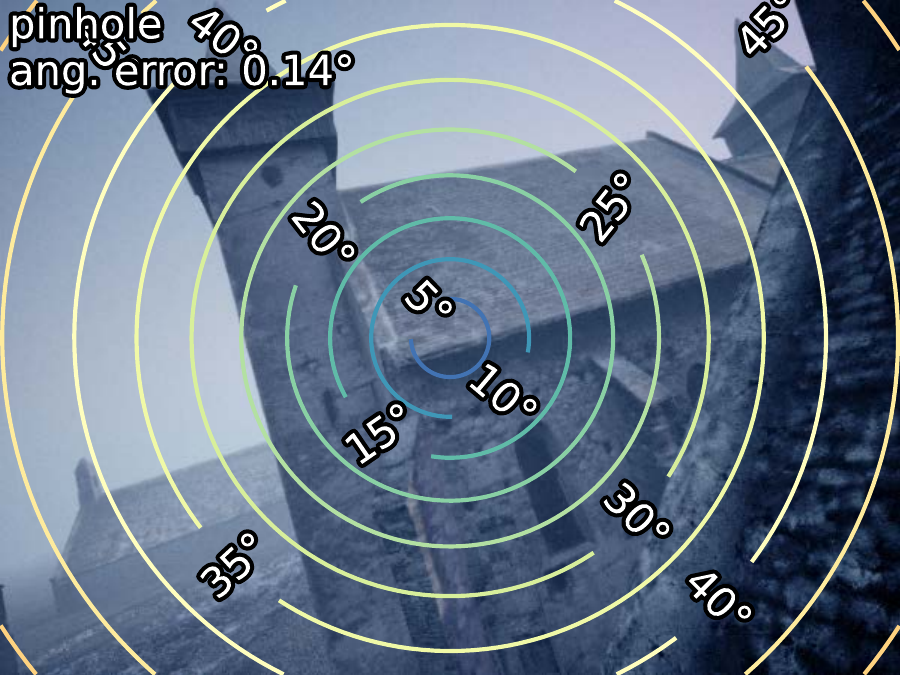} \\
        - & \code{radial:1} && 
        \includegraphics[height=\figh,valign=m]{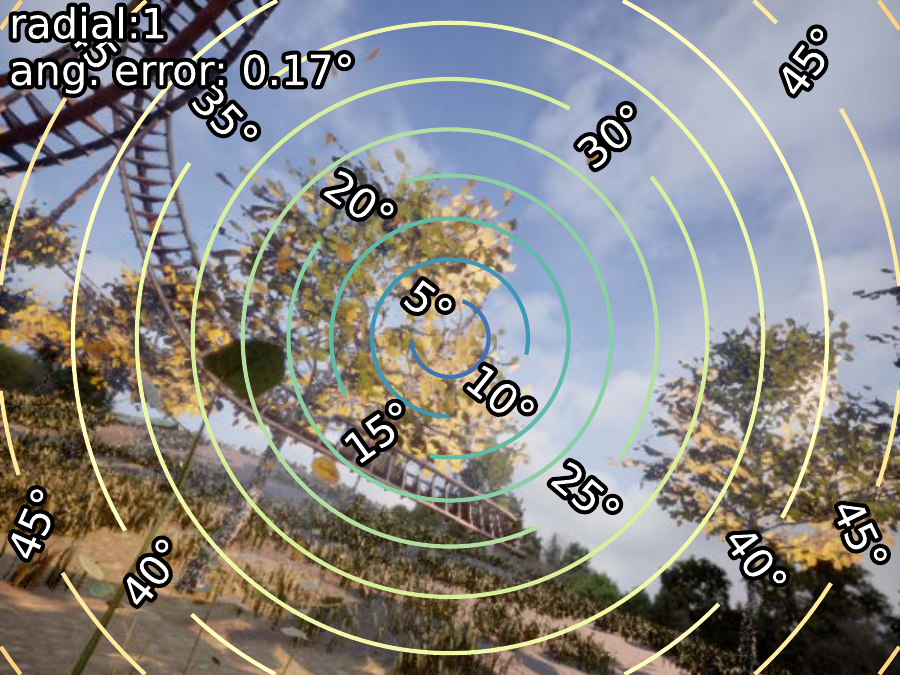} && 
        \includegraphics[height=\figh,valign=m]{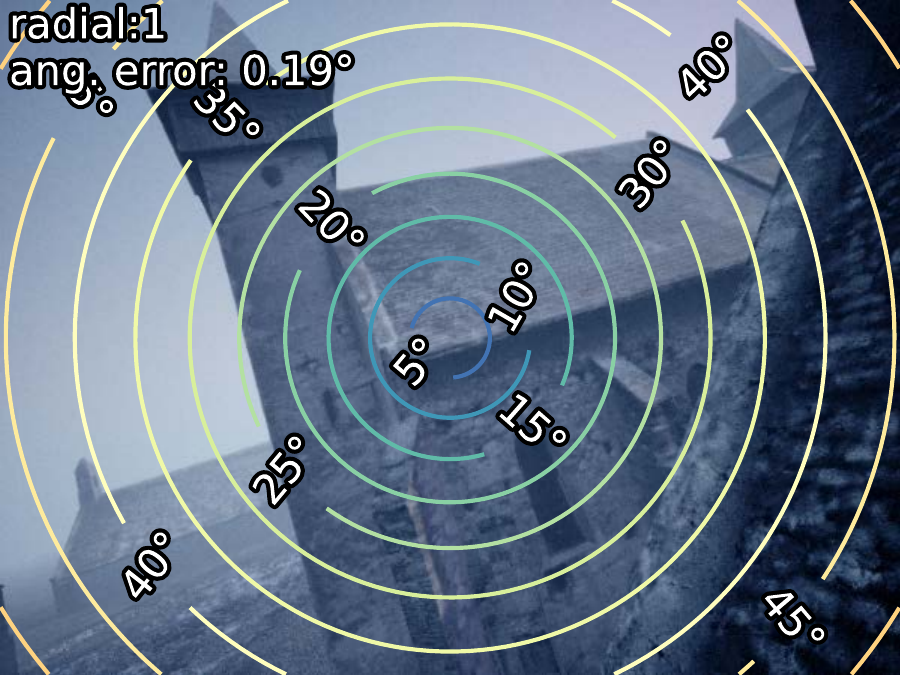} \\
        - & \code{radial:2} && 
        \includegraphics[height=\figh,valign=m]{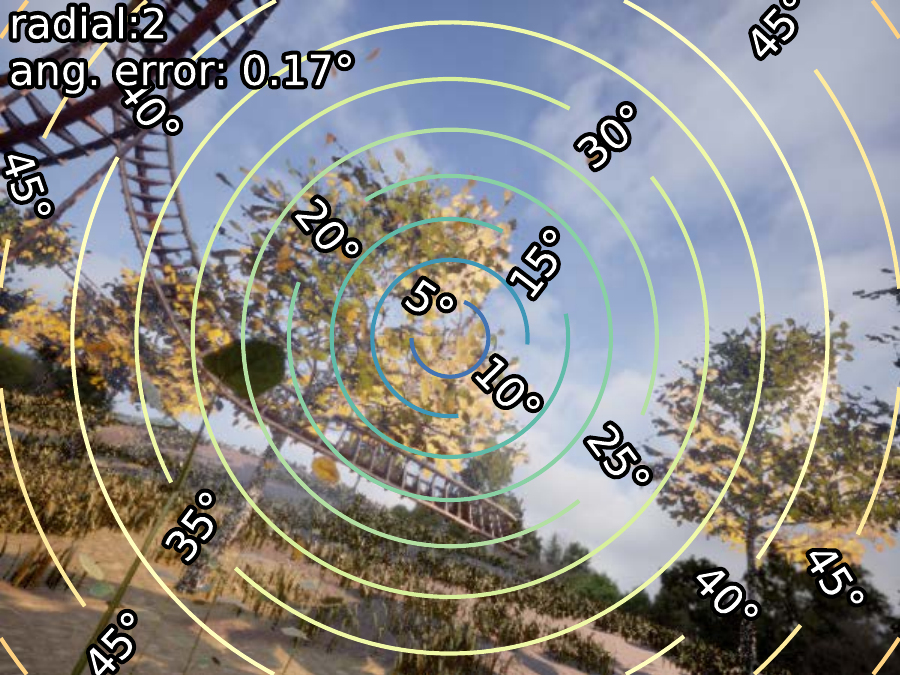} &&
        \includegraphics[height=\figh,valign=m]{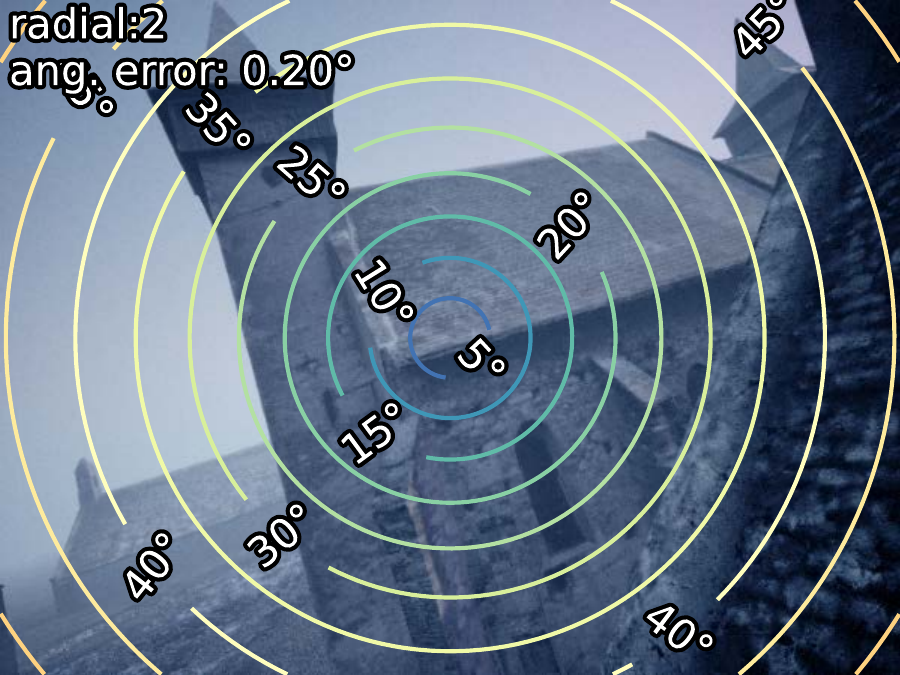} \\
    \end{tblr}
    \caption{\textbf{Qualitative results on perspective images} in TartanAir \cite{wang2020tartanair} with AnyCalib$_{\mathrm{gen}}$---trained on \datg. The FoV field ($\theta_x$ and $\theta_y$) is regressed by the network and $\lVert\boldsymbol\theta\rVert$ represents both its norm and the polar angle of the ray corresponding to each pixel. The predicted FoV field is used to fit the camera model of choice. \code{radial:i} corresponds to the Brown-Conrady model with \code{i} distortion coefficients.}
    \label{fig:supp_qual_tartan}
\end{figure*}

\begin{figure*}[t]

    \setlength{\figh}{2.8cm}

    \definecolor{codecolor}{RGB}{246, 248, 250}
    \newcommand{\code}[1]{\colorbox{codecolor}{\texttt{#1}}}

    \newcommand{\lfnamebase}{figures/supp_qualitative_perspective/149438231719}
    \newcommand{\rfnamebase}{figures/supp_qualitative_perspective/128613160990}
    \newcommand{\rrfnamebase}{figures/supp_qualitative_perspective/43582496299}

    \centering
    \begin{tblr}{
        width=1.0\linewidth,
        vspan=even,
        colspec={*{8}c},
        colsep=1pt,
        rowsep=1pt,
        cell{2}{1}={r=2}{c, cmd=\rotatebox{90}},
        cell{4}{1}={c, cmd={\rotatebox[origin=c]{90}}},
        cell{5,6,7}{2}={c, cmd={\rotatebox[origin=c]{90}}},
        cell{5}{1}={r=3}{c, cmd=\rotatebox{90}},
    }
        && ground-truth & prediction & ground-truth & prediction & ground-truth & prediction \\
        FoV field & $\theta_x$ & 
        \includegraphics[height=\figh,valign=m]{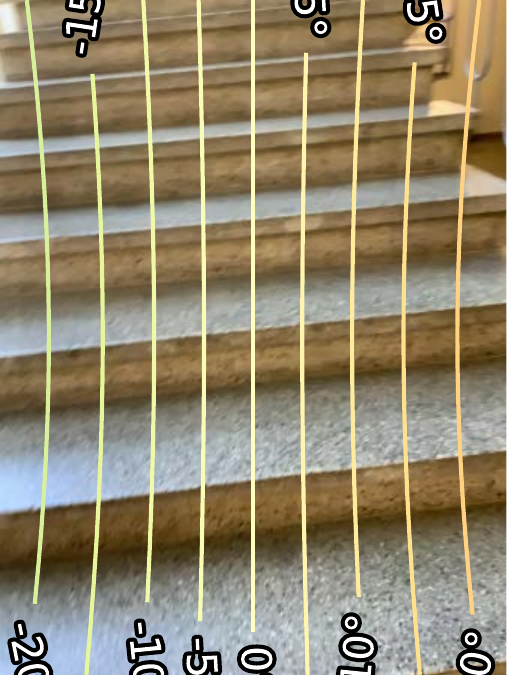} & 
        \includegraphics[height=\figh,valign=m]{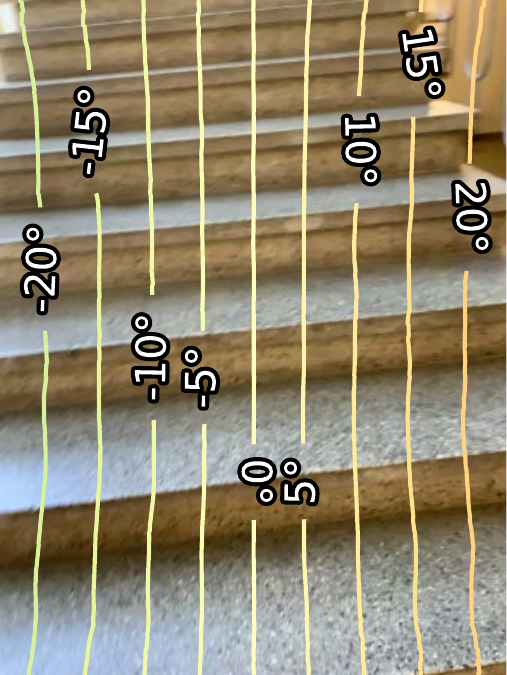} & 
        \includegraphics[height=\figh,valign=m]{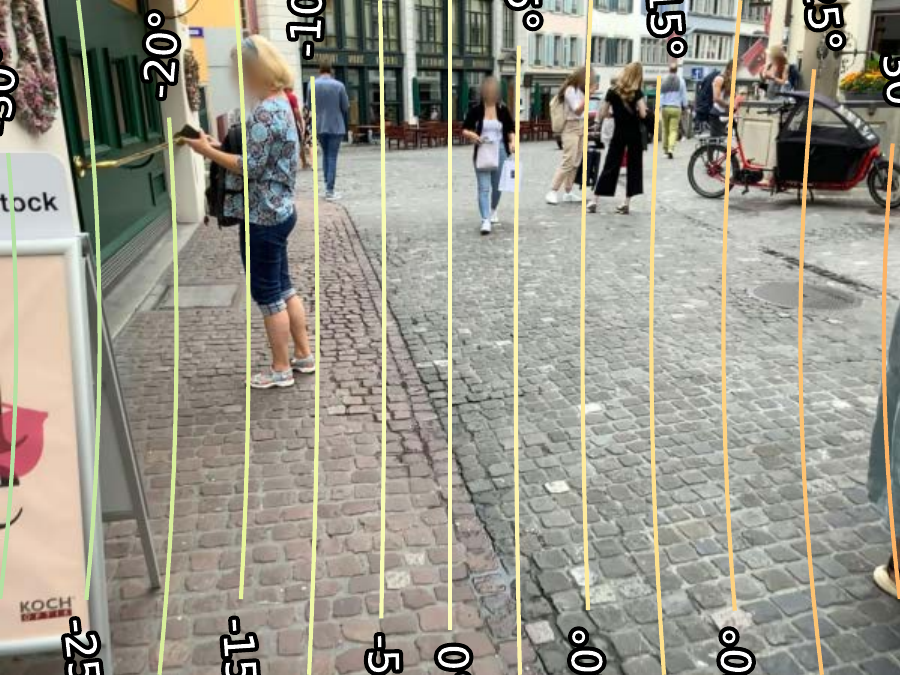} & 
        \includegraphics[height=\figh,valign=m]{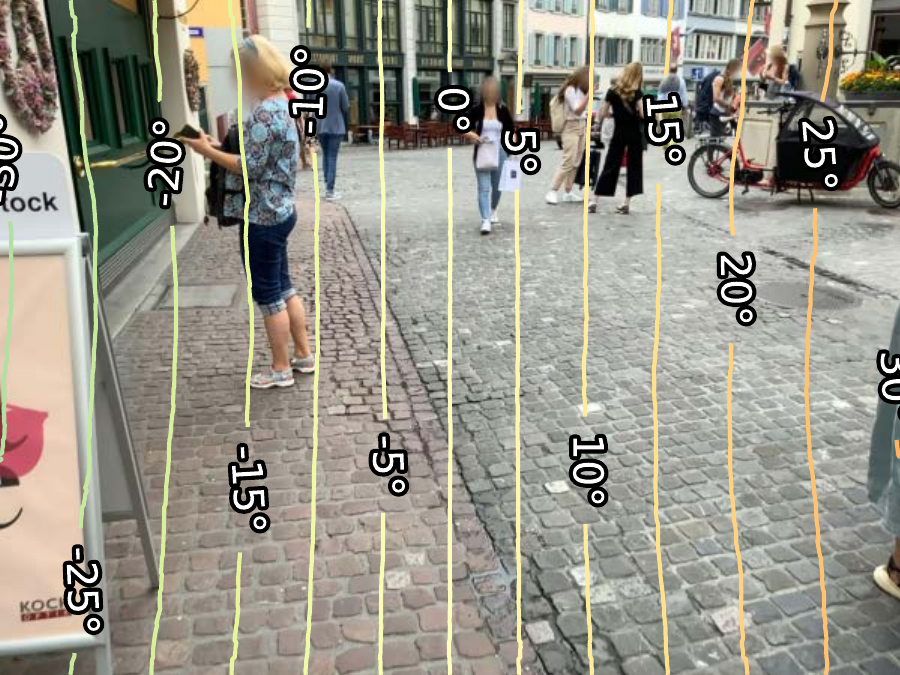} &
        \includegraphics[height=\figh,valign=m]{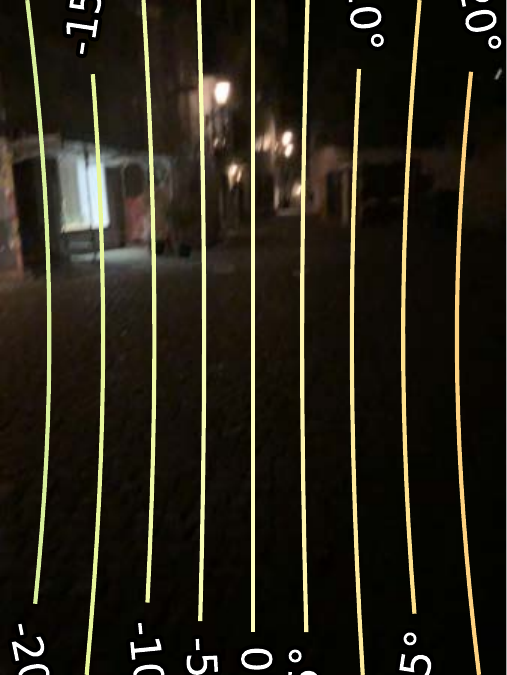} & 
        \includegraphics[height=\figh,valign=m]{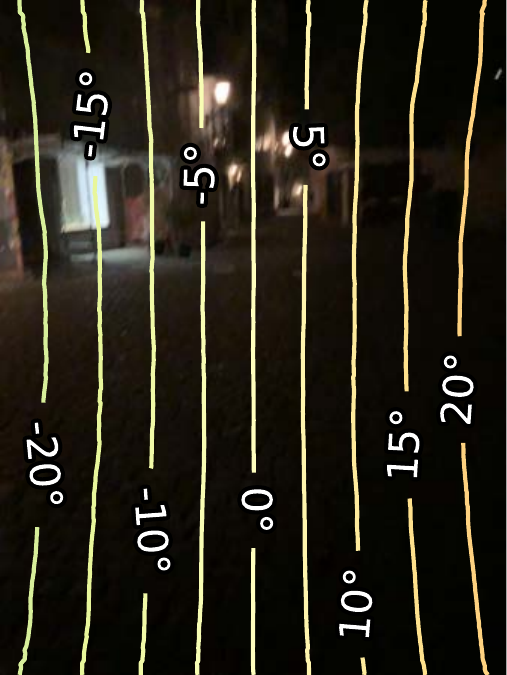} \\
        -         & $\theta_y$ & 
        \includegraphics[height=\figh,valign=m]{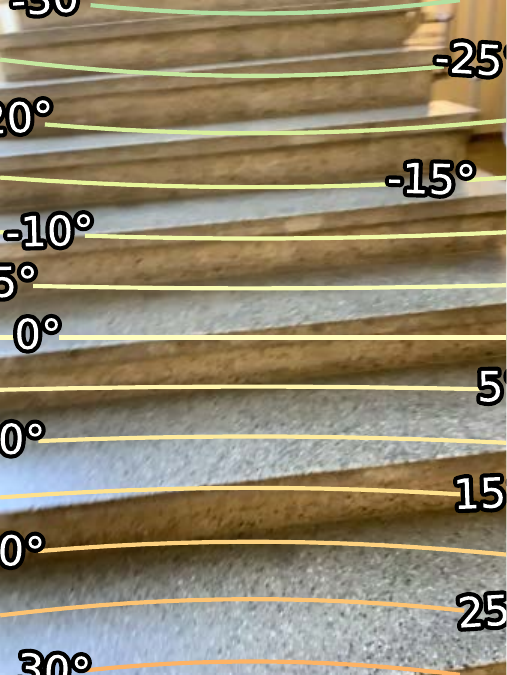} & 
        \includegraphics[height=\figh,valign=m]{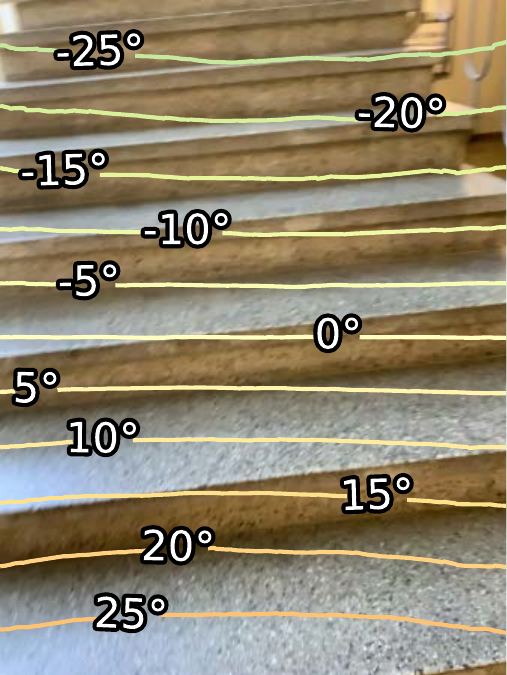} &
        \includegraphics[height=\figh,valign=m]{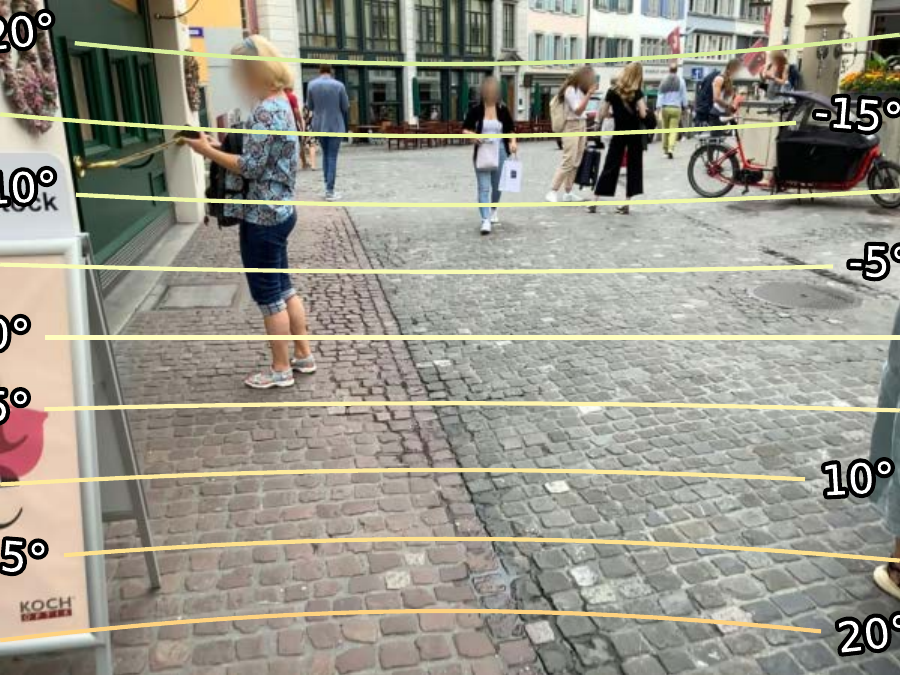} & 
        \includegraphics[height=\figh,valign=m]{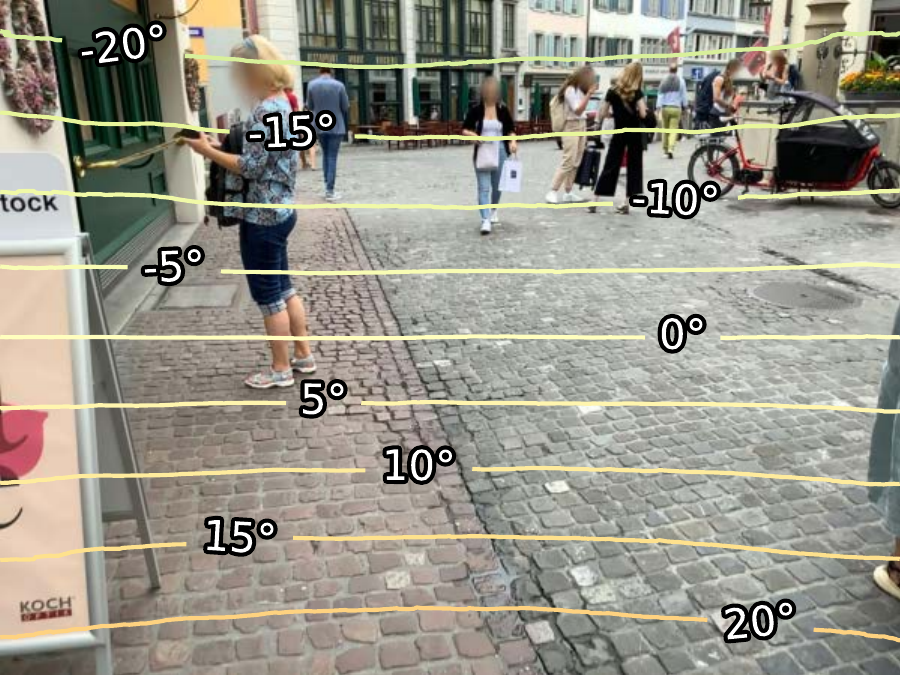} &
        \includegraphics[height=\figh,valign=m]{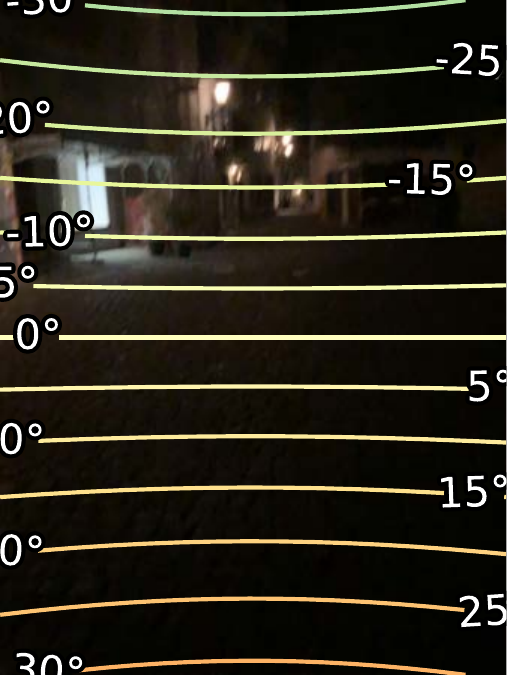} & 
        \includegraphics[height=\figh,valign=m]{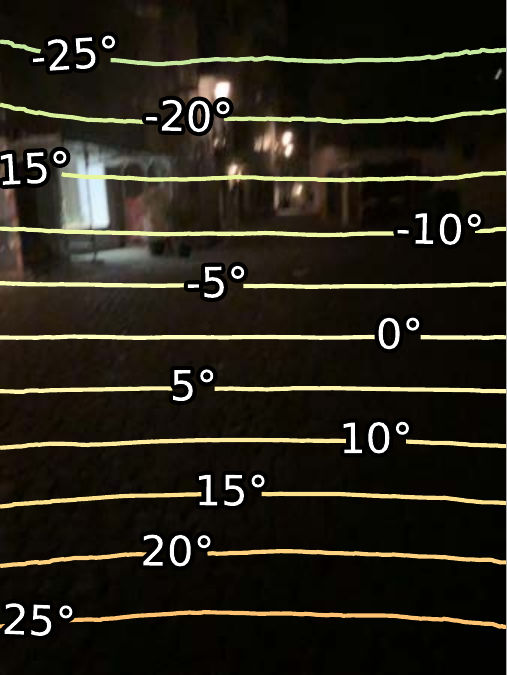} \\
        polar angles $\lVert\boldsymbol{\theta}\rVert$ & & 
        \includegraphics[height=\figh,valign=m]{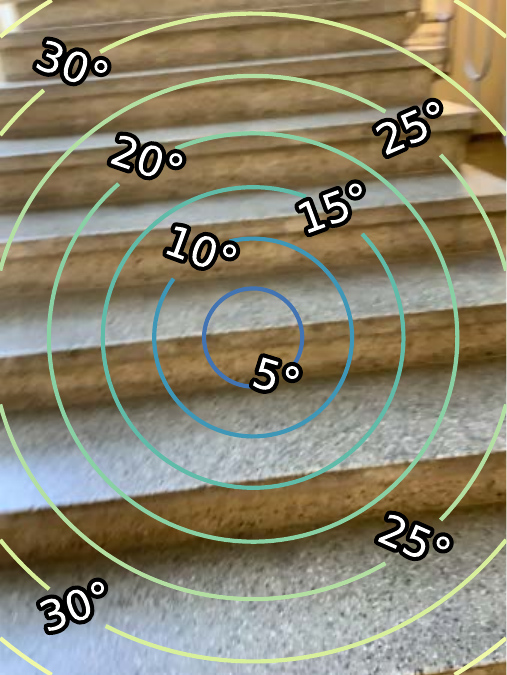} & 
        \includegraphics[height=\figh,valign=m]{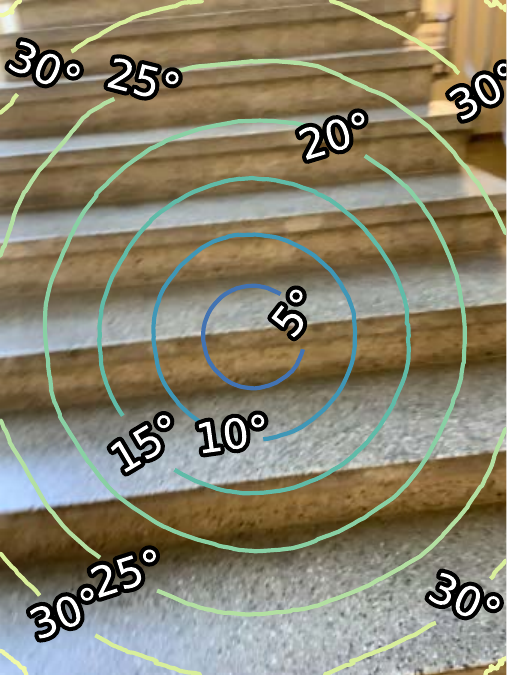} & 
        \includegraphics[height=\figh,valign=m]{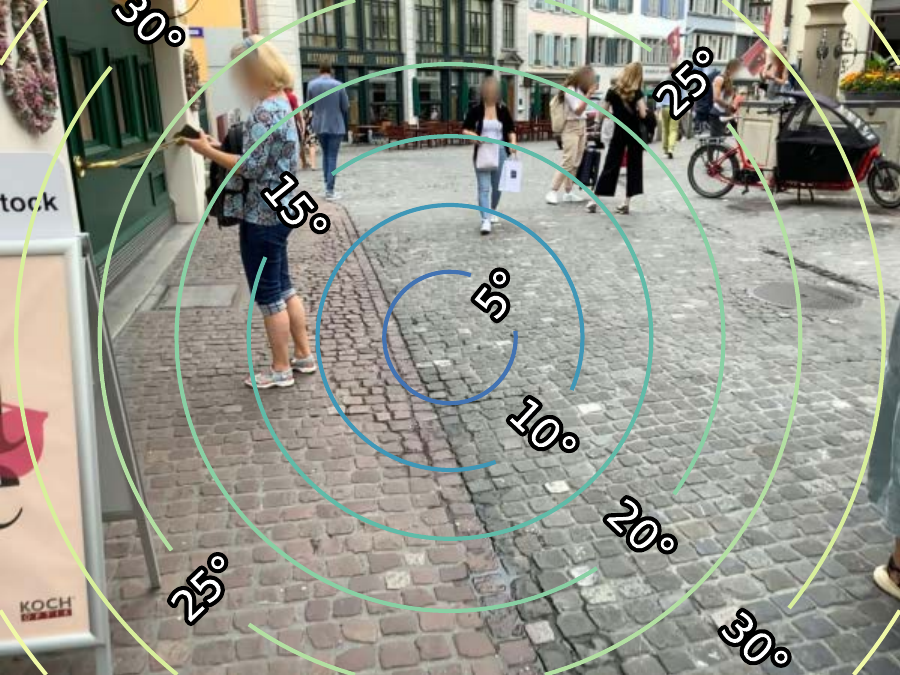} & 
        \includegraphics[height=\figh,valign=m]{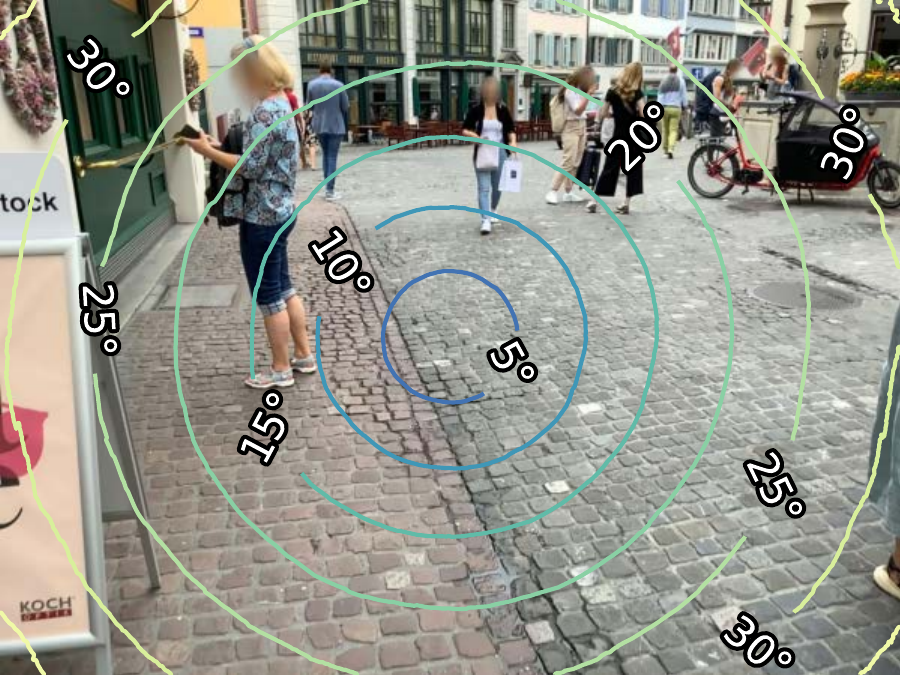} &
        \includegraphics[height=\figh,valign=m]{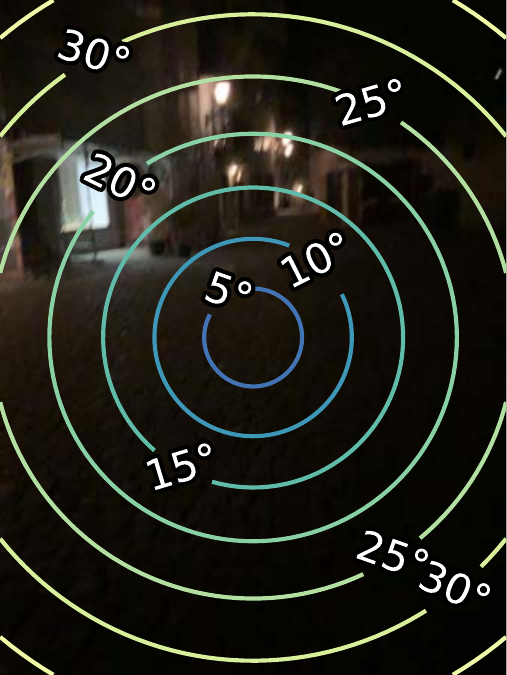} & 
        \includegraphics[height=\figh,valign=m]{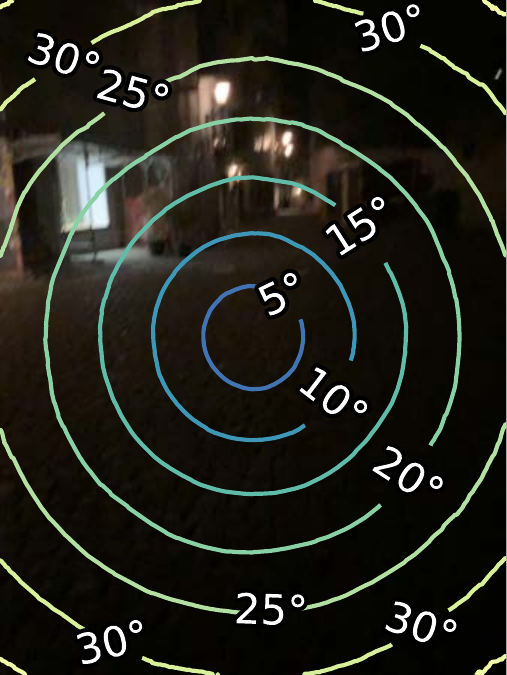} \\
        fits & \code{pinhole} & 
        \includegraphics[height=\figh,valign=m]{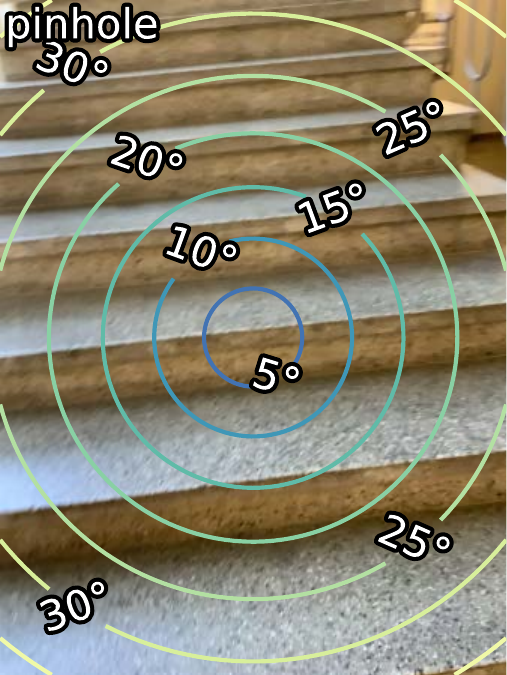} & 
        \includegraphics[height=\figh,valign=m]{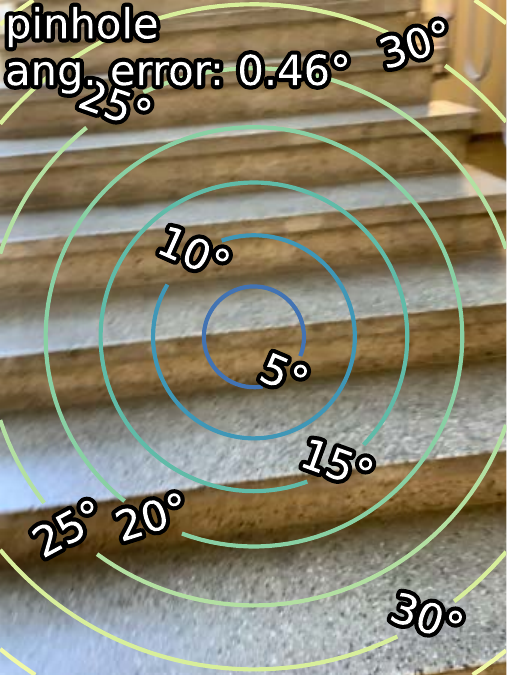} &
        \includegraphics[height=\figh,valign=m]{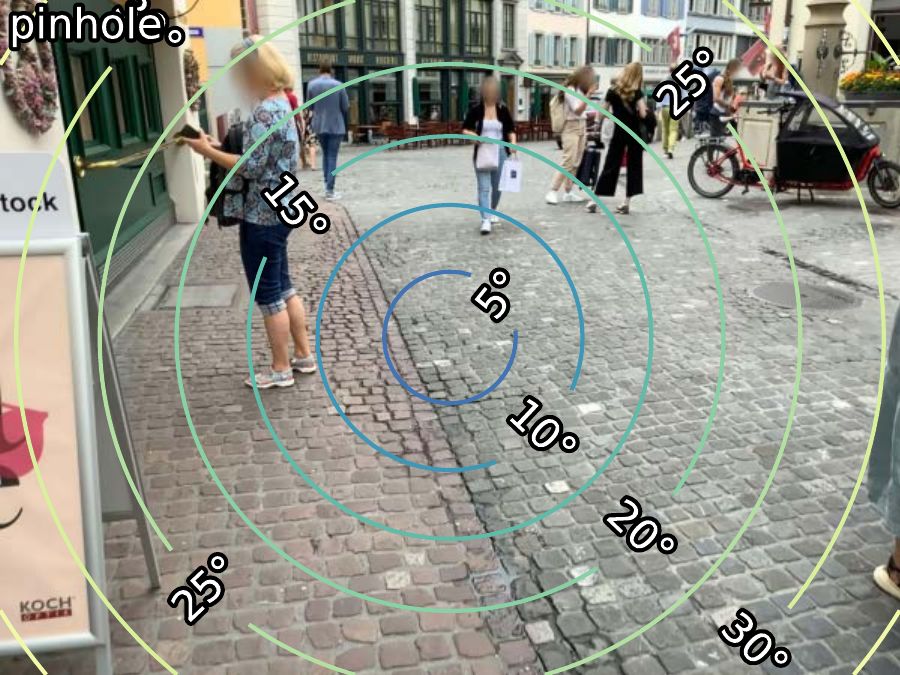} & 
        \includegraphics[height=\figh,valign=m]{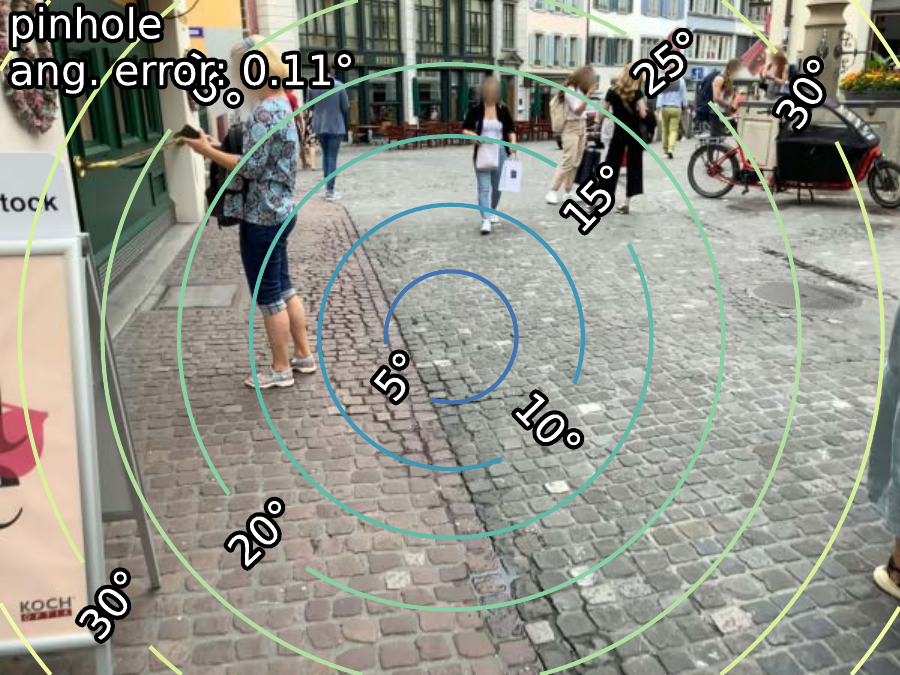} &
        \includegraphics[height=\figh,valign=m]{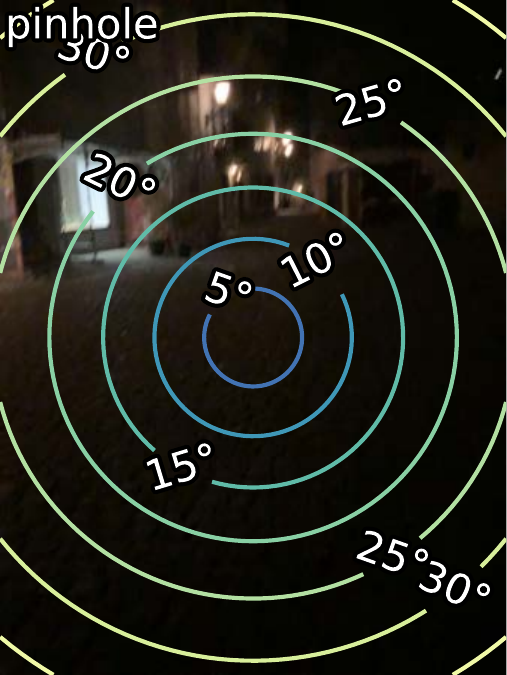} & 
        \includegraphics[height=\figh,valign=m]{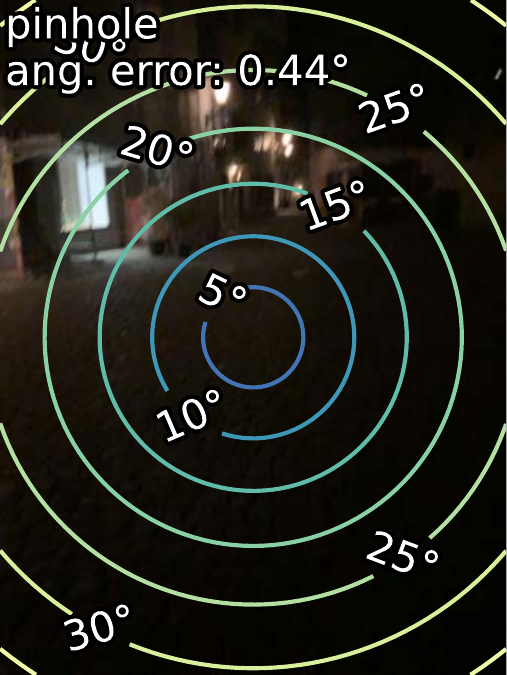} \\
        - & \code{ucm} && 
        \includegraphics[height=\figh,valign=m]{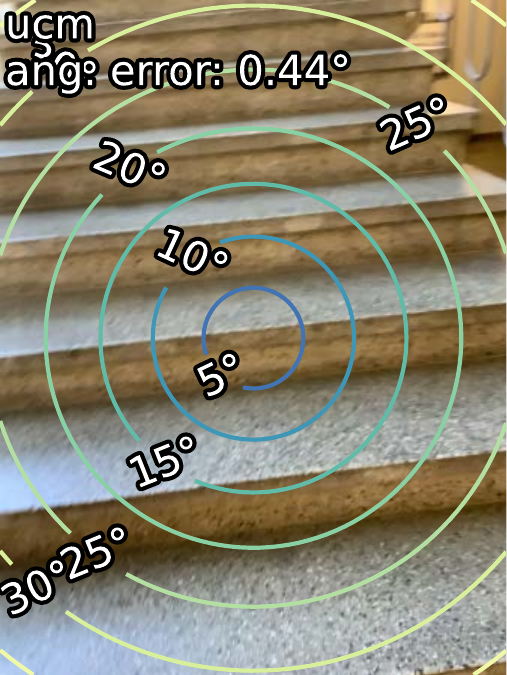} && 
        \includegraphics[height=\figh,valign=m]{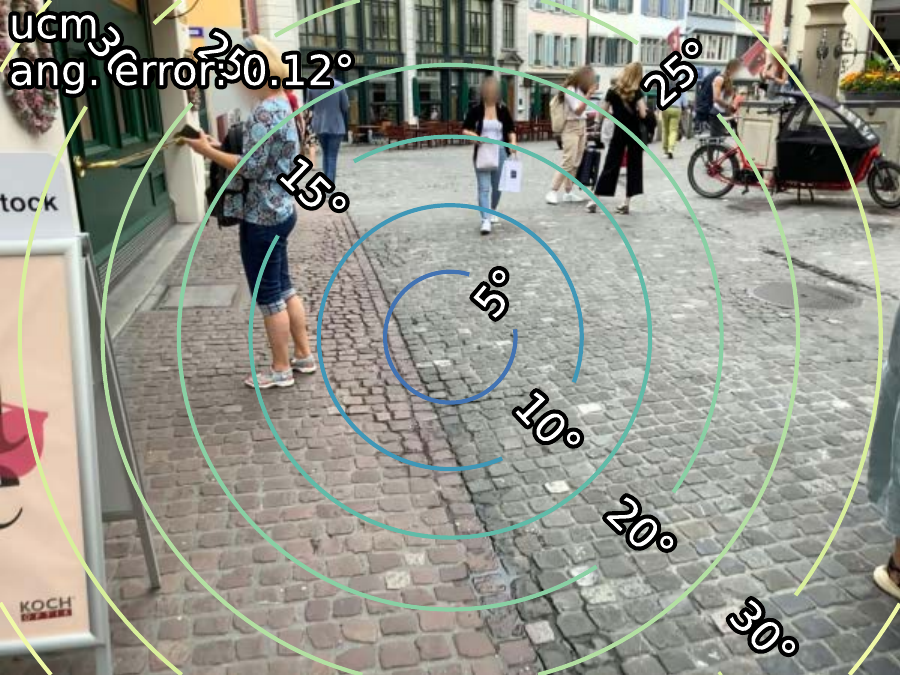} &&
        \includegraphics[height=\figh,valign=m]{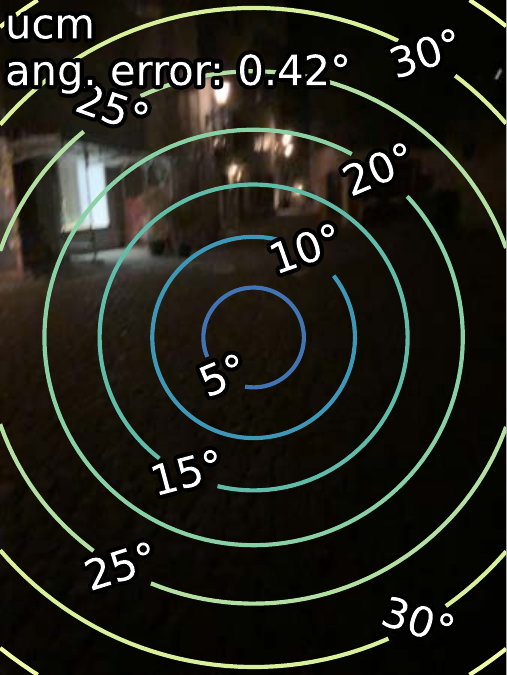} \\
        - & \code{eucm} && 
        \includegraphics[height=\figh,valign=m]{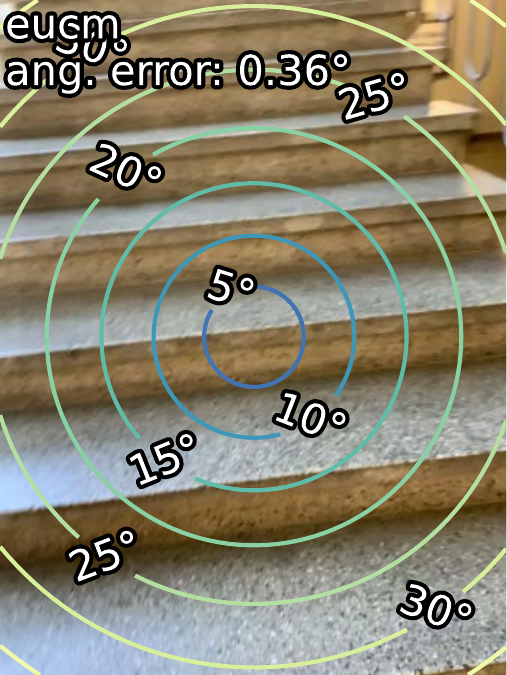} &&
        \includegraphics[height=\figh,valign=m]{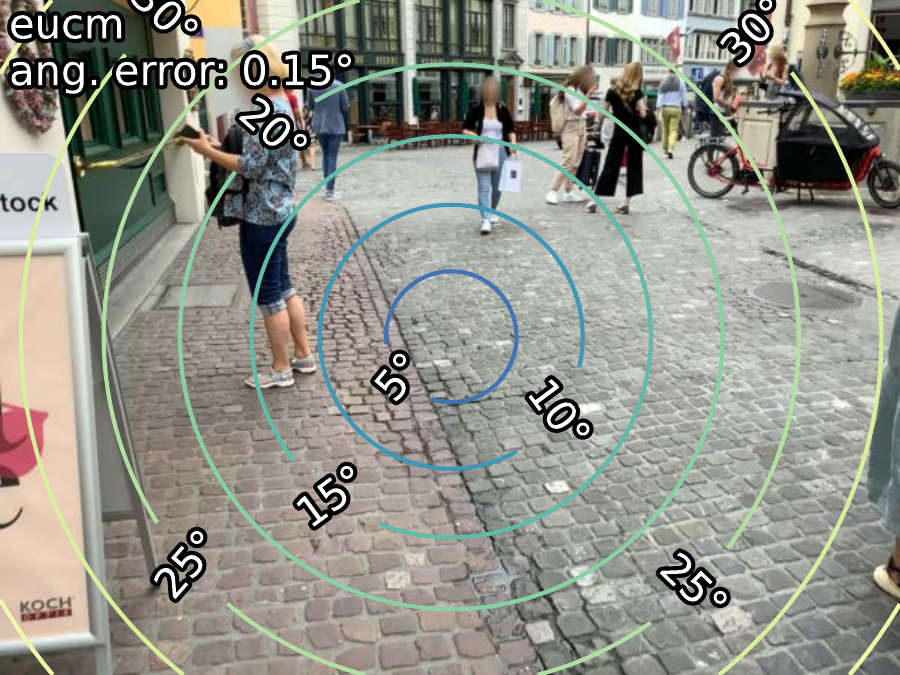} &&
        \includegraphics[height=\figh,valign=m]{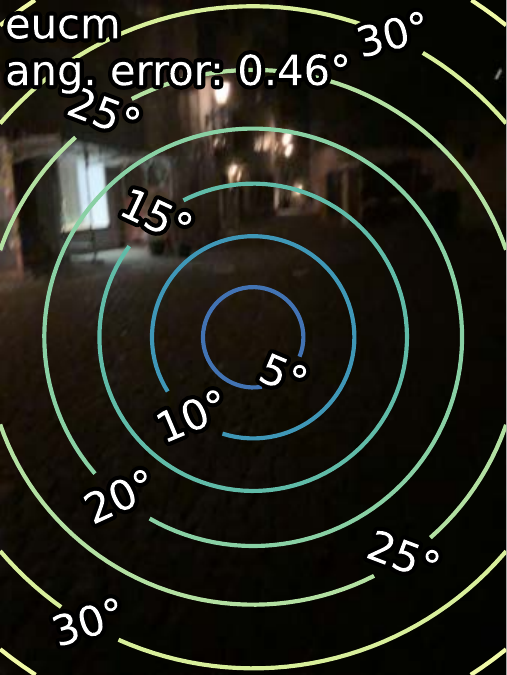} \\
    \end{tblr}
    \caption{\textbf{Qualitative results on perspective images} in LaMAR \cite{sarlin2022lamar} with AnyCalib$_{\mathrm{gen}}$---trained on \datg. The FoV field ($\theta_x$ and $\theta_y$) is regressed by the network and $\lVert\boldsymbol\theta\rVert$ represents both its norm and the polar angle of the ray corresponding to each pixel. The predicted FoV field is used to fit the camera model of choice.}
    \label{fig:supp_qual_lamar}
\end{figure*}

\begin{figure*}[t]

    \setlength{\figh}{2.8cm}

    \definecolor{codecolor}{RGB}{246, 248, 250}
    \newcommand{\code}[1]{\colorbox{codecolor}{\texttt{#1}}}

    \newcommand{\lfnamebase}{figures/supp_qualitative_perspective/374386609_5a71b27a86_o}
    \newcommand{\rfnamebase}{figures/supp_qualitative_perspective/3559880957_3cc52ffa0c_o}

    \centering
    \begin{tblr}{
        width=1.0\linewidth,
        vspan=even,
        colspec={*{6}c},
        colsep=1pt,
        rowsep=1pt,
        cell{2}{1}={r=2}{c, cmd=\rotatebox{90}},
        cell{4}{1}={c, cmd={\rotatebox[origin=c]{90}}},
        cell{5,6,7}{2}={c, cmd={\rotatebox[origin=c]{90}}},
        cell{5}{1}={r=3}{c, cmd=\rotatebox{90}},
    }
        && ground-truth & prediction & ground-truth & prediction \\
        FoV field & $\theta_x$ & 
        \includegraphics[height=\figh,valign=m]{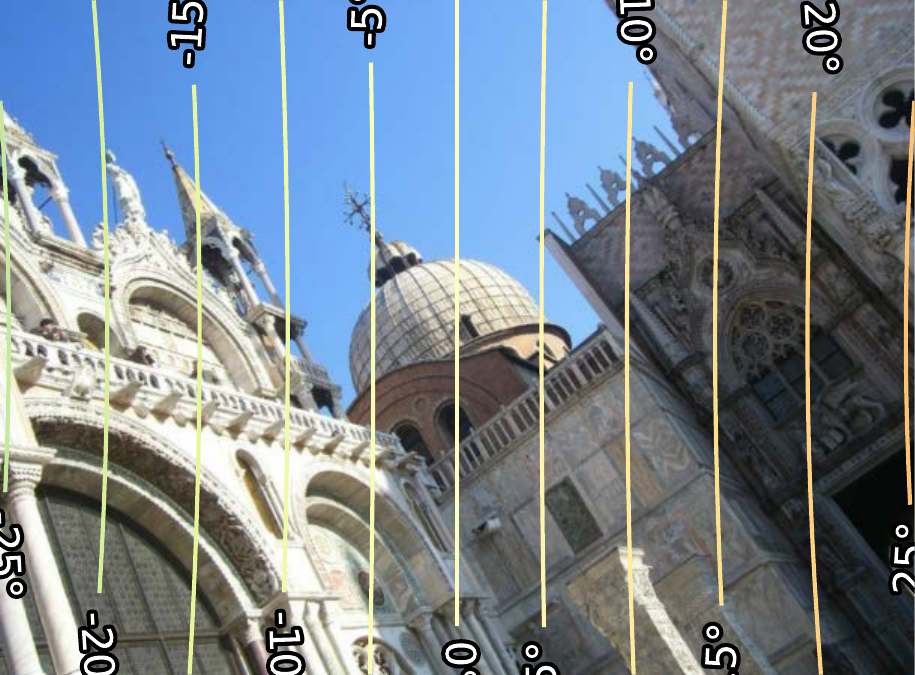} & 
        \includegraphics[height=\figh,valign=m]{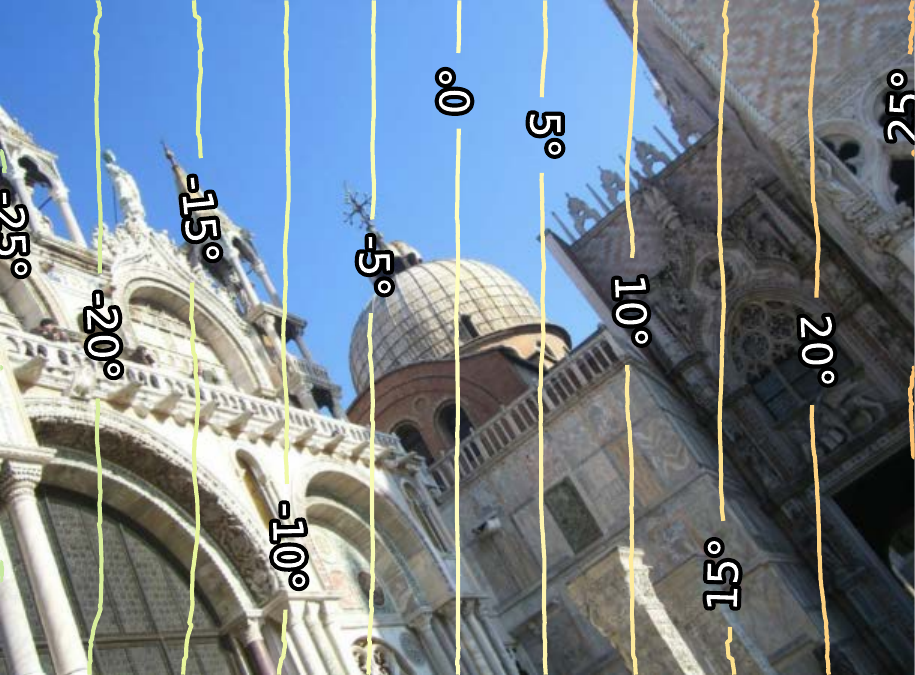} & 
        \includegraphics[height=\figh,valign=m]{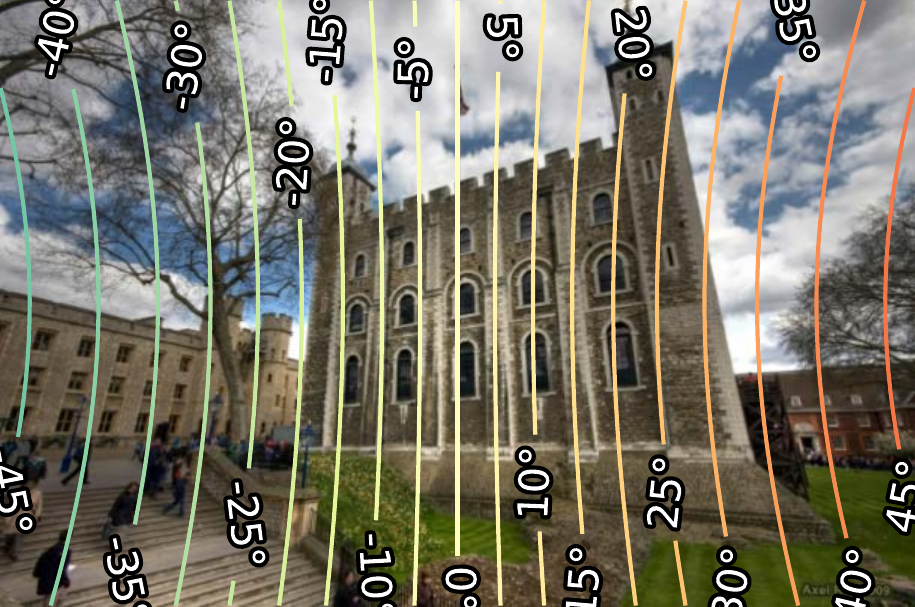} & 
        \includegraphics[height=\figh,valign=m]{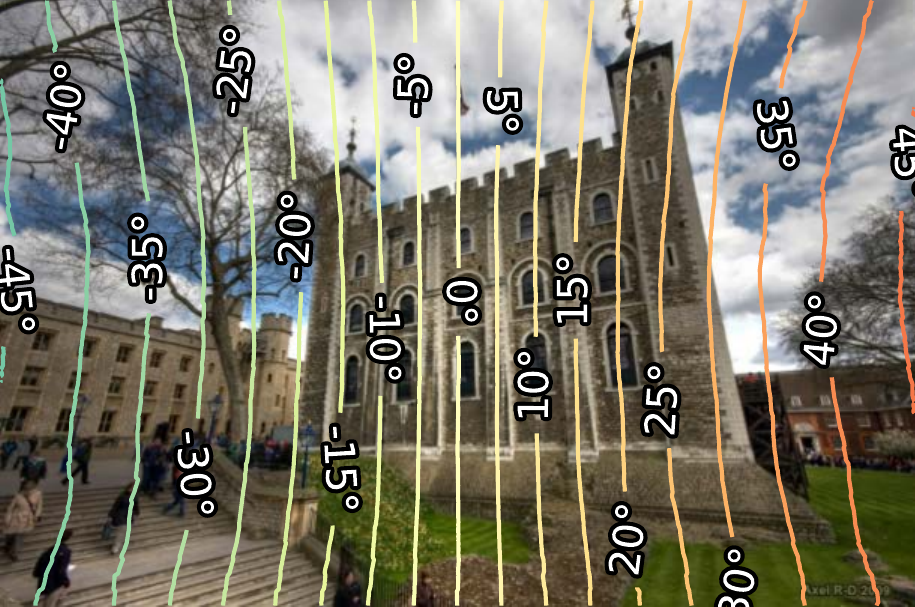} \\
        -         & $\theta_y$ & 
        \includegraphics[height=\figh,valign=m]{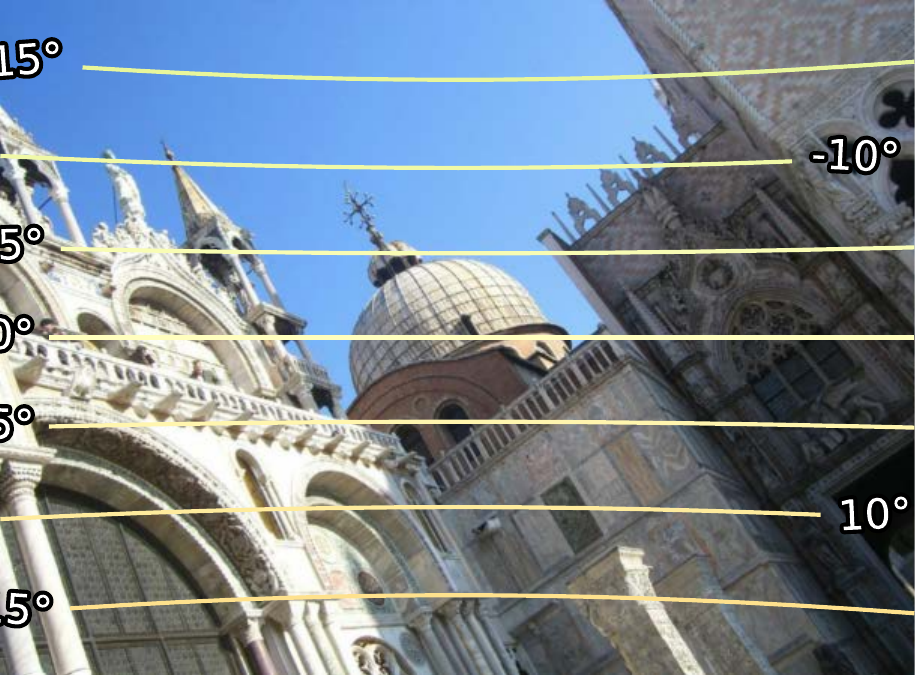} & 
        \includegraphics[height=\figh,valign=m]{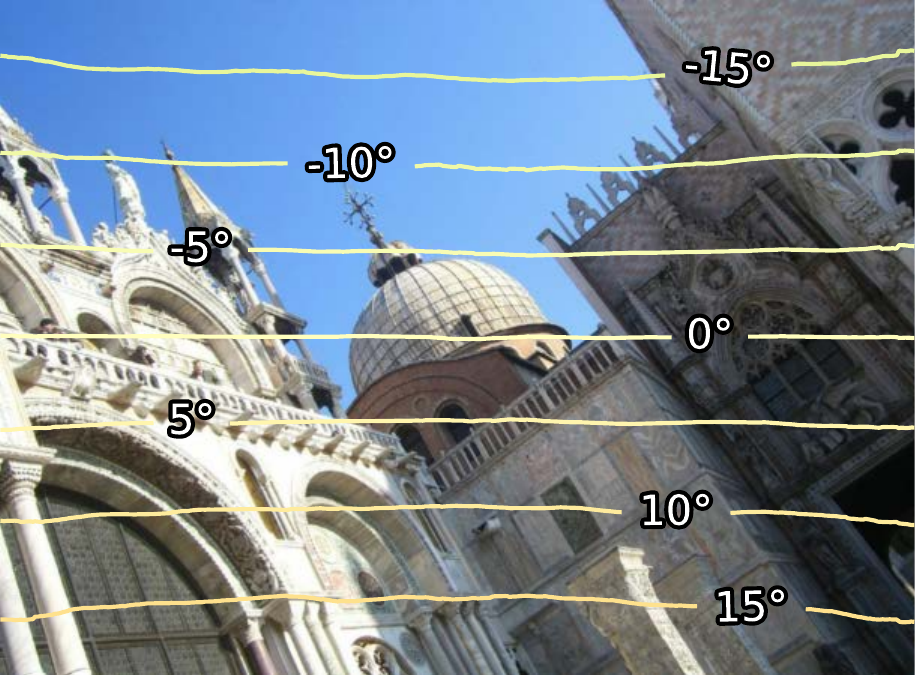} &
        \includegraphics[height=\figh,valign=m]{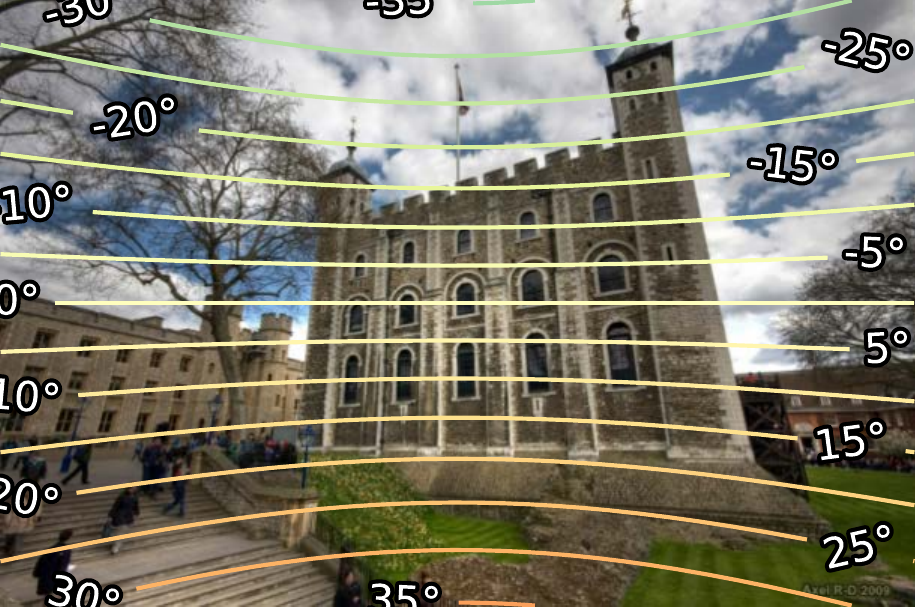} & 
        \includegraphics[height=\figh,valign=m]{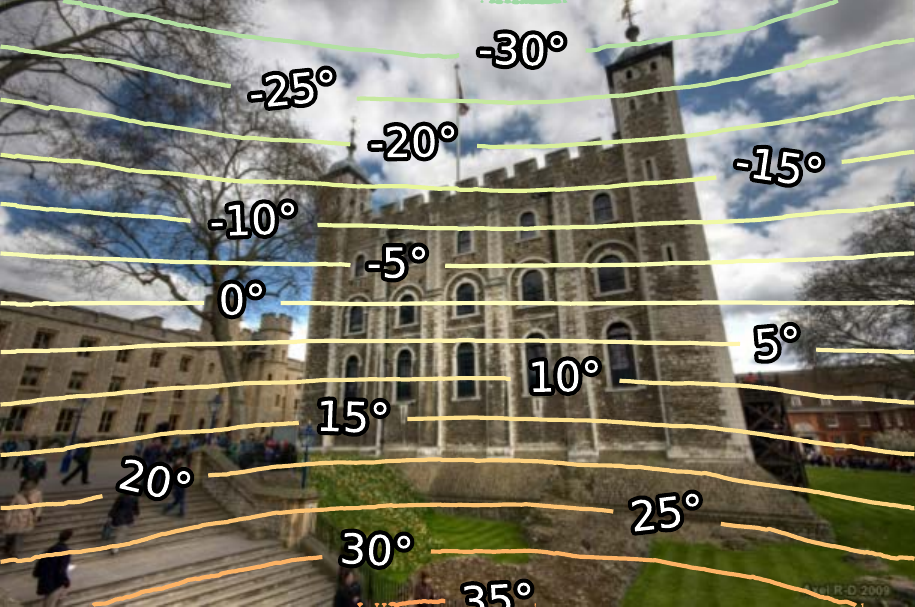} \\
        polar angles $\lVert\boldsymbol{\theta}\rVert$ & & 
        \includegraphics[height=\figh,valign=m]{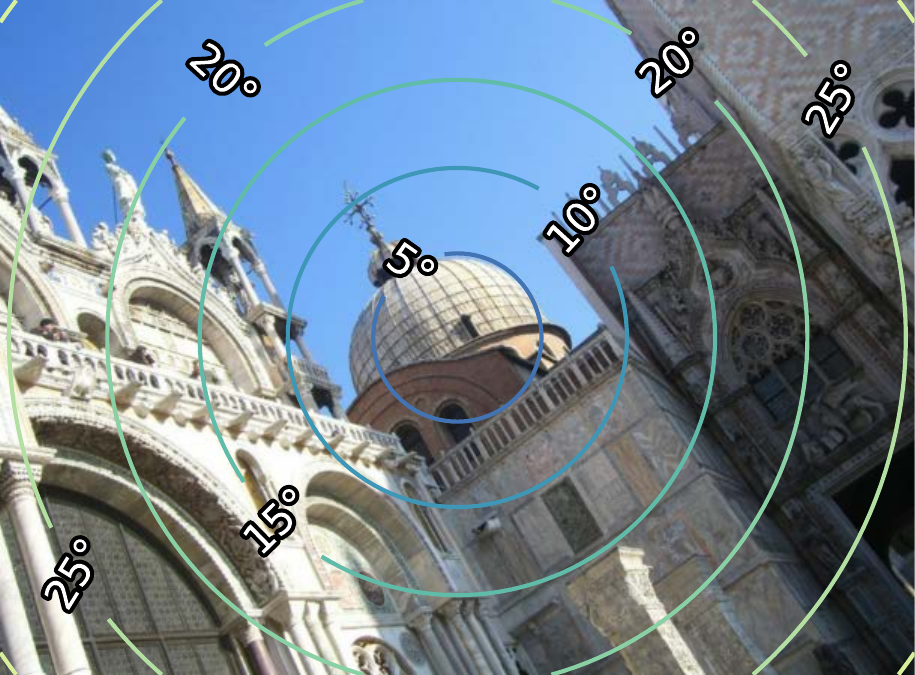} & 
        \includegraphics[height=\figh,valign=m]{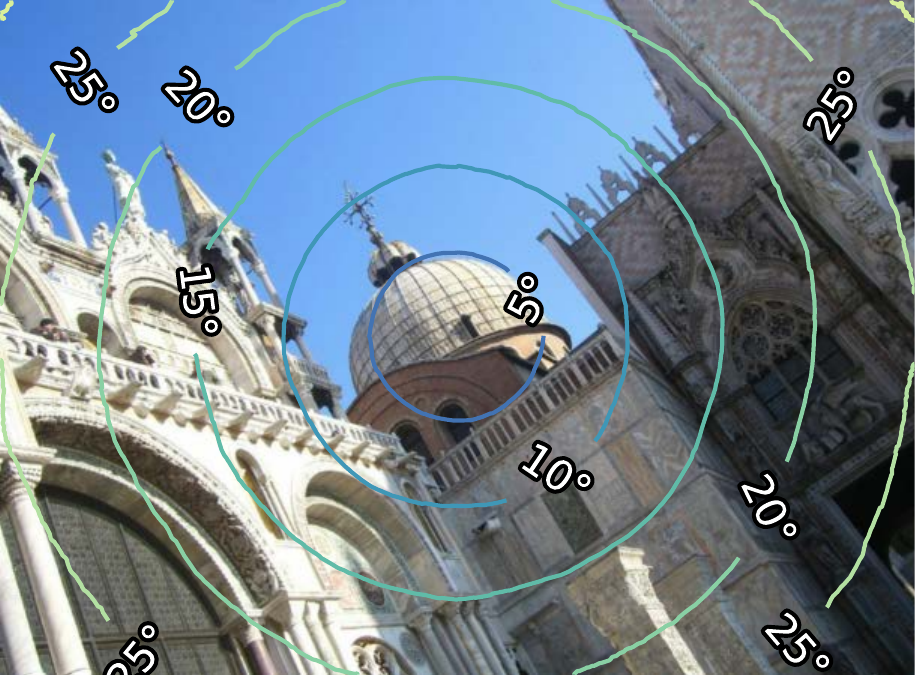} & 
        \includegraphics[height=\figh,valign=m]{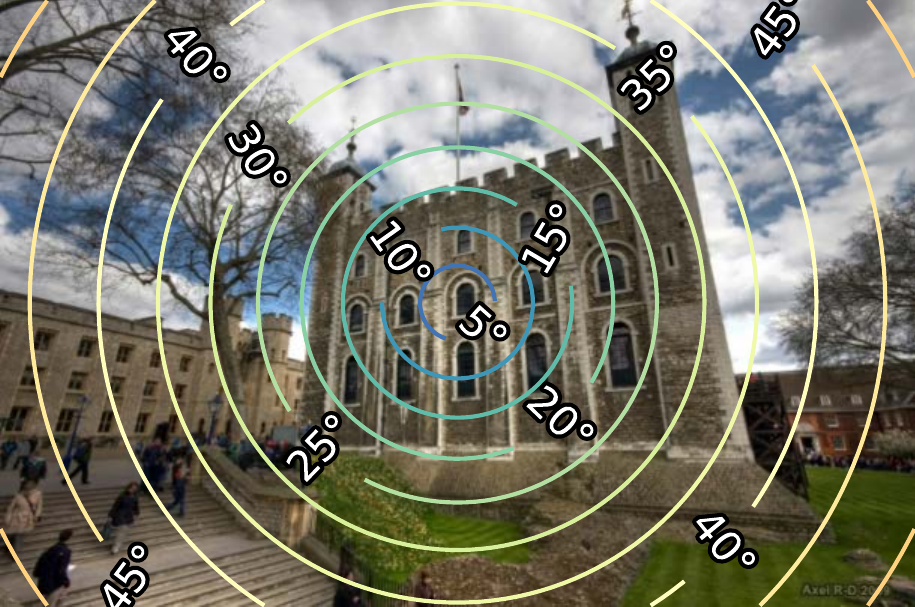} & 
        \includegraphics[height=\figh,valign=m]{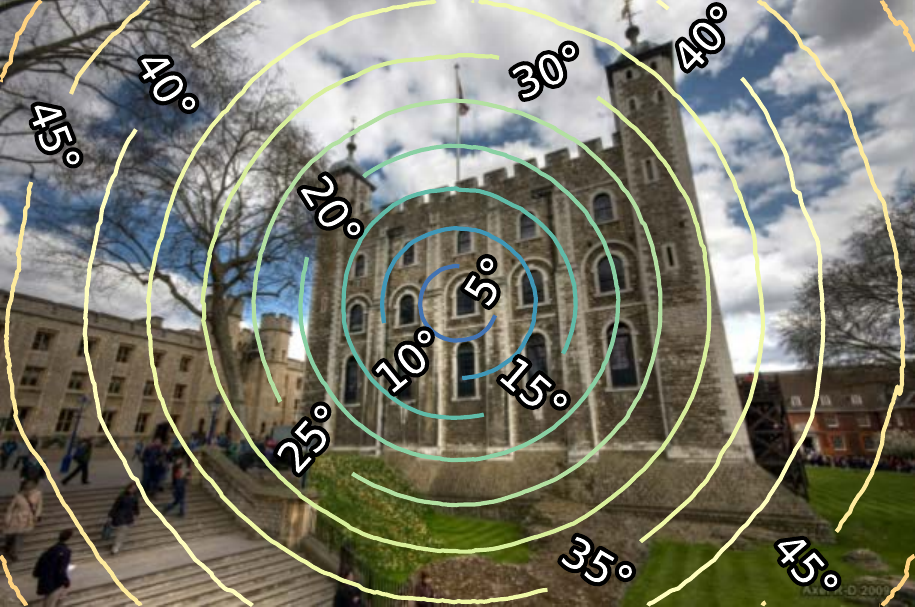} \\
        fits & \code{pinhole} & 
        \includegraphics[height=\figh,valign=m]{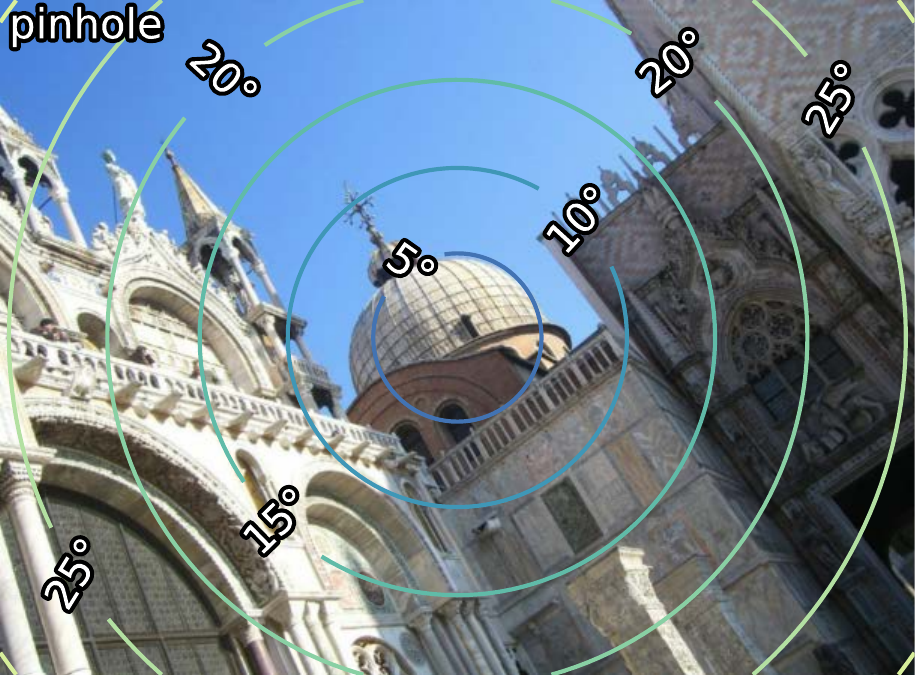} & 
        \includegraphics[height=\figh,valign=m]{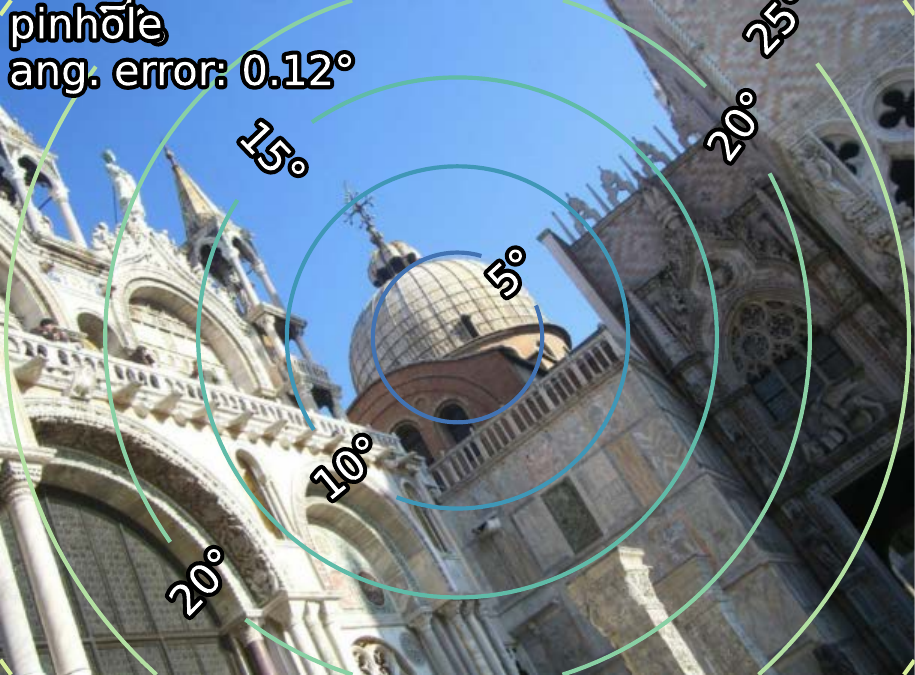} &
        \includegraphics[height=\figh,valign=m]{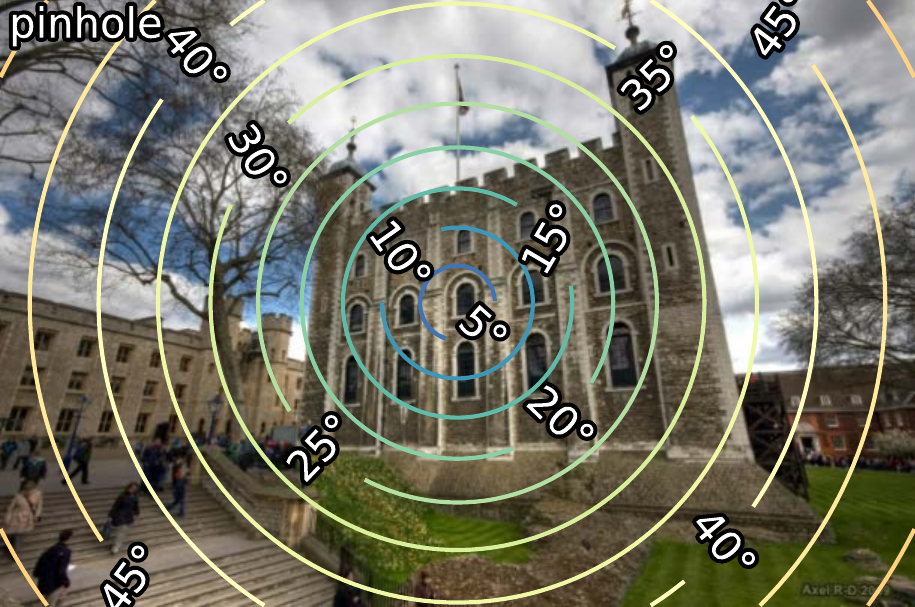} & 
        \includegraphics[height=\figh,valign=m]{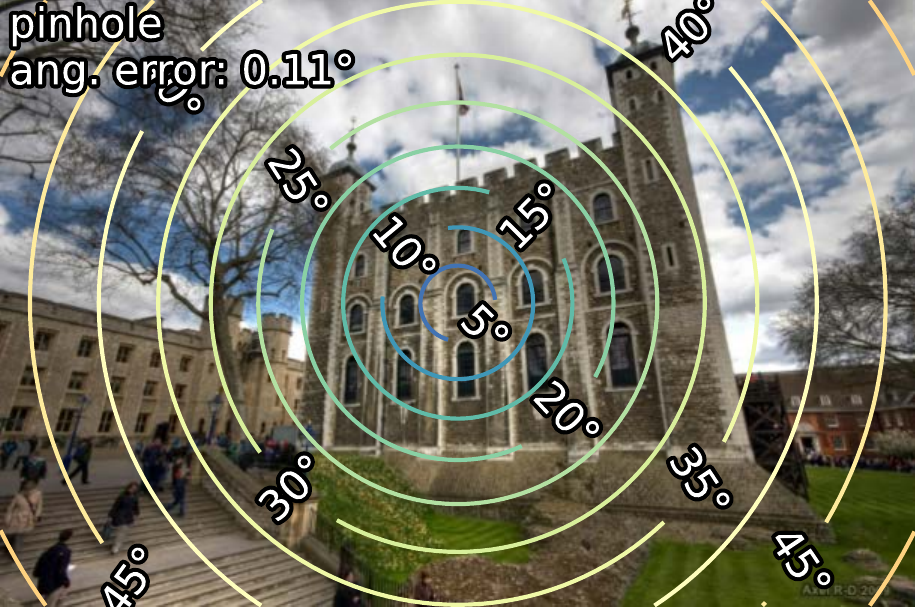} \\
        - & \code{kb:1} && 
        \includegraphics[height=\figh,valign=m]{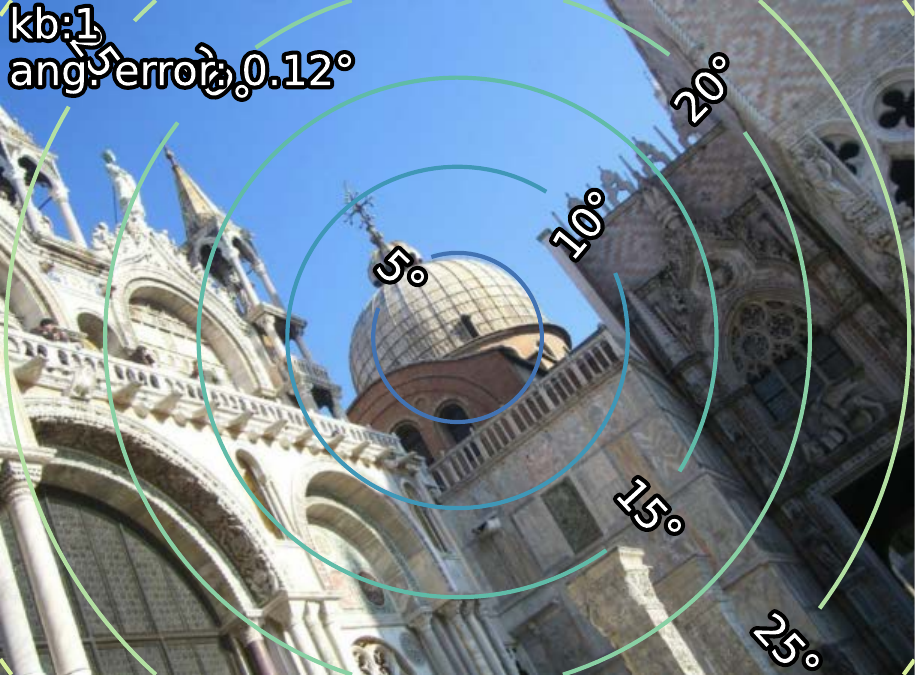} && 
        \includegraphics[height=\figh,valign=m]{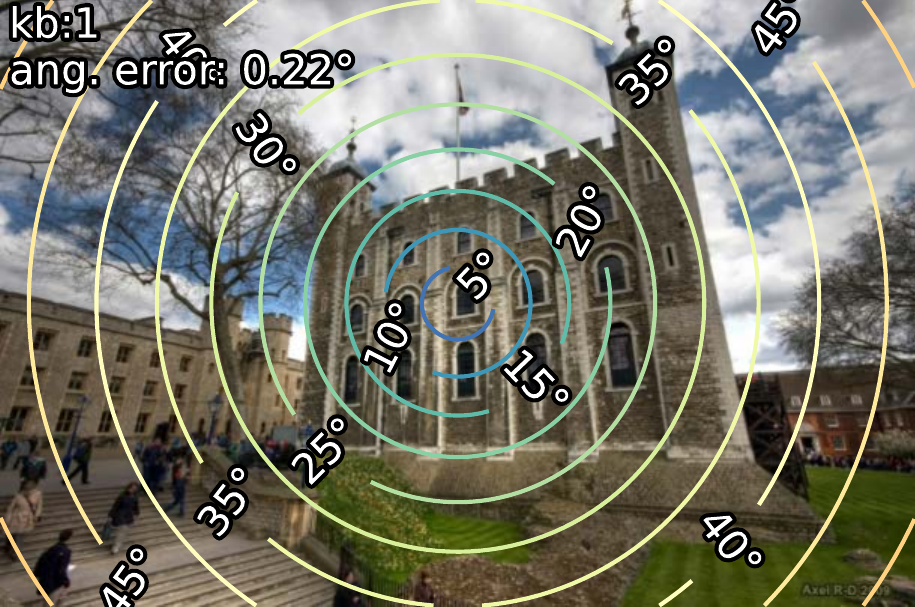} \\
        - & \code{kb:2} && 
        \includegraphics[height=\figh,valign=m]{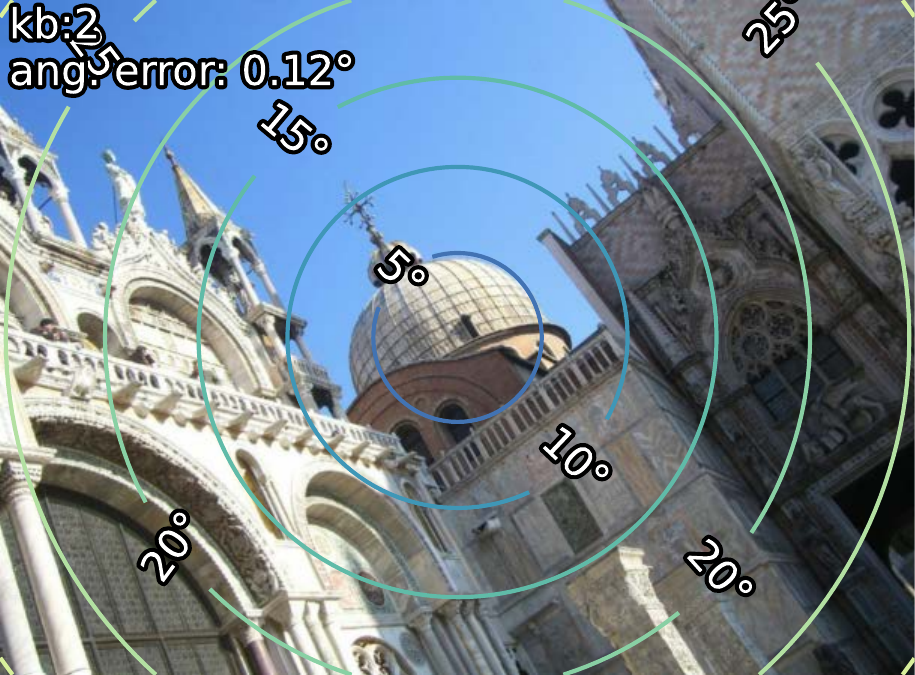} &&
        \includegraphics[height=\figh,valign=m]{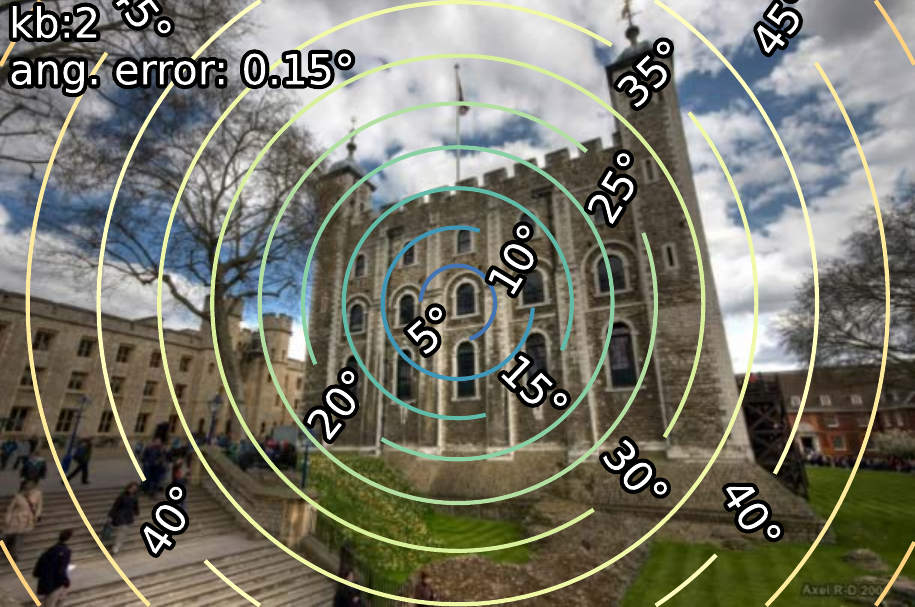} \\
    \end{tblr}
    \caption{\textbf{Qualitative results on perspective images} in MegaDepth \cite{Li2018megadepth} with AnyCalib$_{\mathrm{gen}}$---trained on \datg. The FoV field ($\theta_x$ and $\theta_y$) is regressed by the network and $\lVert\boldsymbol\theta\rVert$ represents both its norm and the polar angle of the ray corresponding to each pixel. The predicted FoV field is used to fit the camera model of choice. \code{kb:i} corresponds to the Kannala-Brandt model with \code{i} distortion coefficients.}
    \label{fig:supp_qual_mega}
\end{figure*}

\begin{figure*}[t]

    \setlength{\figh}{2.7cm}

    \definecolor{codecolor}{RGB}{246, 248, 250}
    \newcommand{\code}[1]{\colorbox{codecolor}{\texttt{#1}}}

    \newcommand{\lfnamebase}{figures/supp_qualitative_distorted/5fb5d2dbf2_DSC00700}
    \newcommand{\rfnamebase}{figures/supp_qualitative_distorted/b08a908f0f_DSC02528}

    \centering
    \begin{tblr}{
        width=1.0\linewidth,
        vspan=even,
        colspec={*{6}c},
        colsep=1pt,
        rowsep=1pt,
        cell{2}{1}={r=2}{c, cmd=\rotatebox{90}},
        cell{4}{1}={c, cmd={\rotatebox[origin=c]{90}}},
        cell{5,6,7}{2}={c, cmd={\rotatebox[origin=c]{90}}},
        cell{5}{1}={r=3}{c, cmd=\rotatebox{90}},
    }
        && ground-truth & prediction & ground-truth & prediction \\
        FoV field & $\theta_x$ & 
        \includegraphics[height=\figh,valign=m]{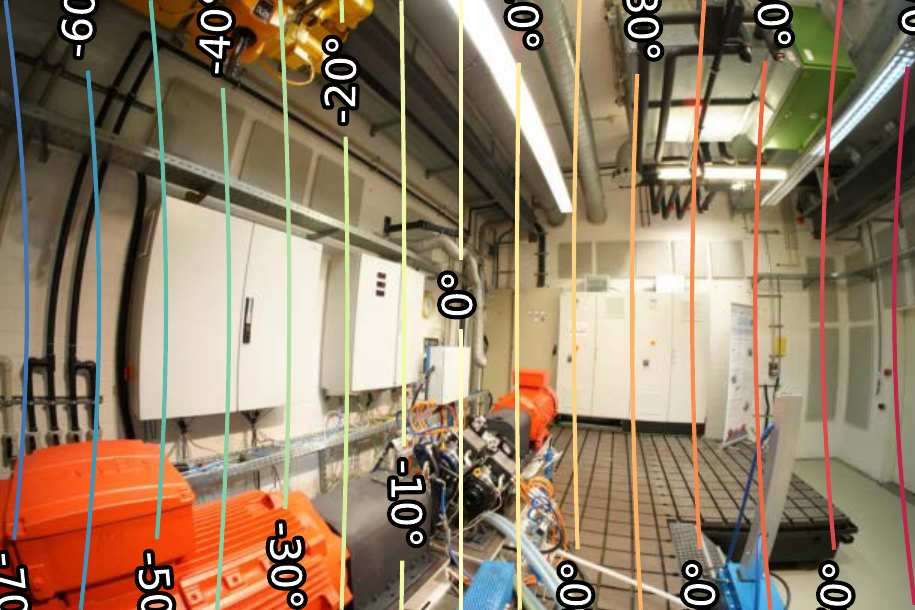} & 
        \includegraphics[height=\figh,valign=m]{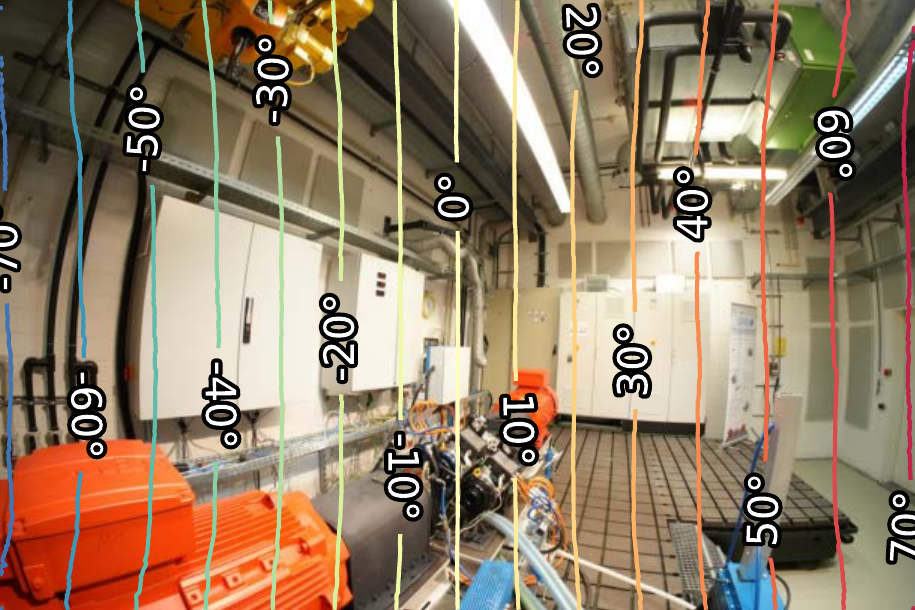} & 
        \includegraphics[height=\figh,valign=m]{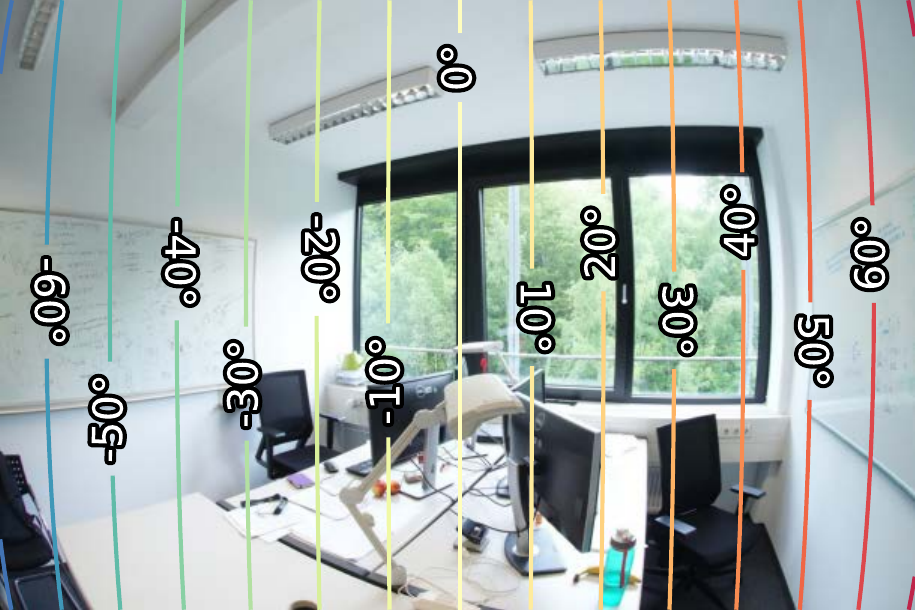} & 
        \includegraphics[height=\figh,valign=m]{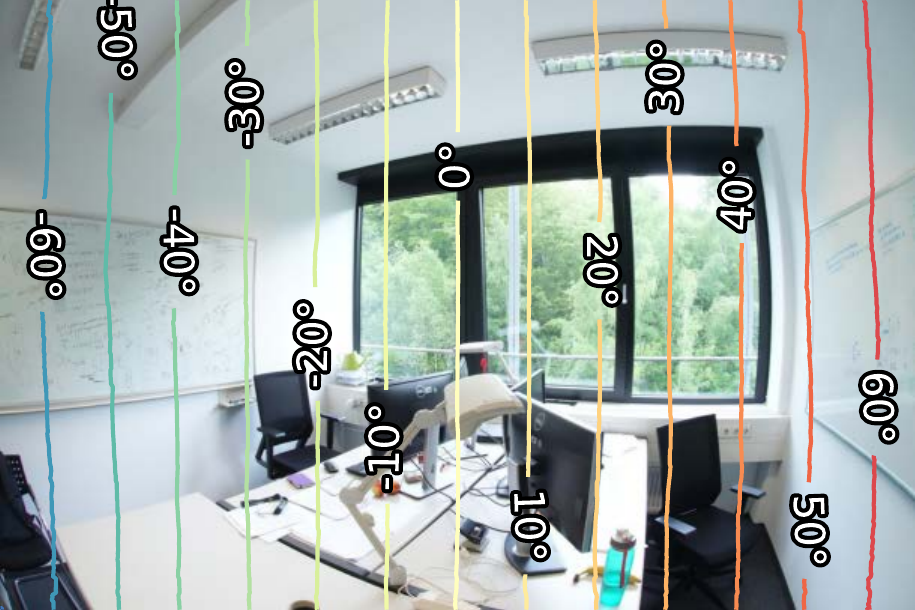} \\
        -         & $\theta_y$ & 
        \includegraphics[height=\figh,valign=m]{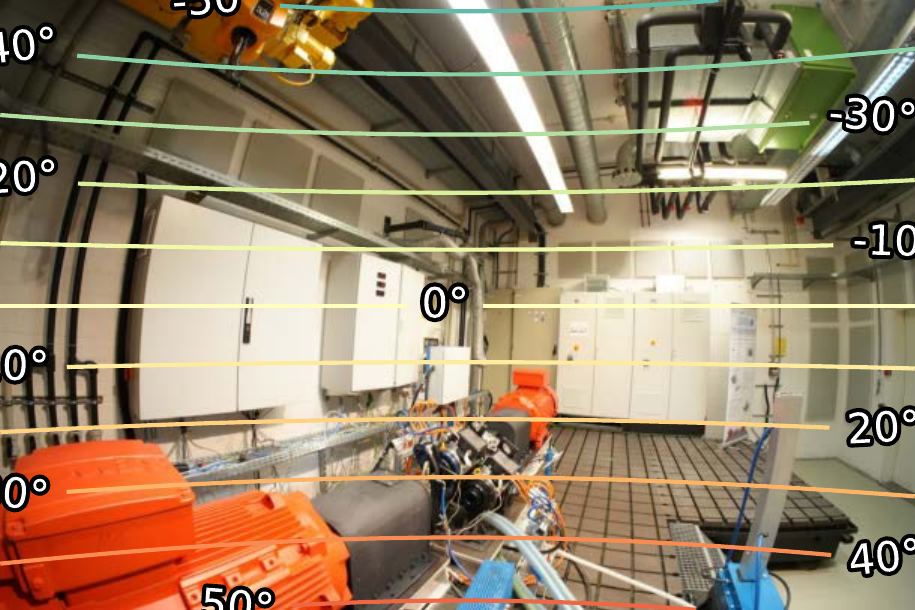} & 
        \includegraphics[height=\figh,valign=m]{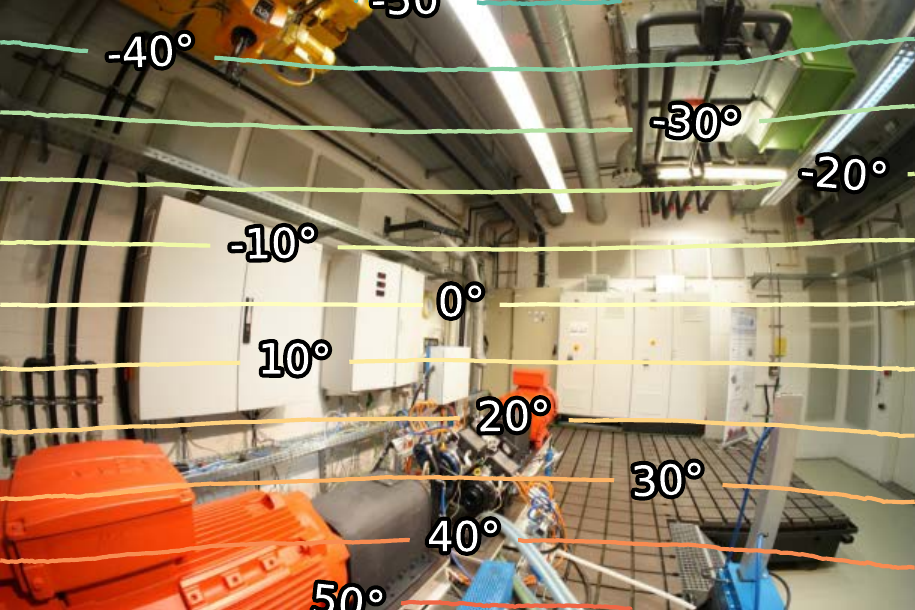} &
        \includegraphics[height=\figh,valign=m]{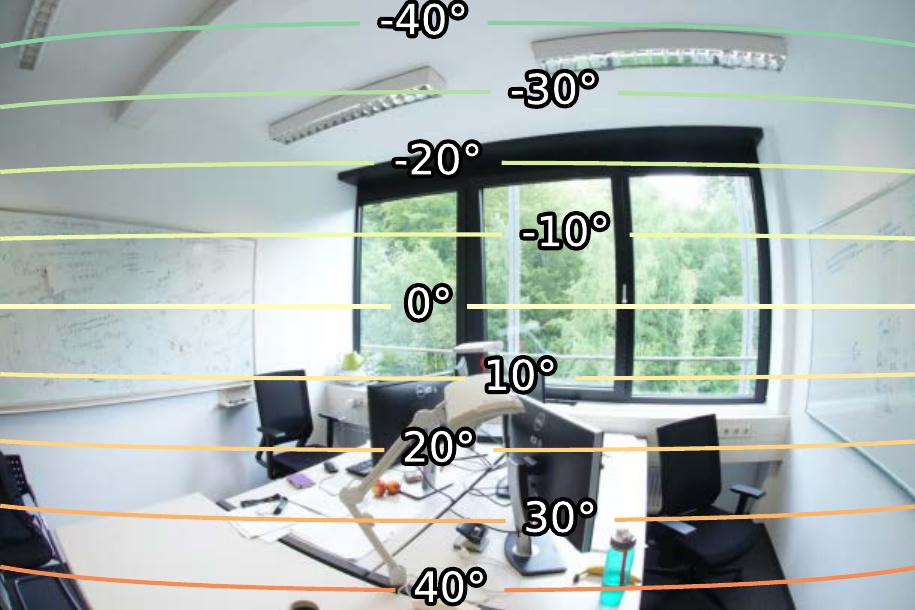} & 
        \includegraphics[height=\figh,valign=m]{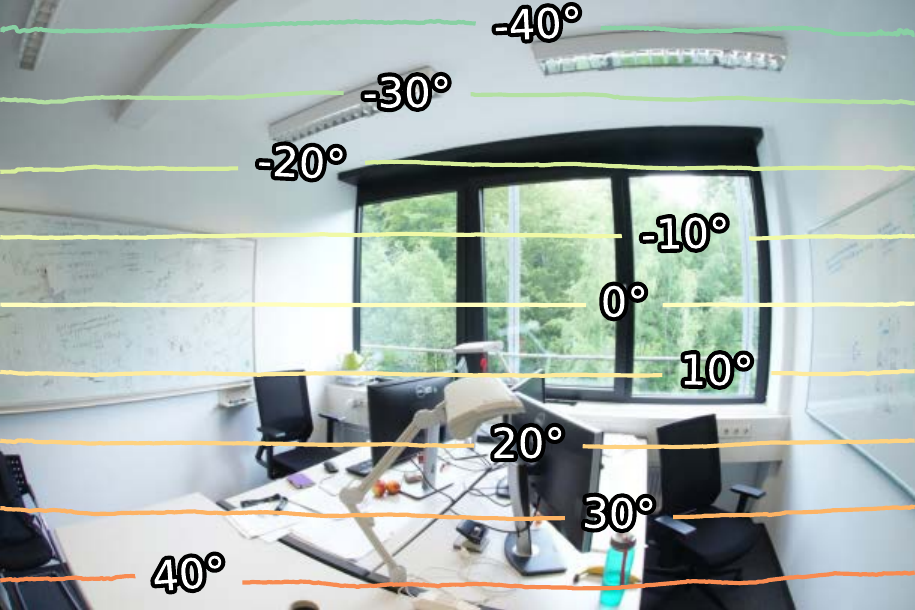} \\
        polar angles $\lVert\boldsymbol{\theta}\rVert$ & & 
        \includegraphics[height=\figh,valign=m]{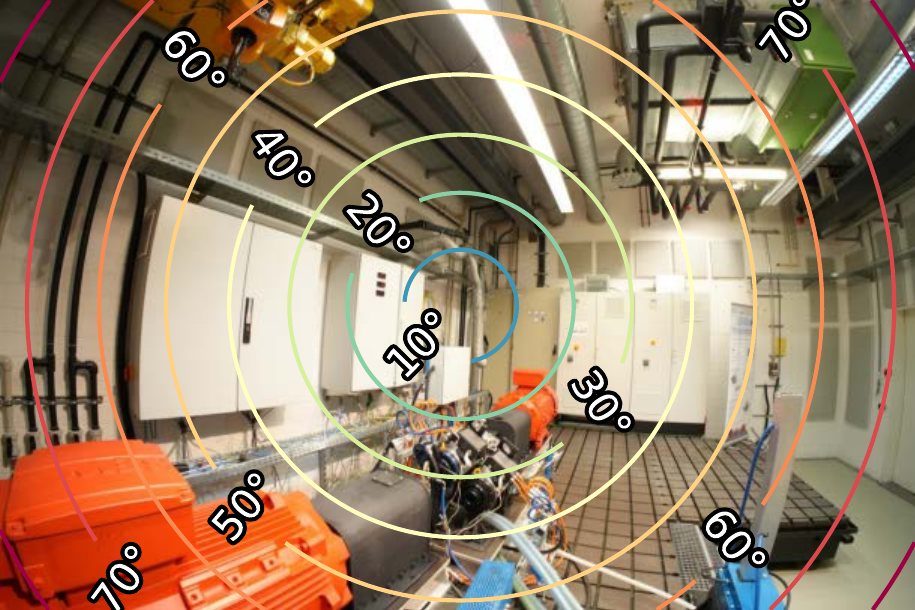} & 
        \includegraphics[height=\figh,valign=m]{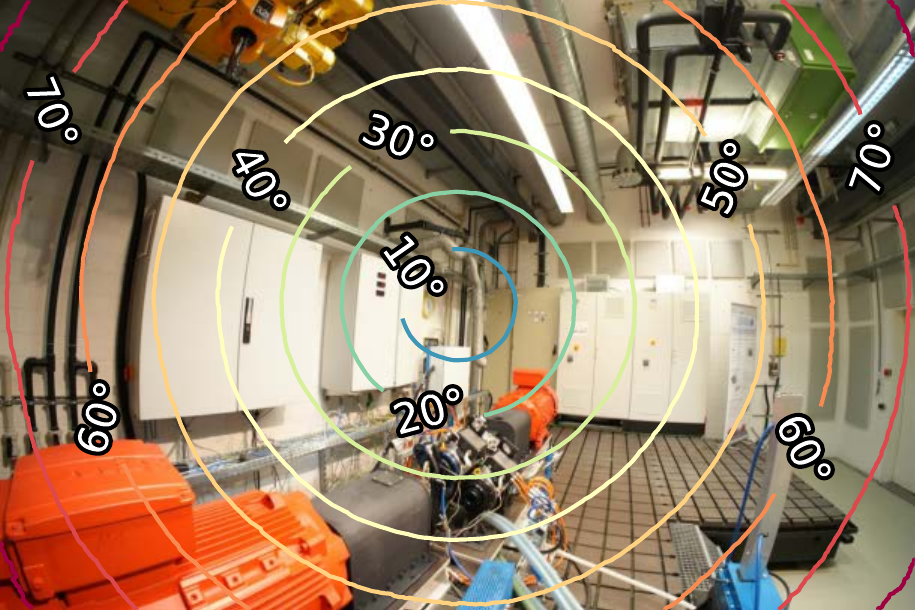} & 
        \includegraphics[height=\figh,valign=m]{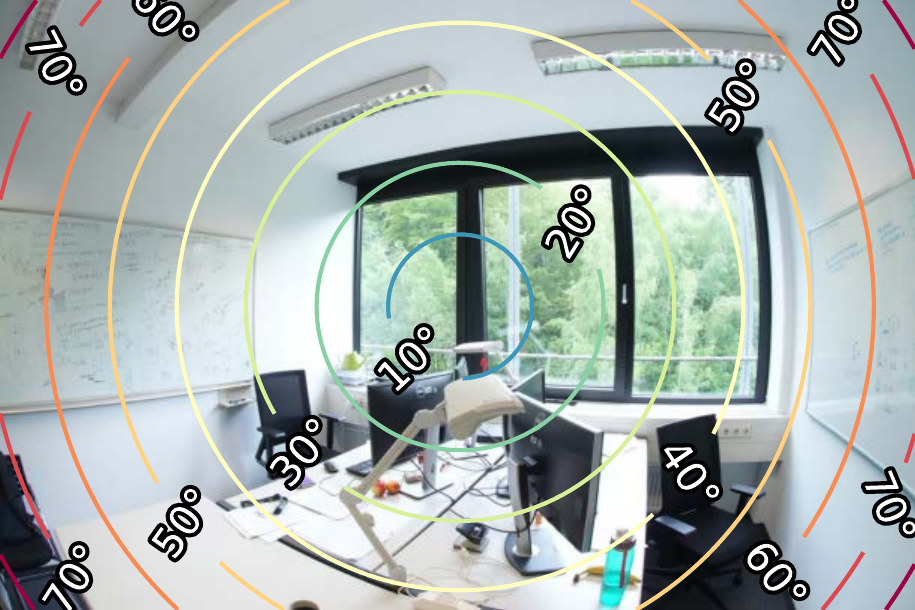} & 
        \includegraphics[height=\figh,valign=m]{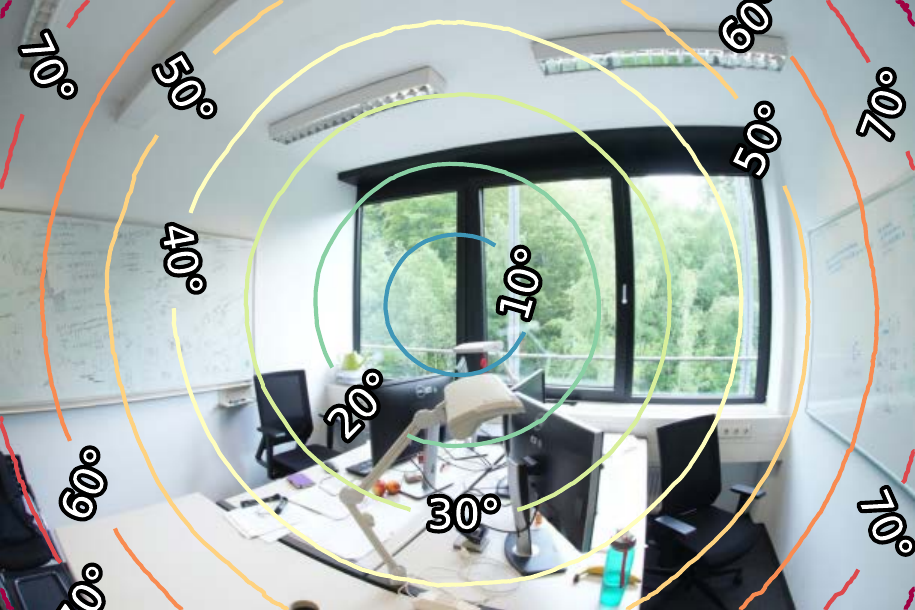} \\
        fits & \code{kb:4} & 
        \includegraphics[height=\figh,valign=m]{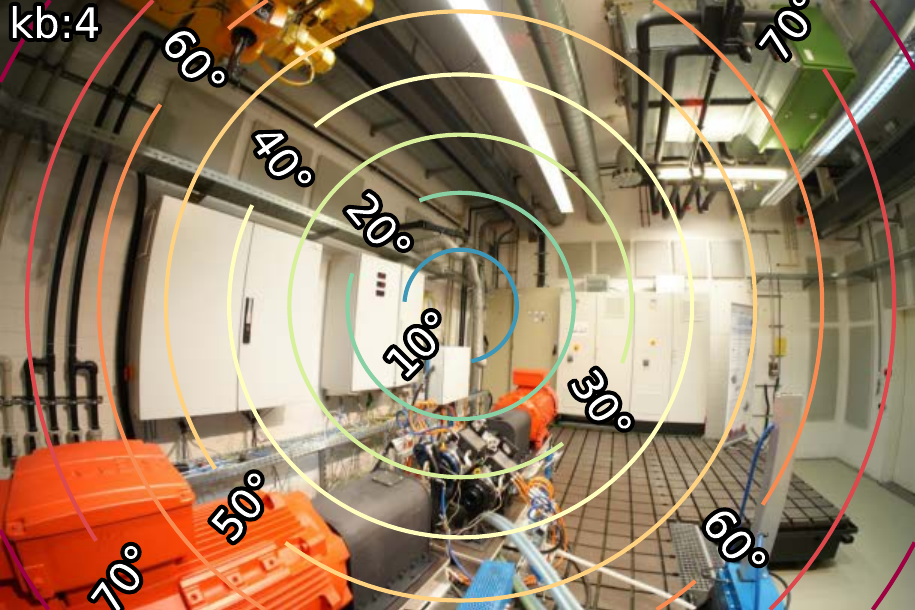} & 
        \includegraphics[height=\figh,valign=m]{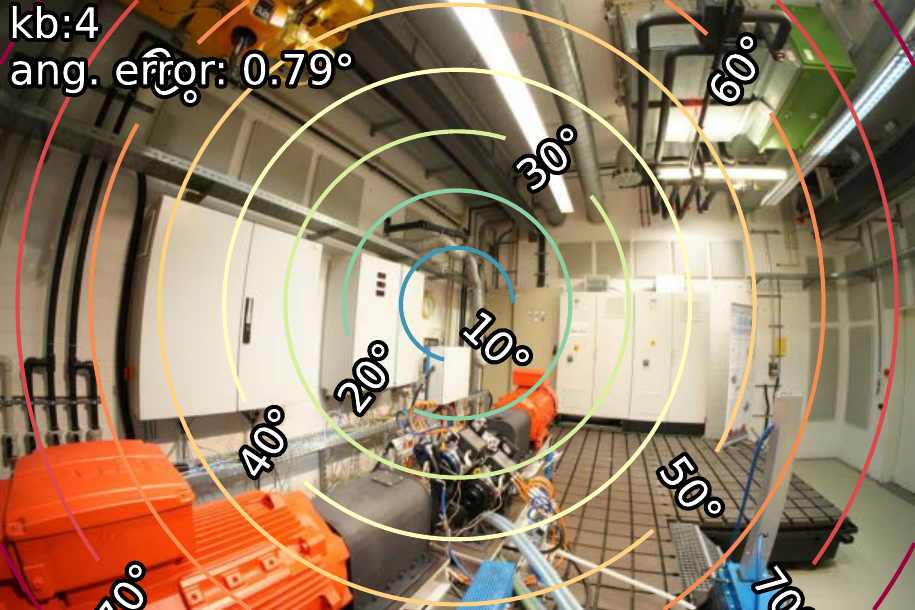} &
        \includegraphics[height=\figh,valign=m]{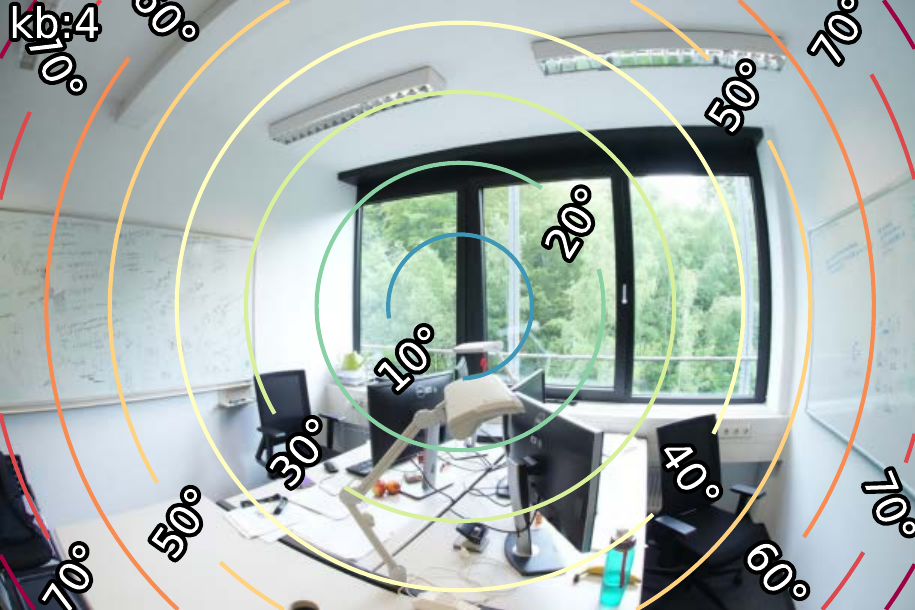} & 
        \includegraphics[height=\figh,valign=m]{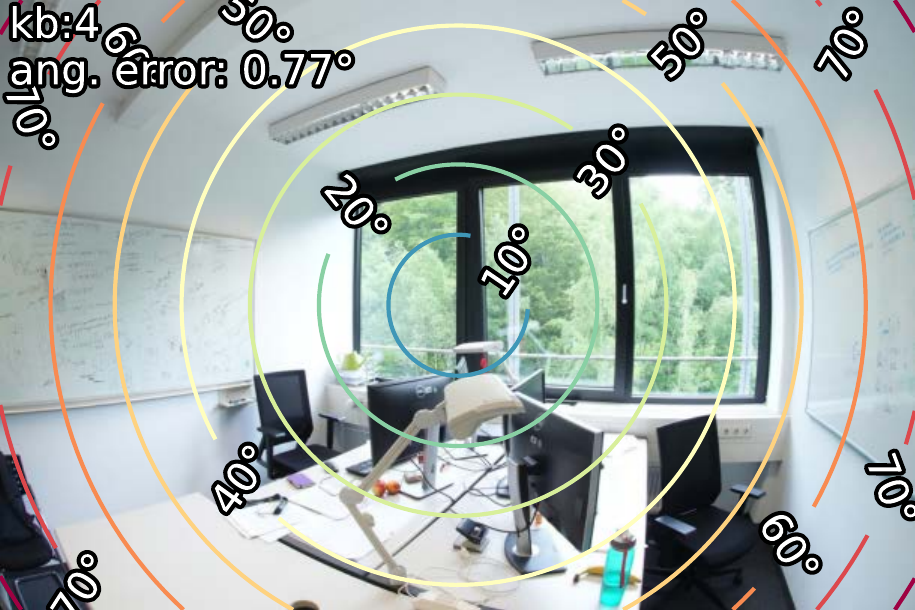} \\
        - & \code{ucm} && 
        \includegraphics[height=\figh,valign=m]{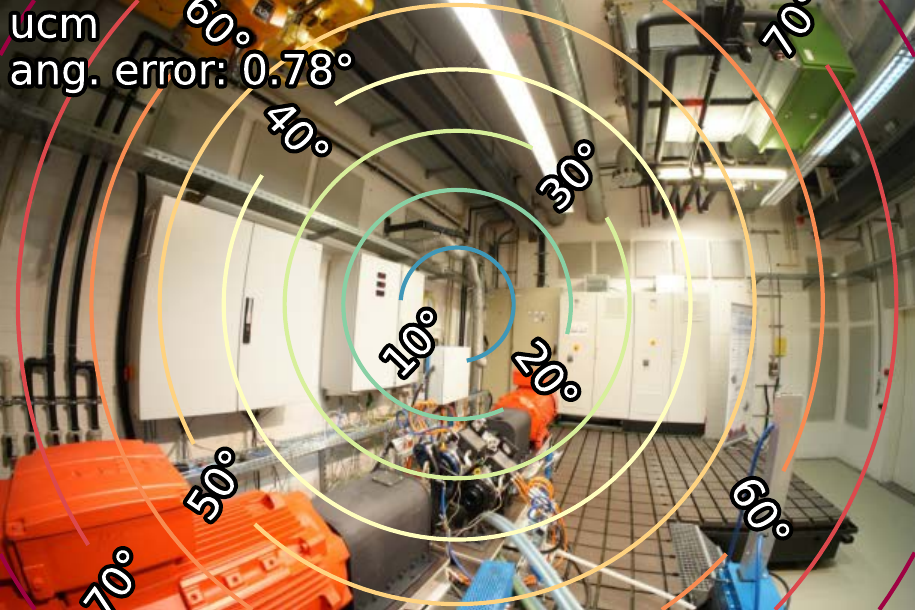} && 
        \includegraphics[height=\figh,valign=m]{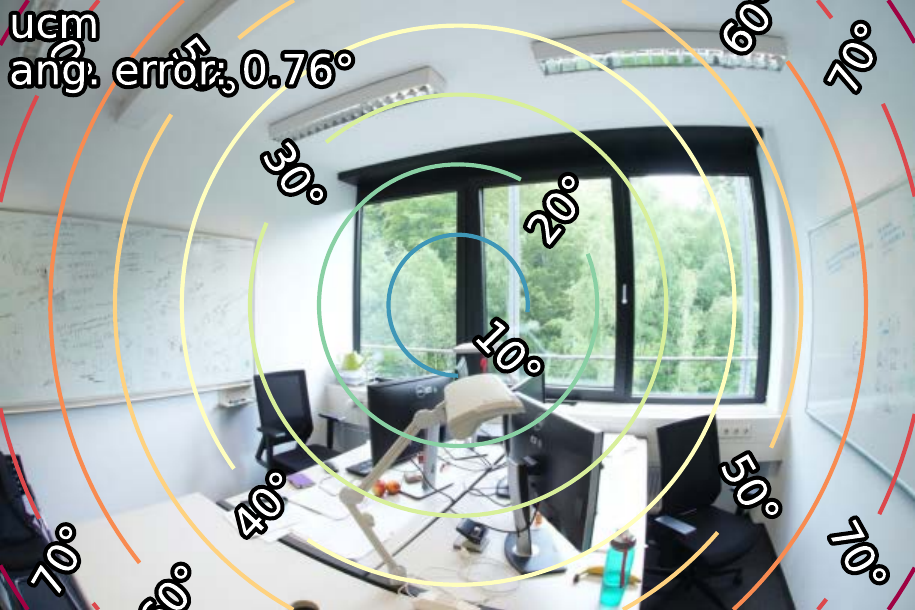} \\
        - & \code{eucm} && 
        \includegraphics[height=\figh,valign=m]{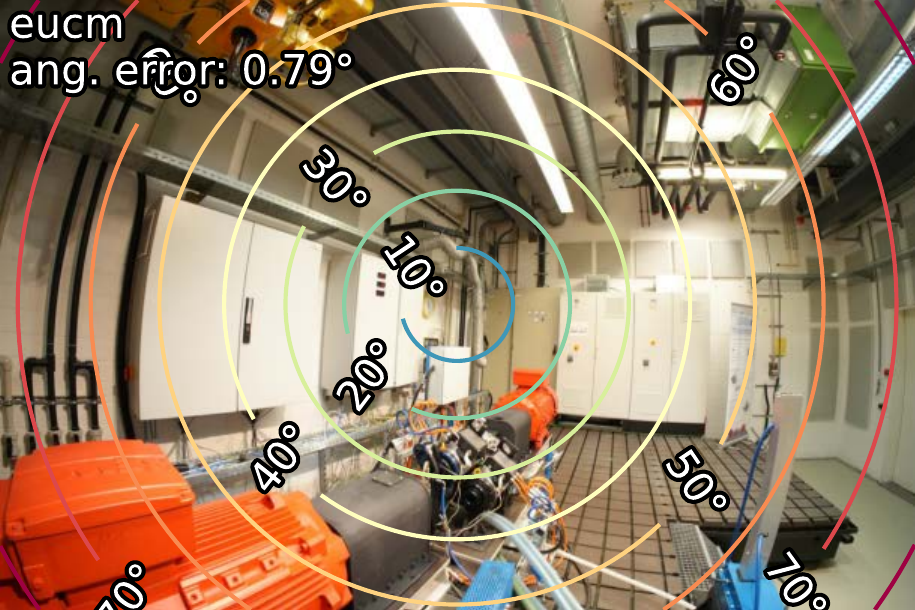} &&
        \includegraphics[height=\figh,valign=m]{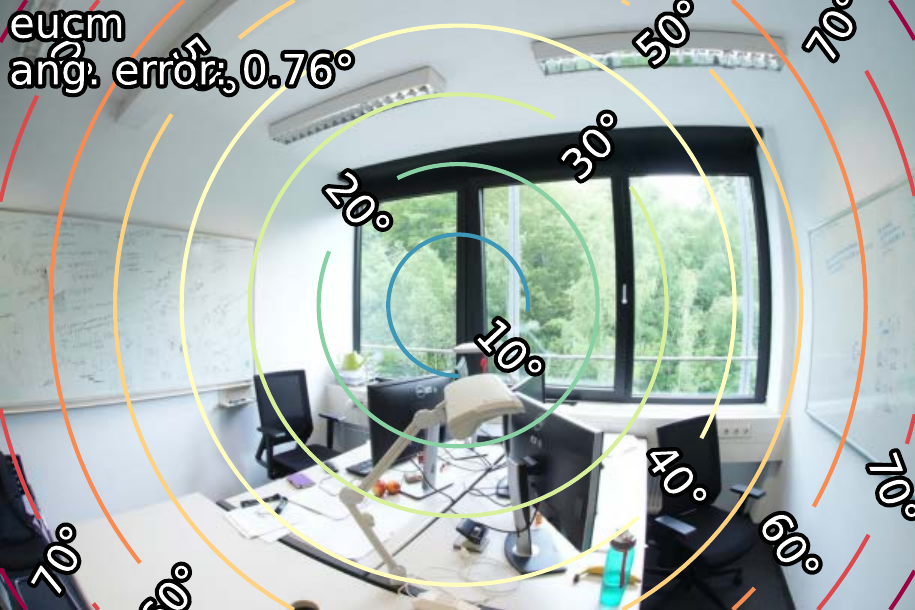} \\
    \end{tblr}
    \caption{\textbf{Qualitative results on distorted images} in ScanNet++ \cite{Yeshwanth2023scannetpp} with AnyCalib$_{\mathrm{gen}}$---trained on \datg. The FoV field ($\theta_x$ and $\theta_y$) is regressed by the network and $\lVert\boldsymbol\theta\rVert$ represents both its norm and the polar angle of the ray corresponding to each pixel. The predicted FoV field is used to fit the camera model of choice. \code{kb:i} corresponds to the Kannala-Brandt model with \code{i} distortion coefficients.}
    \label{fig:supp_qual_scan}
\end{figure*}

\begin{figure*}[t]

    \setlength{\figh}{2.7cm}

    \definecolor{codecolor}{RGB}{246, 248, 250}
    \newcommand{\code}[1]{\colorbox{codecolor}{\texttt{#1}}}

    \newcommand{\lfnamebase}{figures/supp_qualitative_distorted/sequence_42_00058}
    \newcommand{\rfnamebase}{figures/supp_qualitative_distorted/sequence_48_01185}

    \centering
    \begin{tblr}{
        width=1.0\linewidth,
        vspan=even,
        colspec={*{6}c},
        colsep=1pt,
        rowsep=1pt,
        cell{2}{1}={r=2}{c, cmd=\rotatebox{90}},
        cell{4}{1}={c, cmd={\rotatebox[origin=c]{90}}},
        cell{5,6,7}{2}={c, cmd={\rotatebox[origin=c]{90}}},
        cell{5}{1}={r=3}{c, cmd=\rotatebox{90}},
    }
        && ground-truth & prediction & ground-truth & prediction \\
        FoV field & $\theta_x$ & 
        \includegraphics[height=\figh,valign=m]{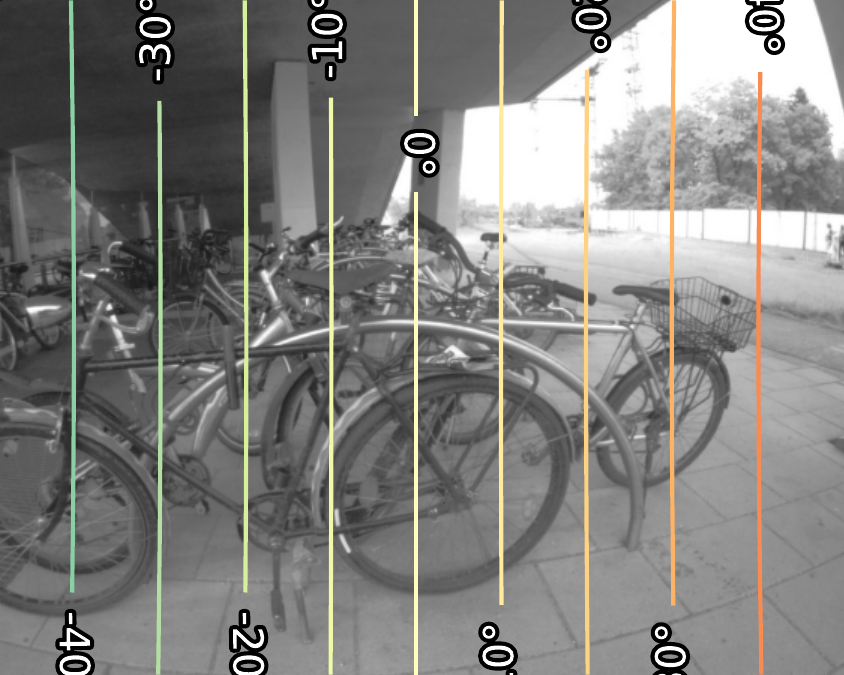} & 
        \includegraphics[height=\figh,valign=m]{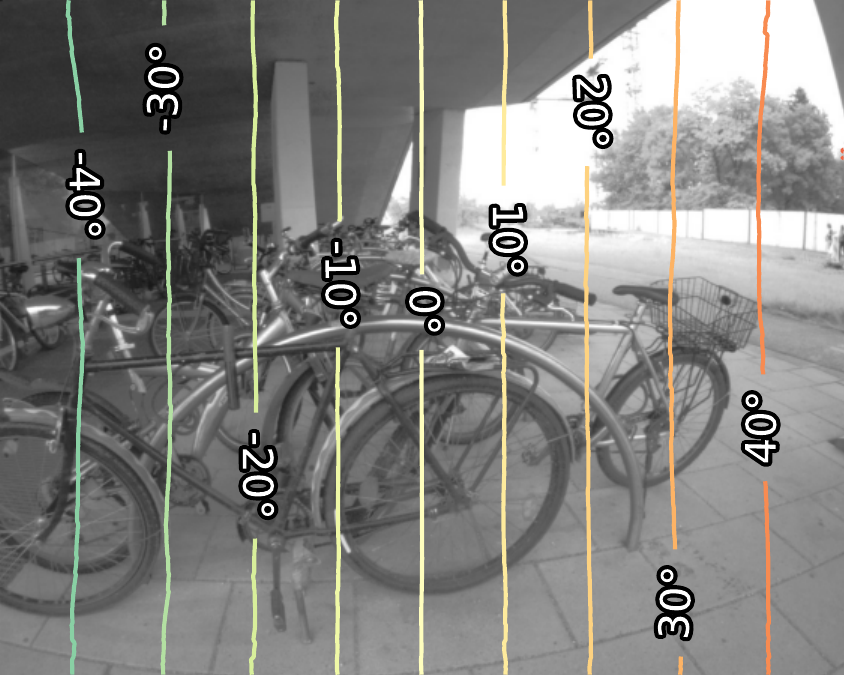} & 
        \includegraphics[height=\figh,valign=m]{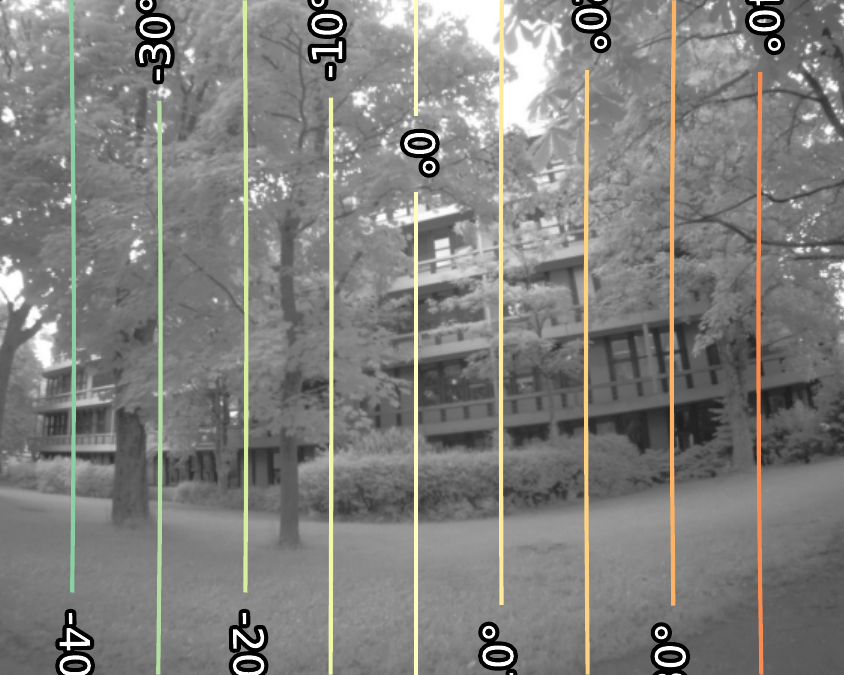} & 
        \includegraphics[height=\figh,valign=m]{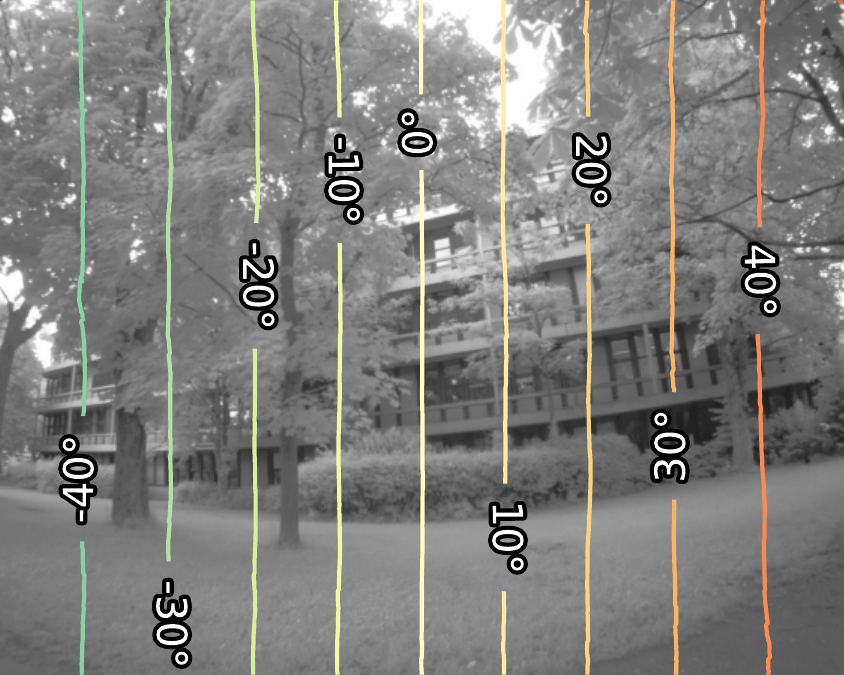} \\
        -         & $\theta_y$ & 
        \includegraphics[height=\figh,valign=m]{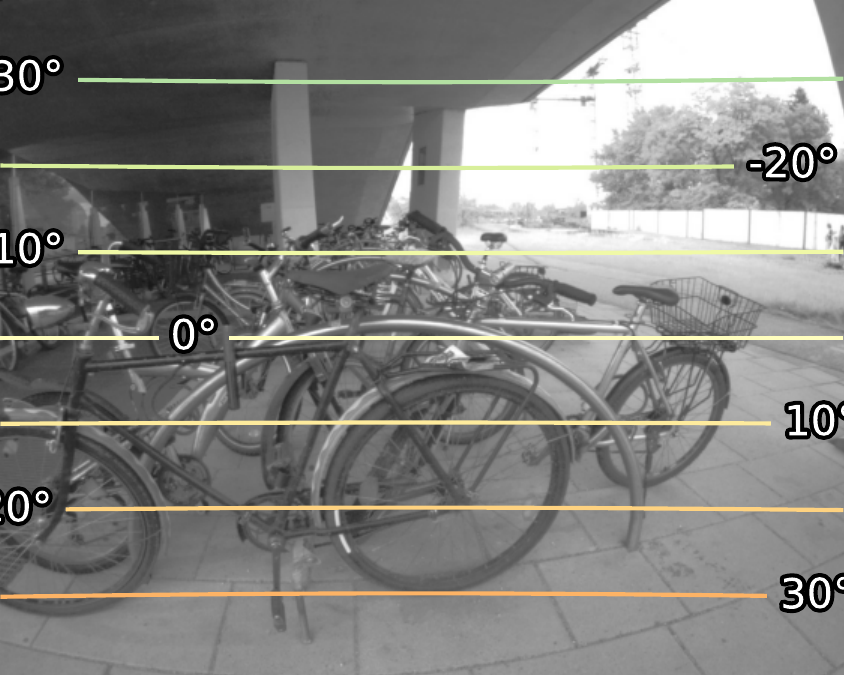} & 
        \includegraphics[height=\figh,valign=m]{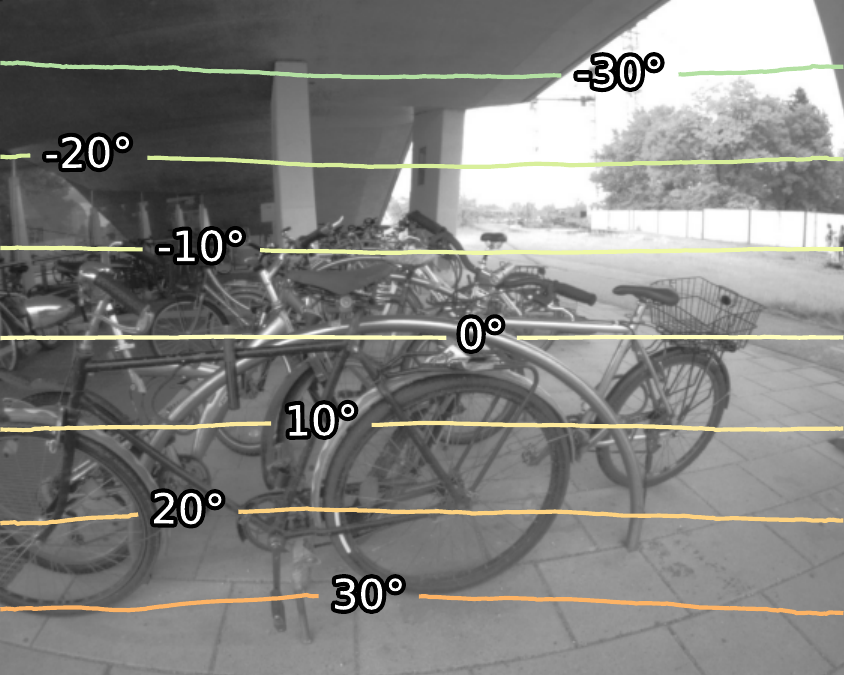} &
        \includegraphics[height=\figh,valign=m]{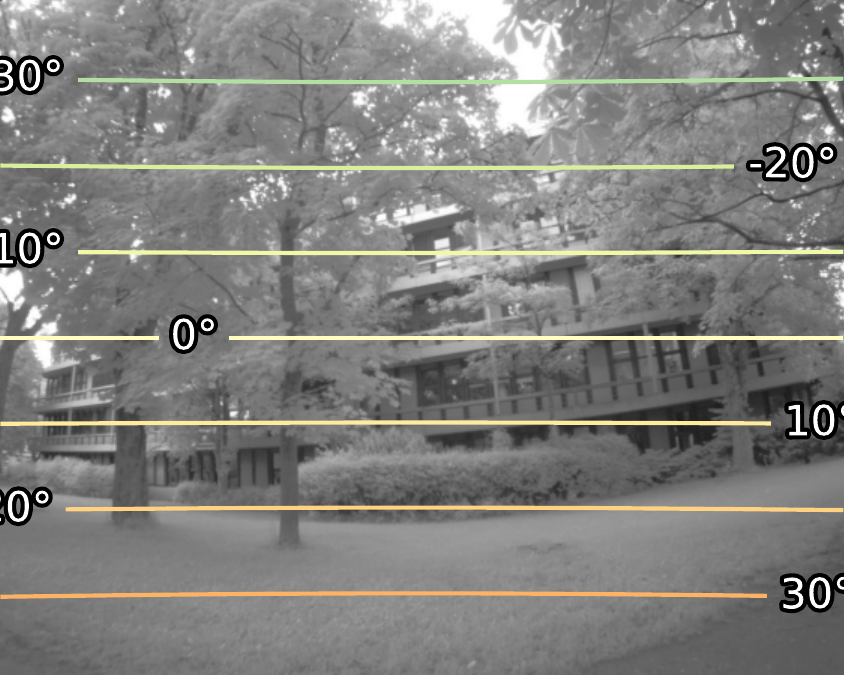} & 
        \includegraphics[height=\figh,valign=m]{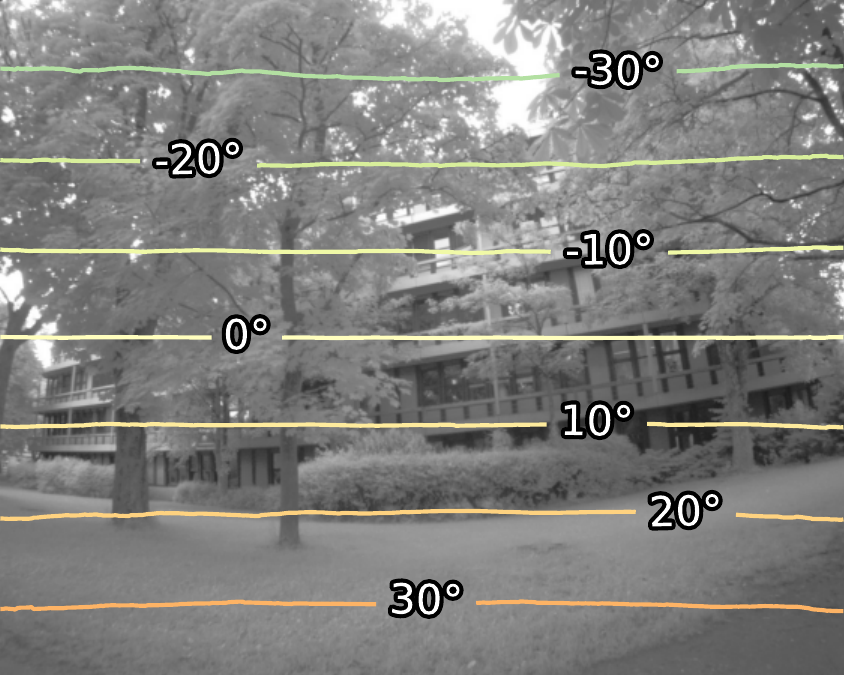} \\
        polar angles $\lVert\boldsymbol{\theta}\rVert$ & & 
        \includegraphics[height=\figh,valign=m]{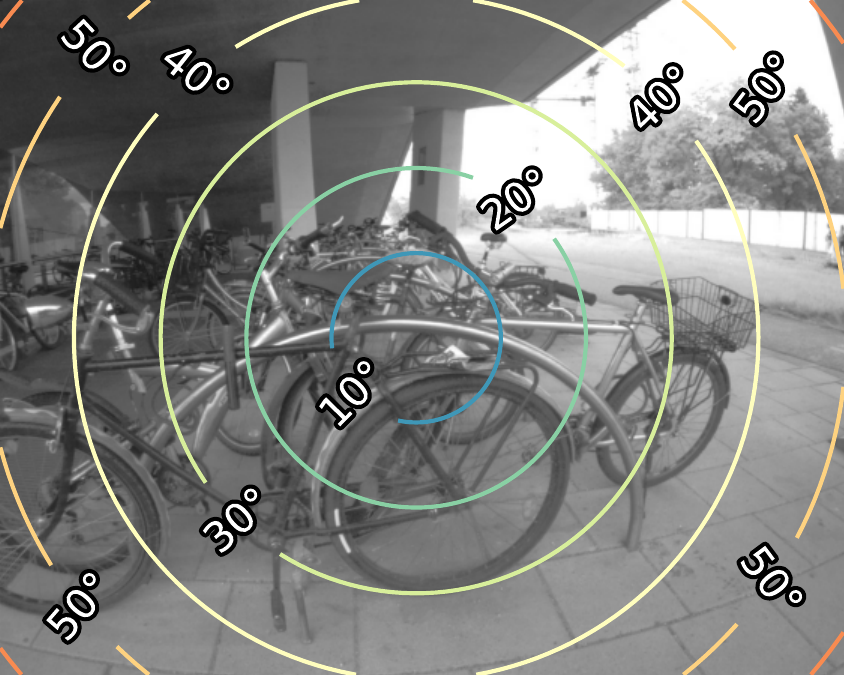} & 
        \includegraphics[height=\figh,valign=m]{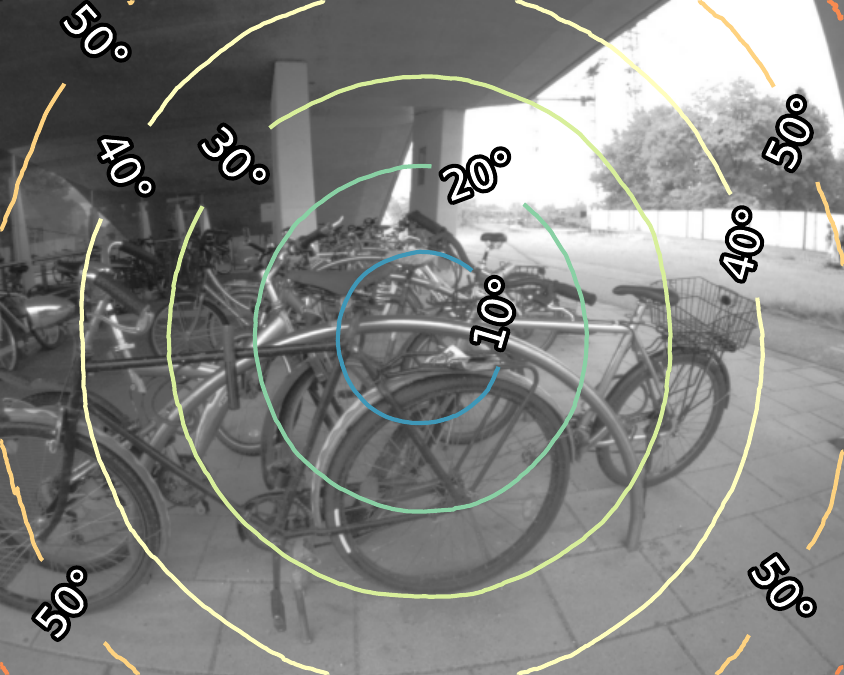} & 
        \includegraphics[height=\figh,valign=m]{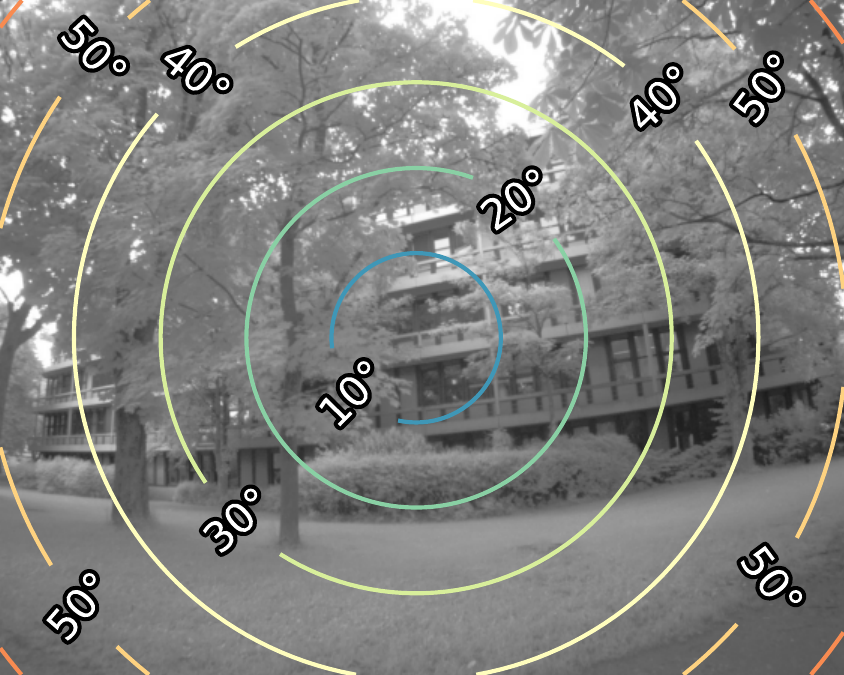} & 
        \includegraphics[height=\figh,valign=m]{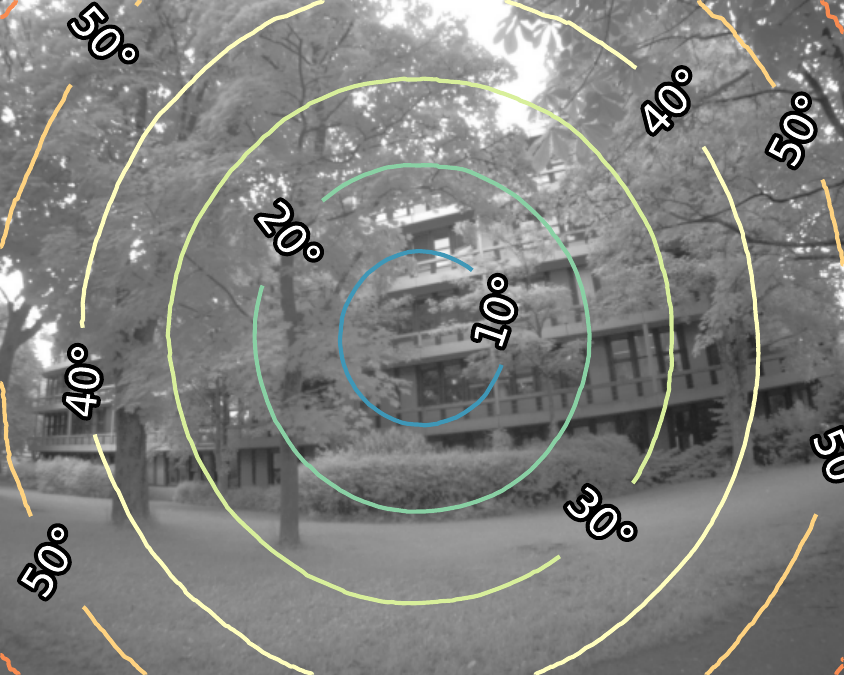} \\
        fits & \code{kb:4} & 
        \includegraphics[height=\figh,valign=m]{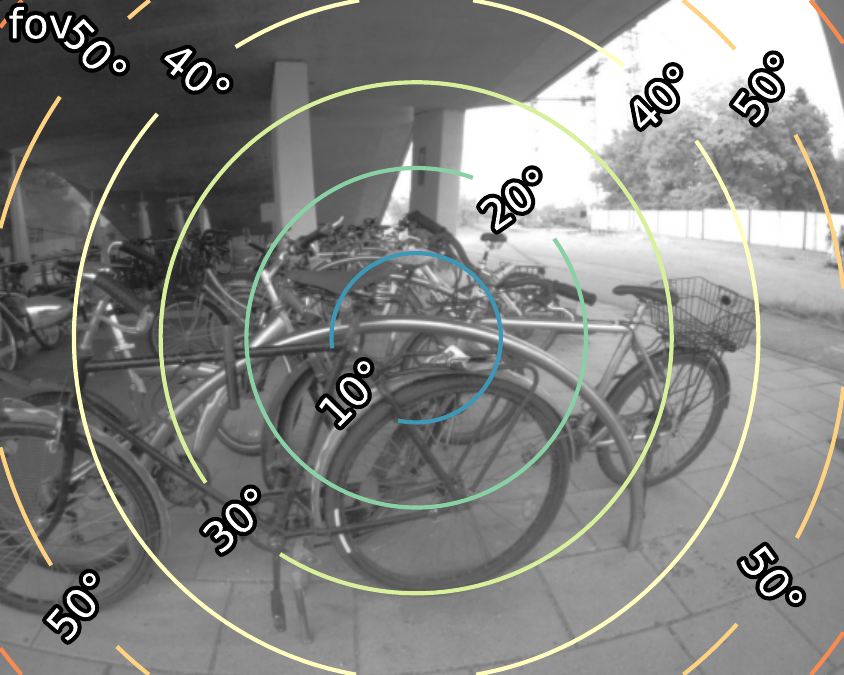} & 
        \includegraphics[height=\figh,valign=m]{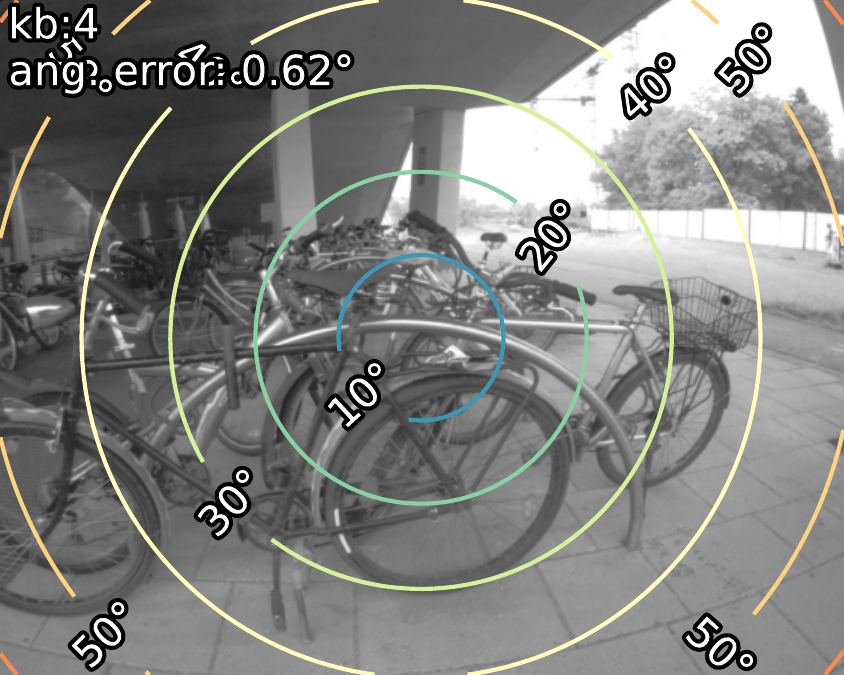} &
        \includegraphics[height=\figh,valign=m]{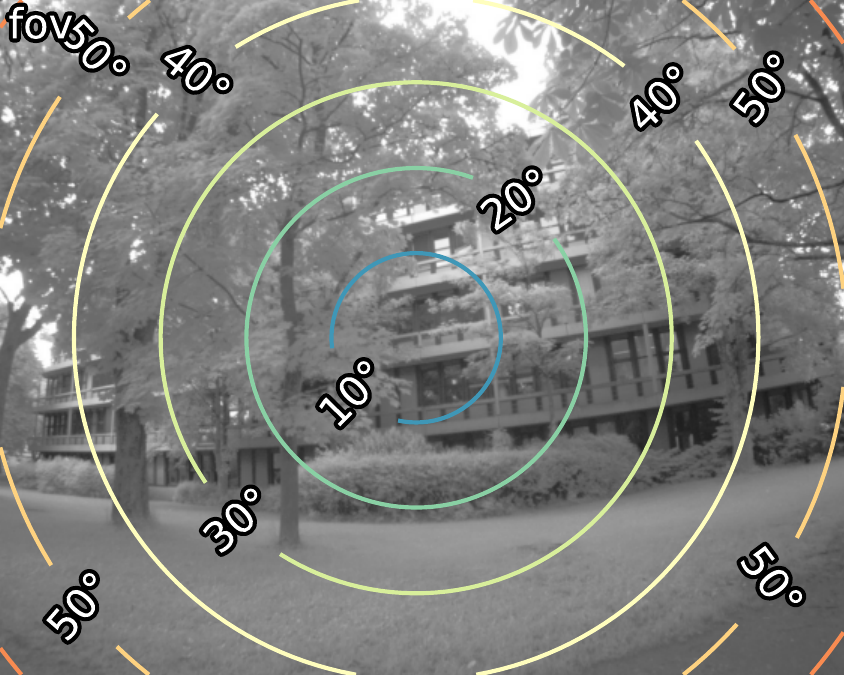} & 
        \includegraphics[height=\figh,valign=m]{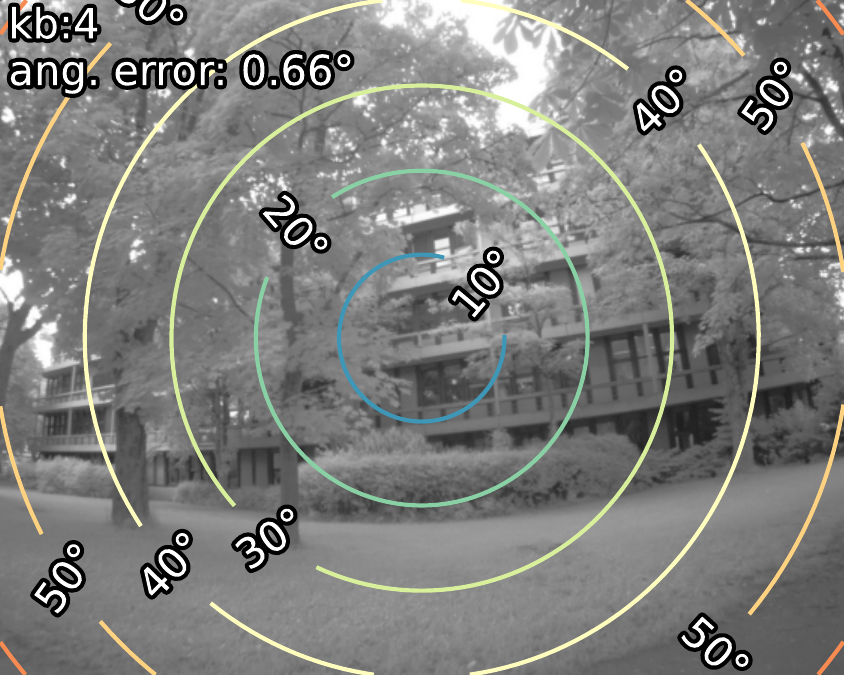} \\
        - & \code{division:2} && 
        \includegraphics[height=\figh,valign=m]{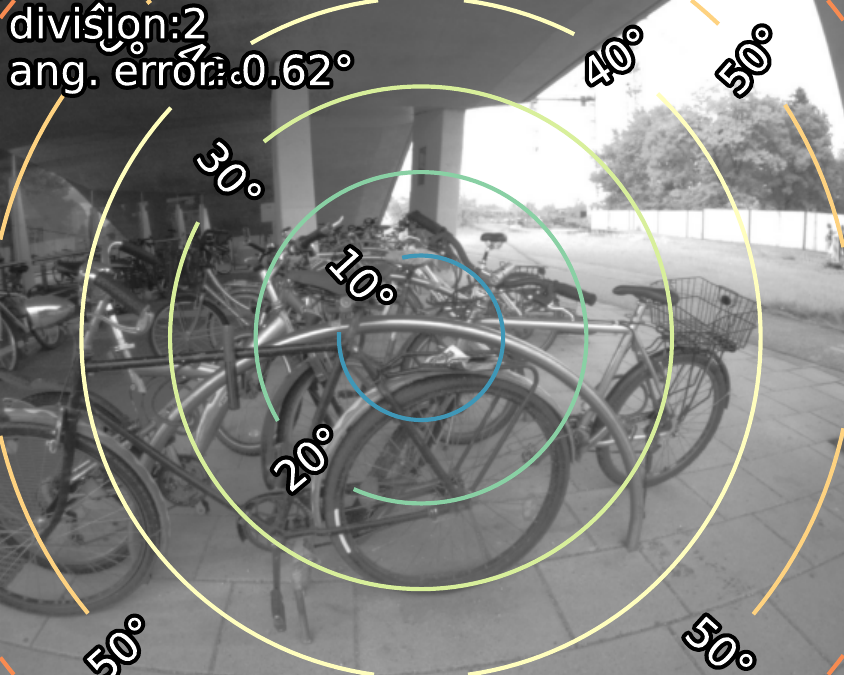} && 
        \includegraphics[height=\figh,valign=m]{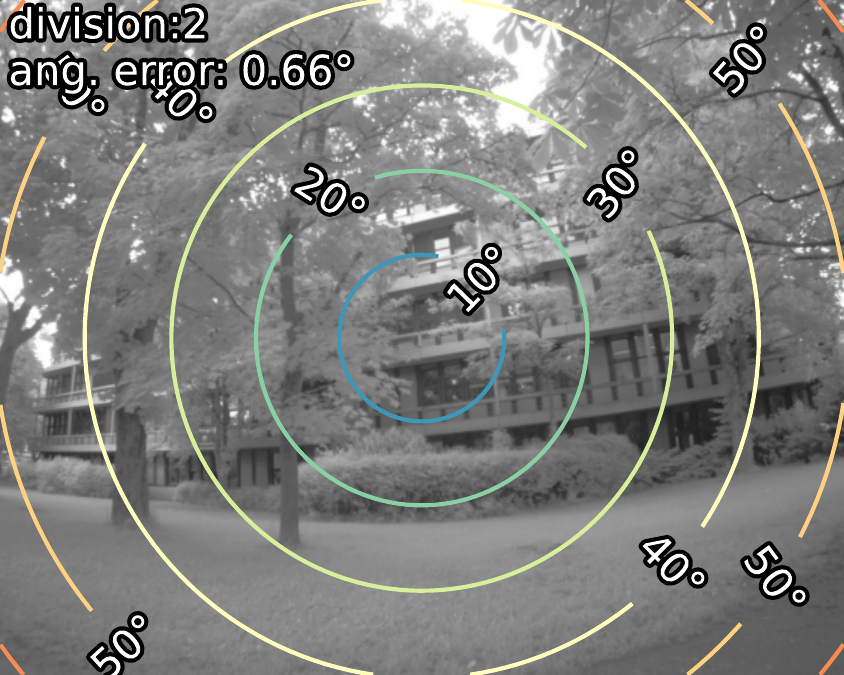} \\
        - & \code{division:3} && 
        \includegraphics[height=\figh,valign=m]{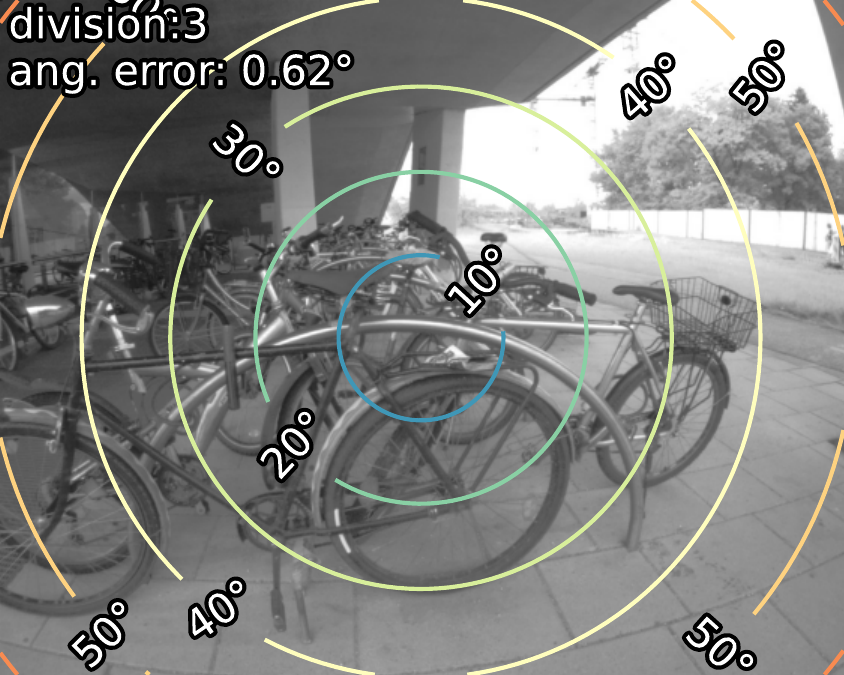} &&
        \includegraphics[height=\figh,valign=m]{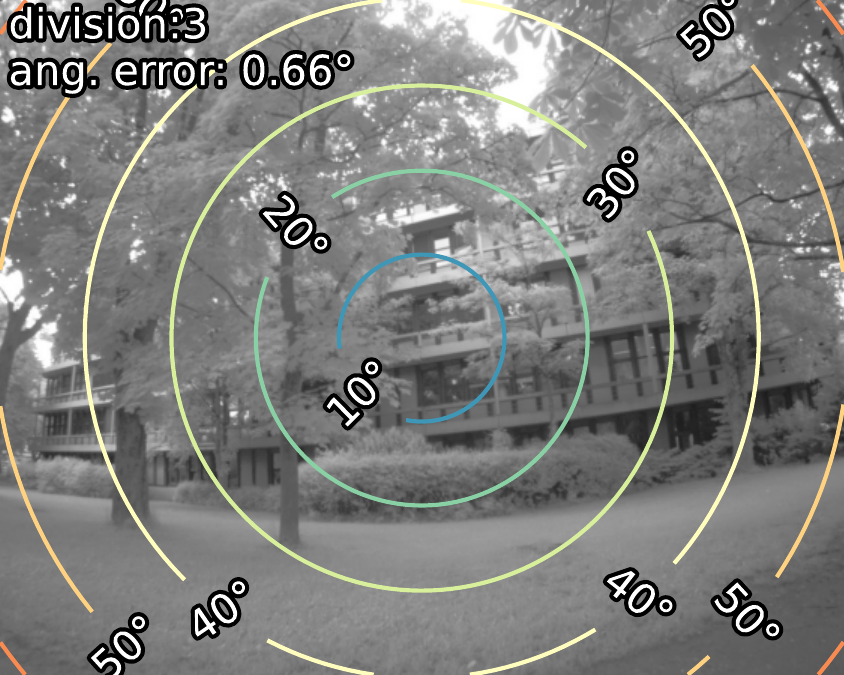} \\
    \end{tblr}
    \caption{\textbf{Qualitative results on distorted images} in the Mono Dataset \cite{engel2016monodataset} with AnyCalib$_{\mathrm{gen}}$---trained on \datg. The FoV field ($\theta_x$ and $\theta_y$) is regressed by the network and $\lVert\boldsymbol\theta\rVert$ represents both its norm and the polar angle of the ray corresponding to each pixel. The predicted FoV field is used to fit the camera model of choice. \code{division:i} corresponds to the division model with \code{i} distortion coefficients.}
    \label{fig:supp_qual_monovo}
\end{figure*}

\begin{figure*}[t]

    \newlength{\ufigh}
    \setlength{\ufigh}{2.8cm}
    \newlength{\bfigh}
    \setlength{\bfigh}{2.3cm}

    \definecolor{codecolor}{RGB}{246, 248, 250}
    \newcommand{\code}[1]{\colorbox{codecolor}{\texttt{#1}}}

    \newcommand{\lufnamebase}{figures/supp_qualitative_edited/edit_101085075990}
    \newcommand{\lbfnamebase}{figures/supp_qualitative_edited/edit_area_6-camera_31f0030b31a8419282da58dd6d507591_office_31_frame_equirectangular_domain_rgb_r-24_p-1_y32_f60}
    \newcommand{\rbfnamebase}{figures/supp_qualitative_edited/edit_japanesealley-japanesealley-Easy-P002-image_left-000485_left}
    \newcommand{\rufnamebase}{figures/supp_qualitative_edited/edit_neighborhood-neighborhood-Easy-P018-image_left-000074_left}

    \centering
    \begin{tblr}{
        width=1.0\linewidth,
        vspan=even,
        colspec={*{6}c},
        colsep=1pt,
        rowsep=1pt,
        cell{2,6}{1}={r=2}{c, cmd=\rotatebox{90}},
        cell{4,8}{1}={c, cmd={\rotatebox[origin=c]{90}}},
        cell{5}{1,2}={c, cmd={\rotatebox[origin=c]{90}}},
        cell{9}{1,2}={c, cmd={\rotatebox[origin=c]{90}}},
    }
        && ground-truth & prediction & ground-truth & prediction \\
        FoV field & $\theta_x$ & 
        \includegraphics[height=\ufigh,valign=m]{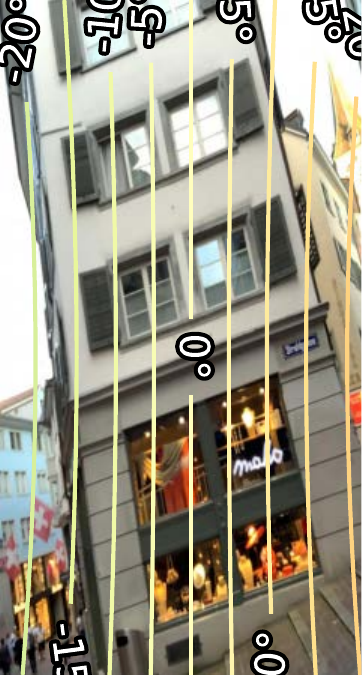} & 
        \includegraphics[height=\ufigh,valign=m]{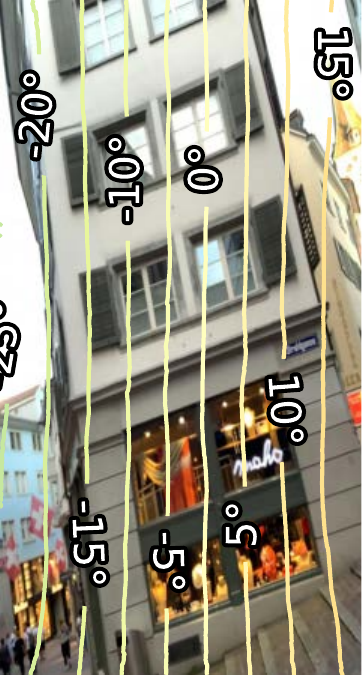} & 
        \includegraphics[height=\ufigh,valign=m]{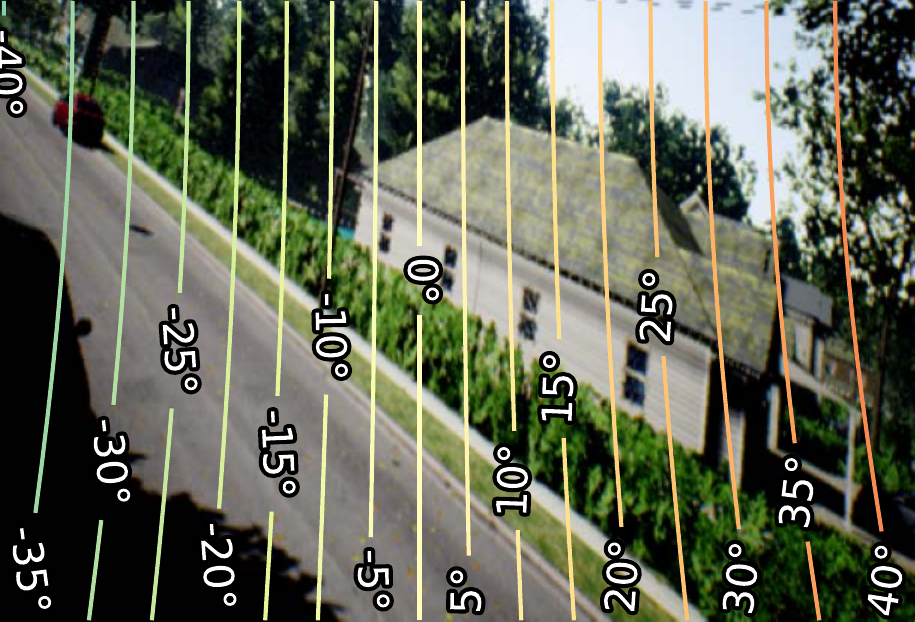} & 
        \includegraphics[height=\ufigh,valign=m]{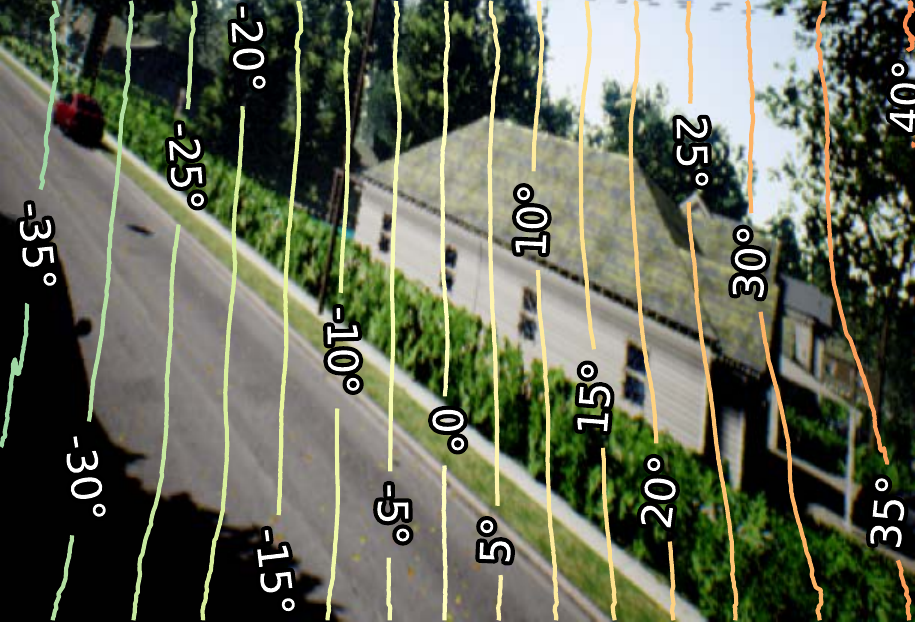} \\
        -         & $\theta_y$ & 
        \includegraphics[height=\ufigh,valign=m]{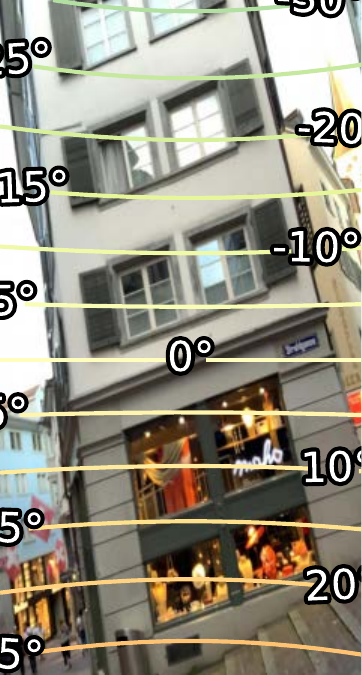} & 
        \includegraphics[height=\ufigh,valign=m]{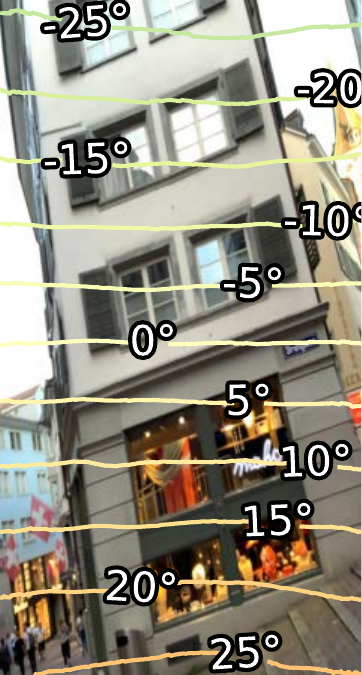} &
        \includegraphics[height=\ufigh,valign=m]{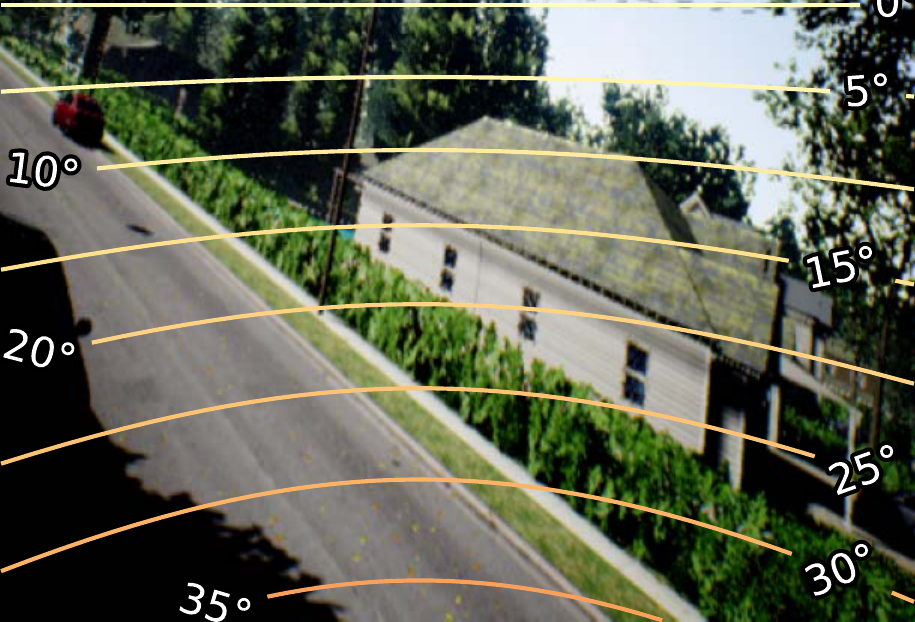} & 
        \includegraphics[height=\ufigh,valign=m]{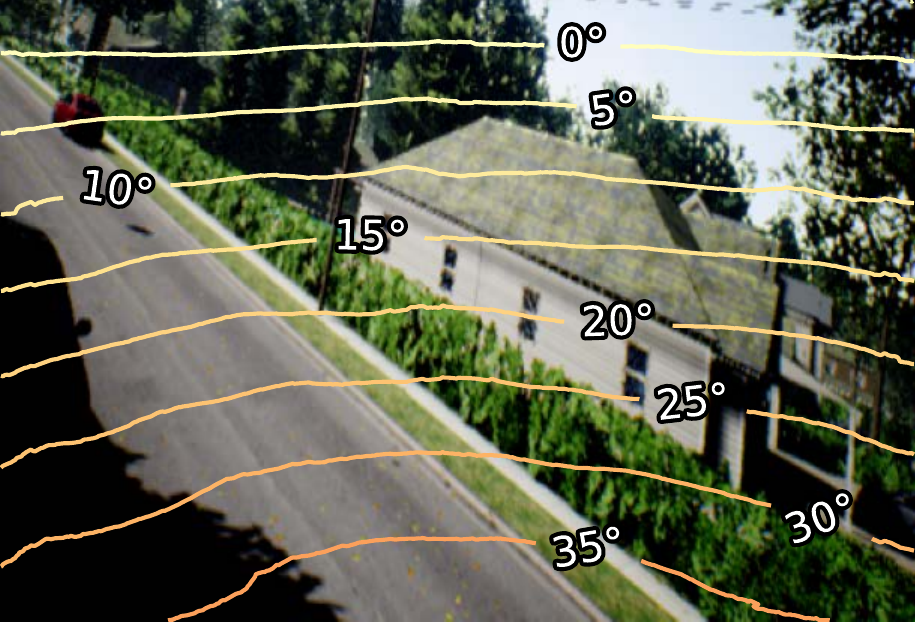} \\
        polar angles $\lVert\boldsymbol{\theta}\rVert$ & & 
        \includegraphics[height=\ufigh,valign=m]{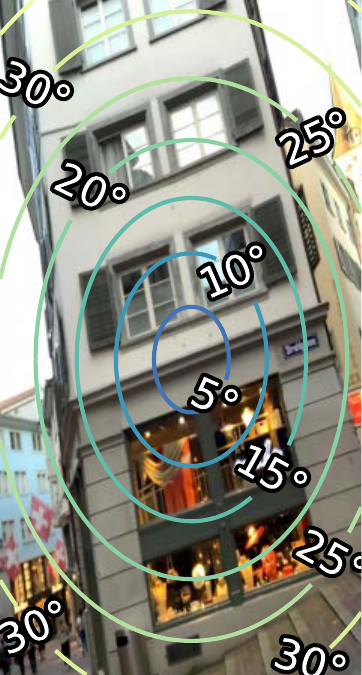} & 
        \includegraphics[height=\ufigh,valign=m]{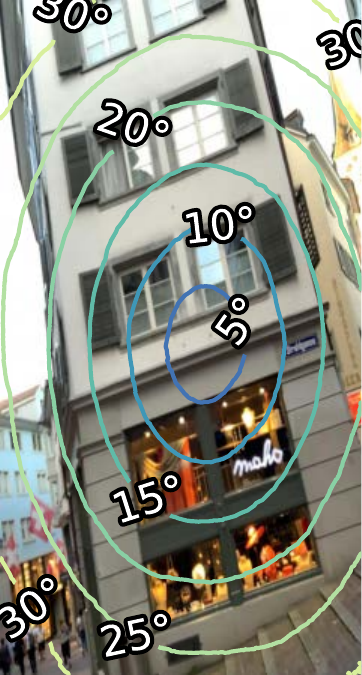} & 
        \includegraphics[height=\ufigh,valign=m]{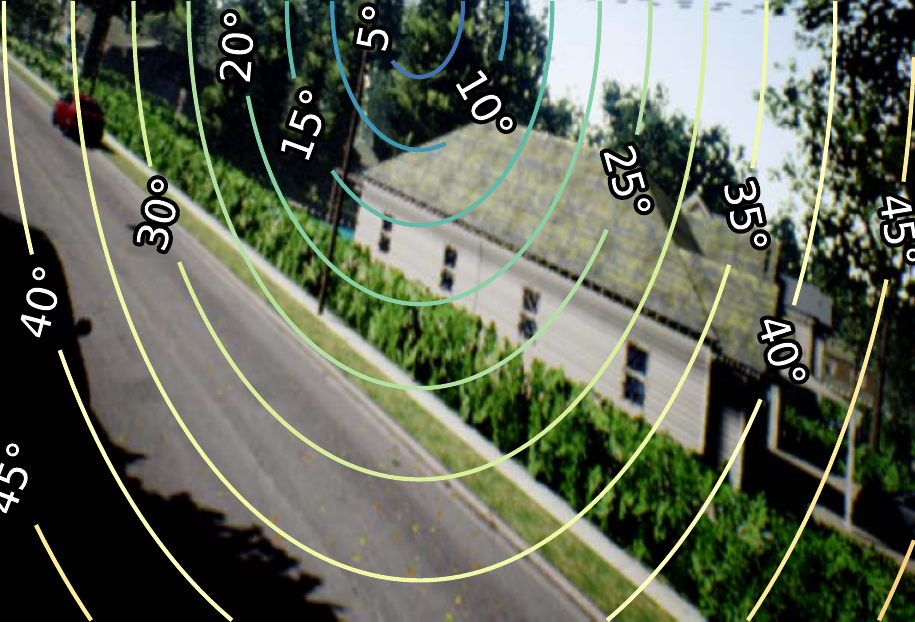} & 
        \includegraphics[height=\ufigh,valign=m]{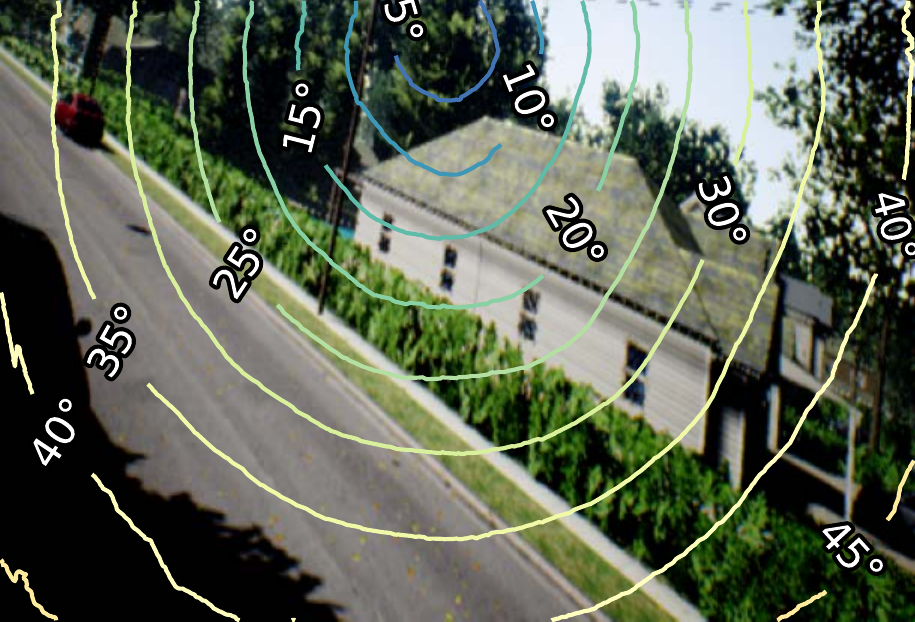} \\
        fit & \code{pinhole} & 
        \includegraphics[height=\ufigh,valign=m]{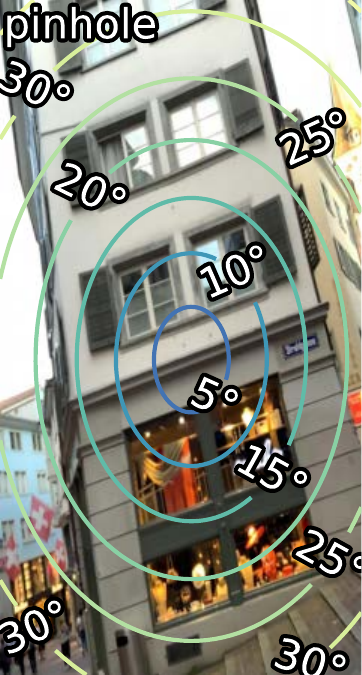} & 
        \includegraphics[height=\ufigh,valign=m]{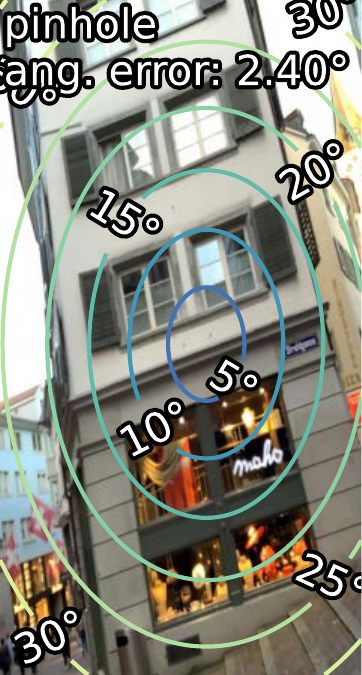} &
        \includegraphics[height=\ufigh,valign=m]{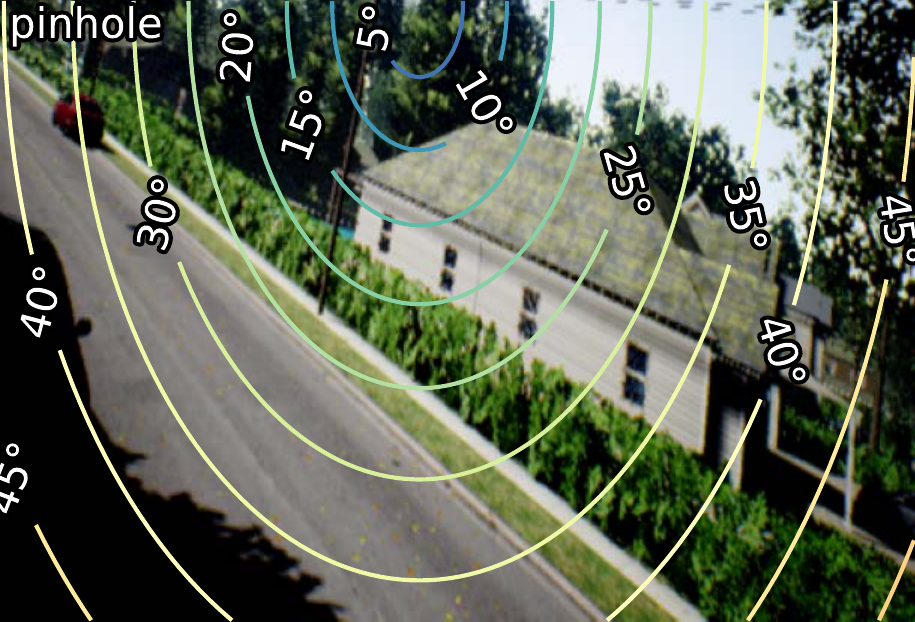} & 
        \includegraphics[height=\ufigh,valign=m]{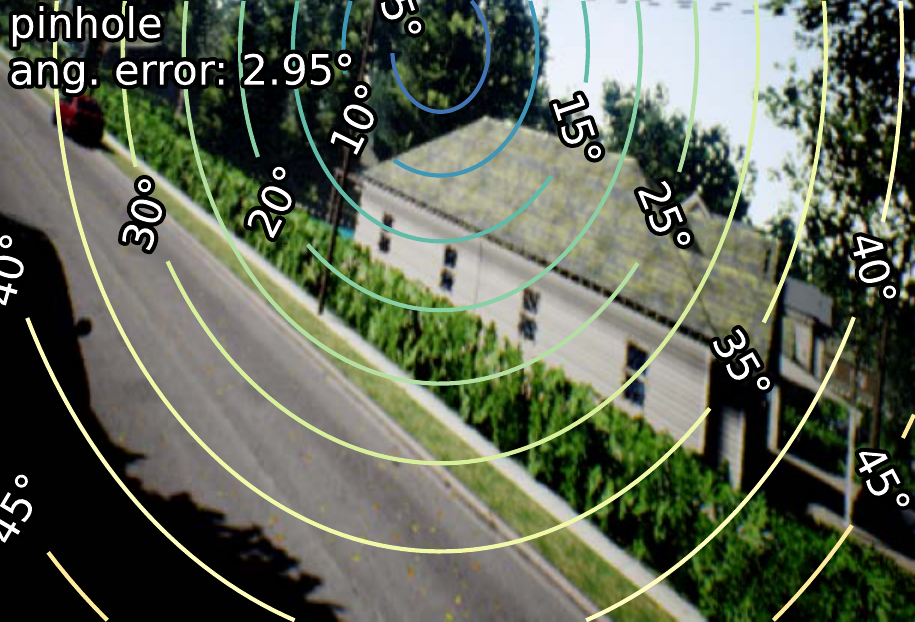} \\
        FoV field & $\theta_x$ & 
        \includegraphics[height=\bfigh,valign=m]{\lbfnamebase/theta_x_gt.pdf} & 
        \includegraphics[height=\bfigh,valign=m]{\lbfnamebase/theta_x.pdf} & 
        \includegraphics[height=\bfigh,valign=m]{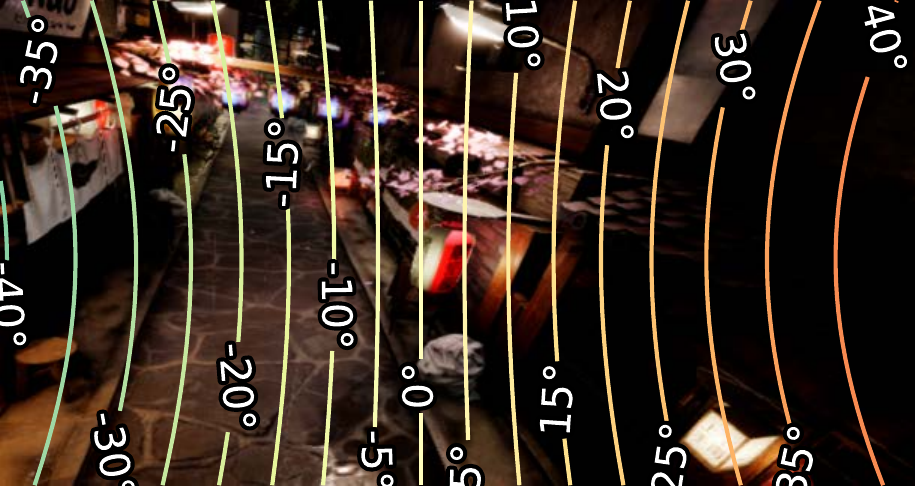} & 
        \includegraphics[height=\bfigh,valign=m]{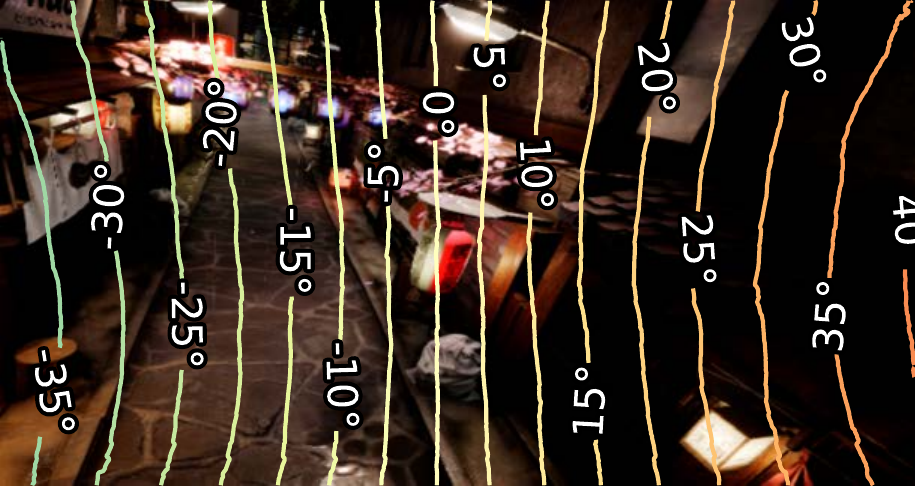} \\
        -         & $\theta_y$ & 
        \includegraphics[height=\bfigh,valign=m]{\lbfnamebase/theta_y_gt.pdf} & 
        \includegraphics[height=\bfigh,valign=m]{\lbfnamebase/theta_y.pdf} &
        \includegraphics[height=\bfigh,valign=m]{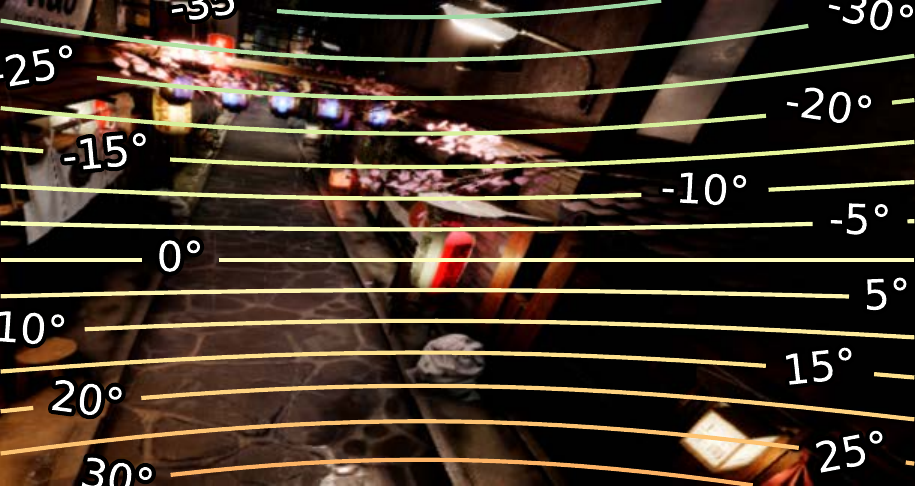} & 
        \includegraphics[height=\bfigh,valign=m]{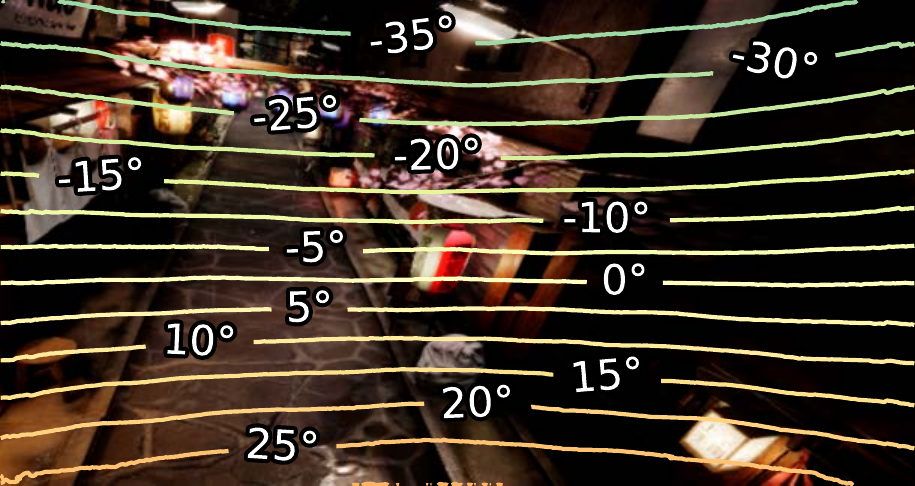} \\
        polar angles $\lVert\boldsymbol{\theta}\rVert$ & & 
        \includegraphics[height=\bfigh,valign=m]{\lbfnamebase/theta_norm_gt.pdf} & 
        \includegraphics[height=\bfigh,valign=m]{\lbfnamebase/theta_norm.pdf} & 
        \includegraphics[height=\bfigh,valign=m]{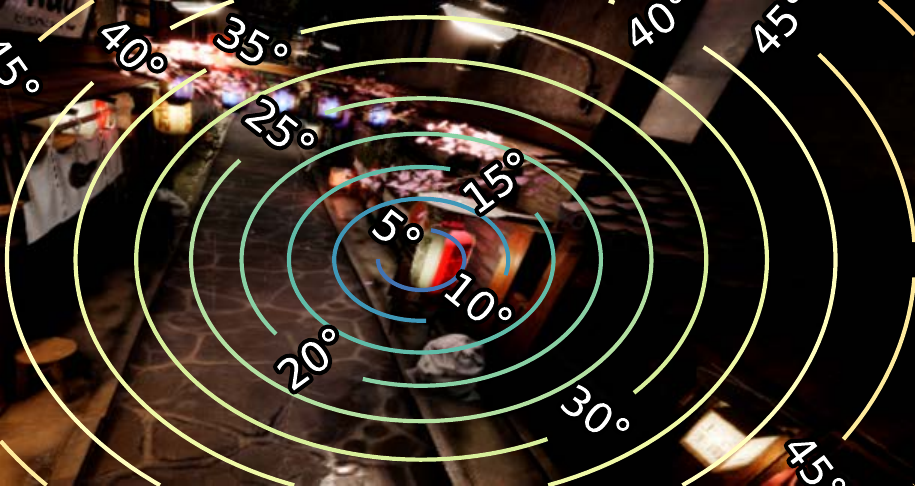} & 
        \includegraphics[height=\bfigh,valign=m]{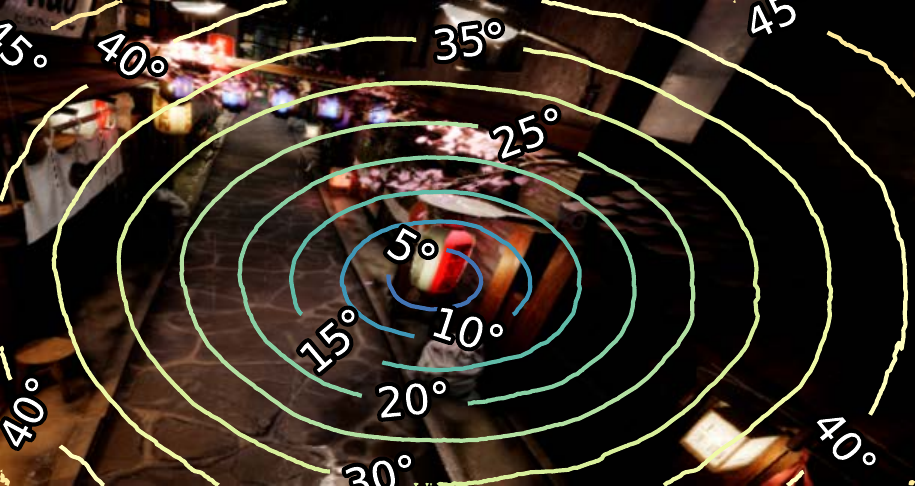} \\
        fit & \code{pinhole} & 
        \includegraphics[height=\bfigh,valign=m]{\lbfnamebase/theta_norm_cam_gt.pdf} & 
        \includegraphics[height=\bfigh,valign=m]{\lbfnamebase/theta_norm_cam_pinhole.pdf} &
        \includegraphics[height=\bfigh,valign=m]{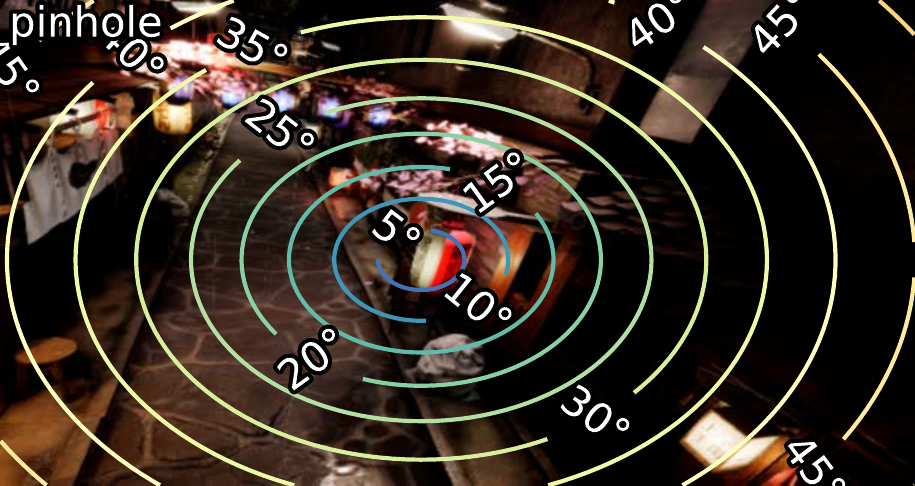} & 
        \includegraphics[height=\bfigh,valign=m]{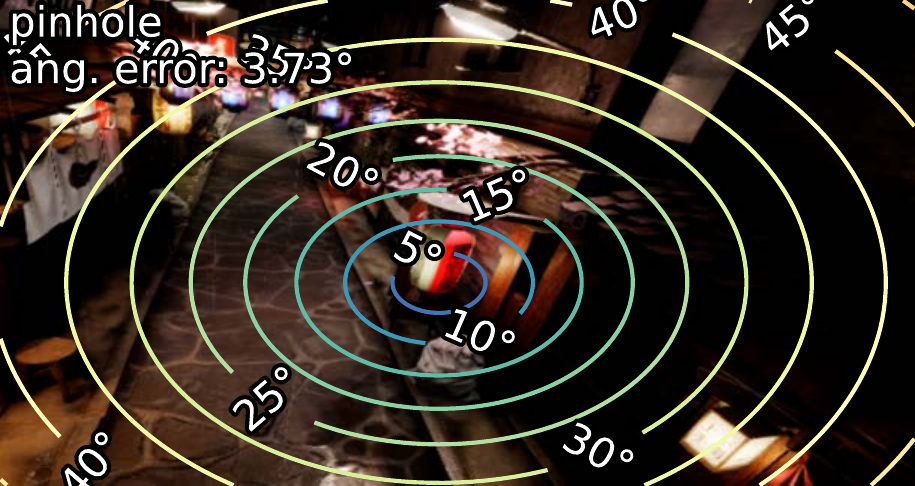} \\
    \end{tblr}
    \caption{\textbf{Qualitative results on edited images} with AnyCalib being trained following \cite{he2025diffcalib,zhu2023wildcam} (\cref{sec:imp}).}
    \label{fig:supp_qual_edit}
\end{figure*}

\begin{figure*}[t]

    \newlength{\figw}
    \setlength{\figw}{4.2cm}

    \centering
    \begin{tblr}{
        width=1.0\linewidth,
        vspan=even,
        colspec={*{4}c},
        colsep=1pt,
        rowsep=1pt,
    }
        original & undistorted & original & undistorted \\
        \includegraphics[width=\figw,valign=m]{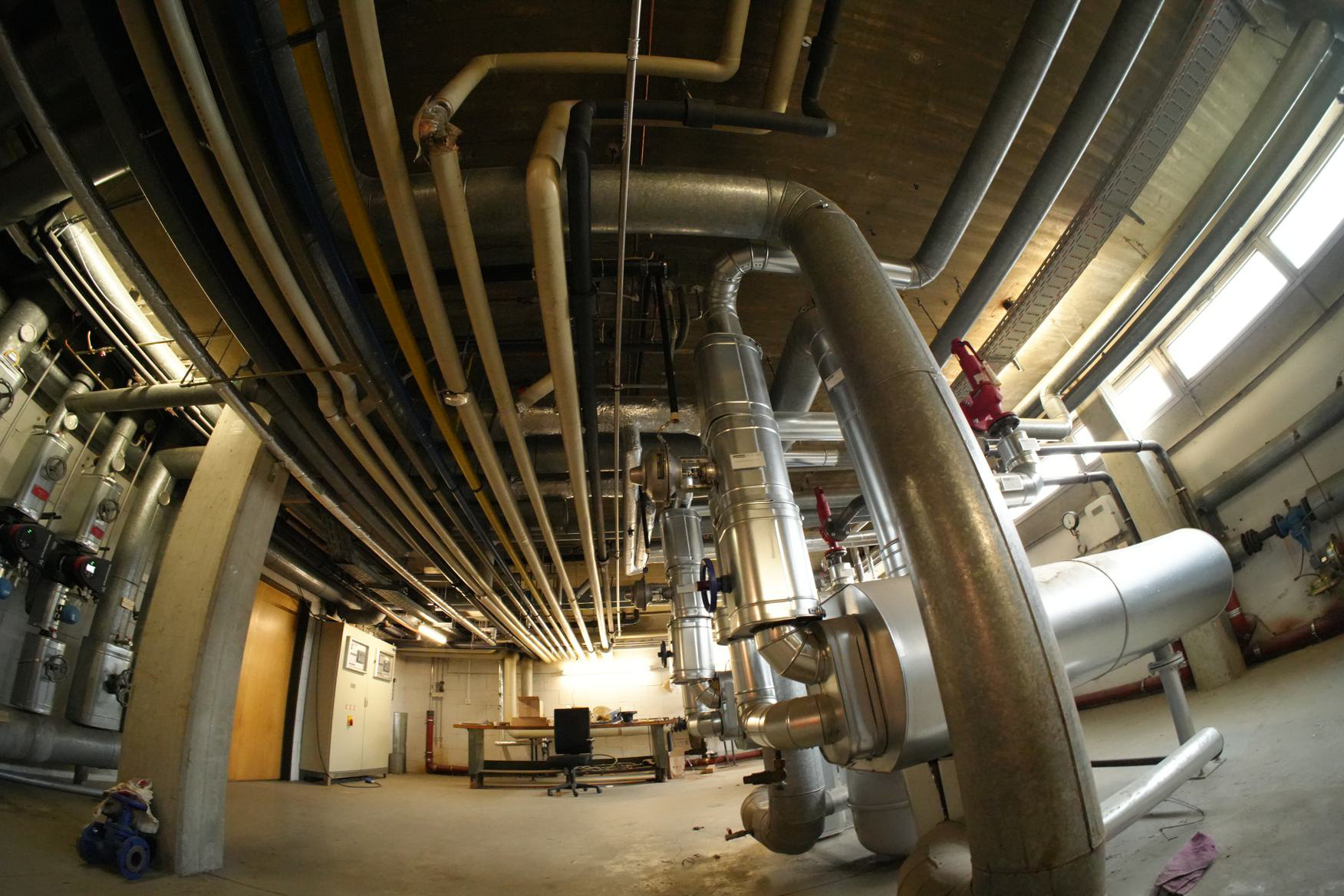} & 
        \includegraphics[width=\figw,valign=m]{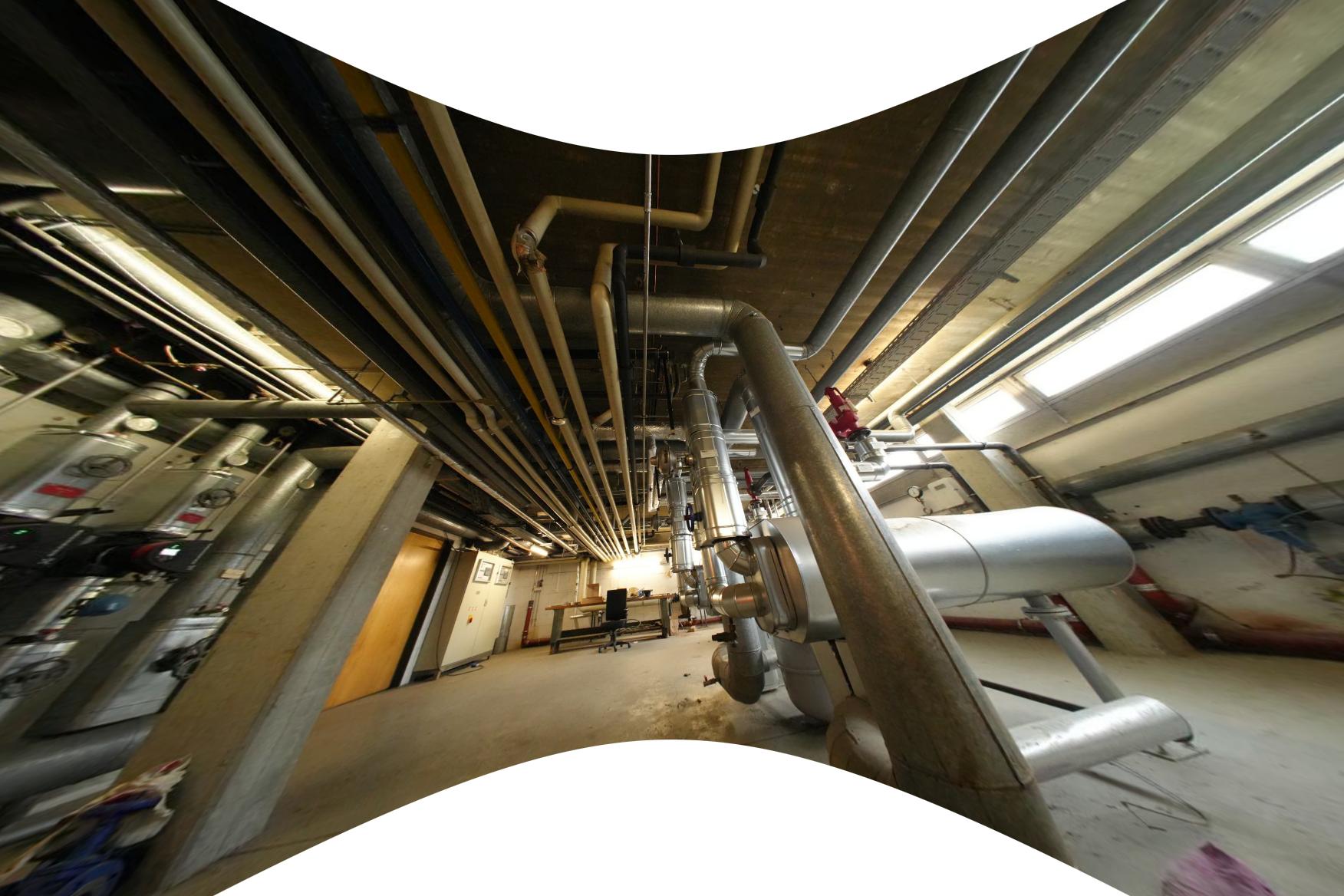} & 
        \includegraphics[width=\figw,valign=m]{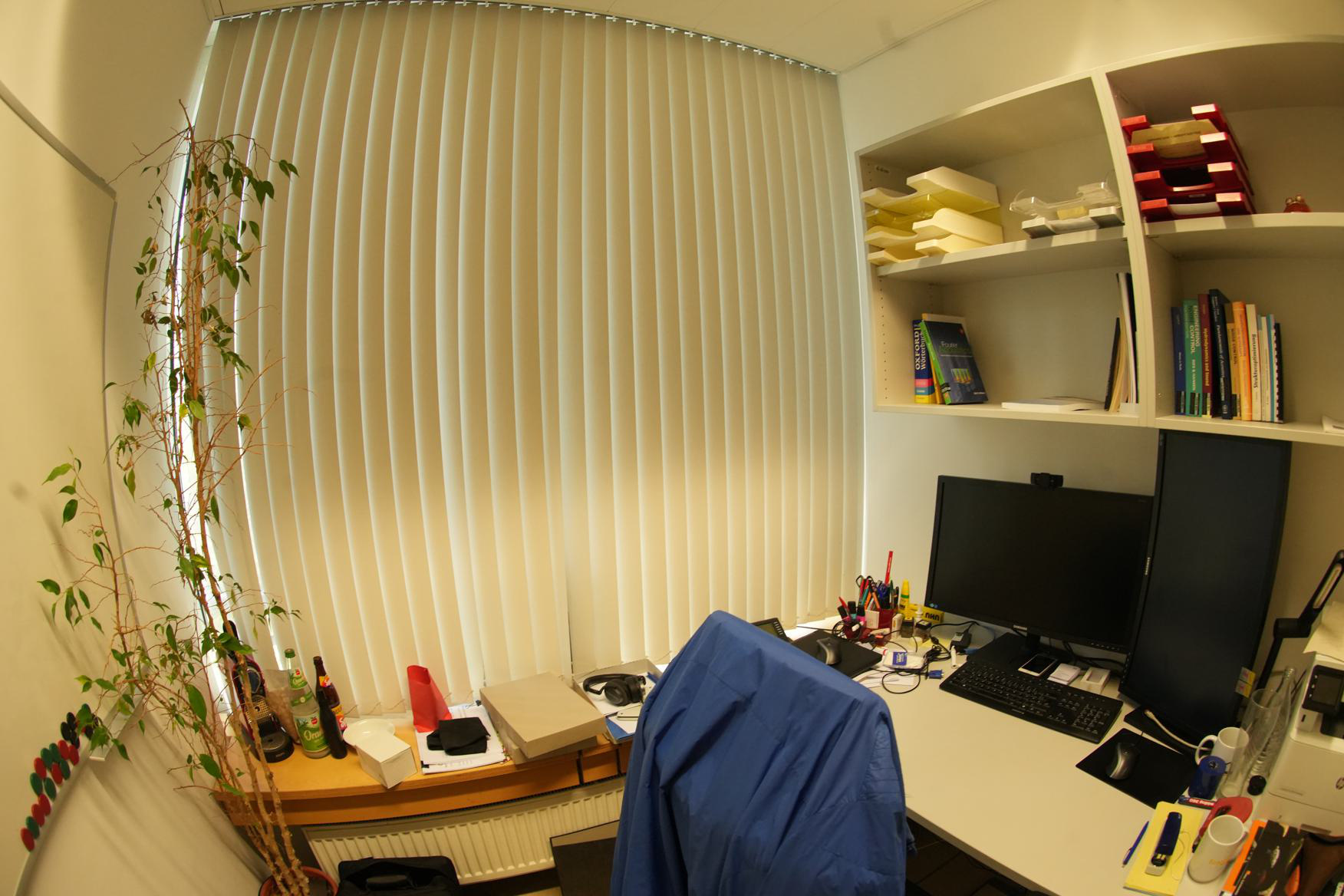} & 
        \includegraphics[width=\figw,valign=m]{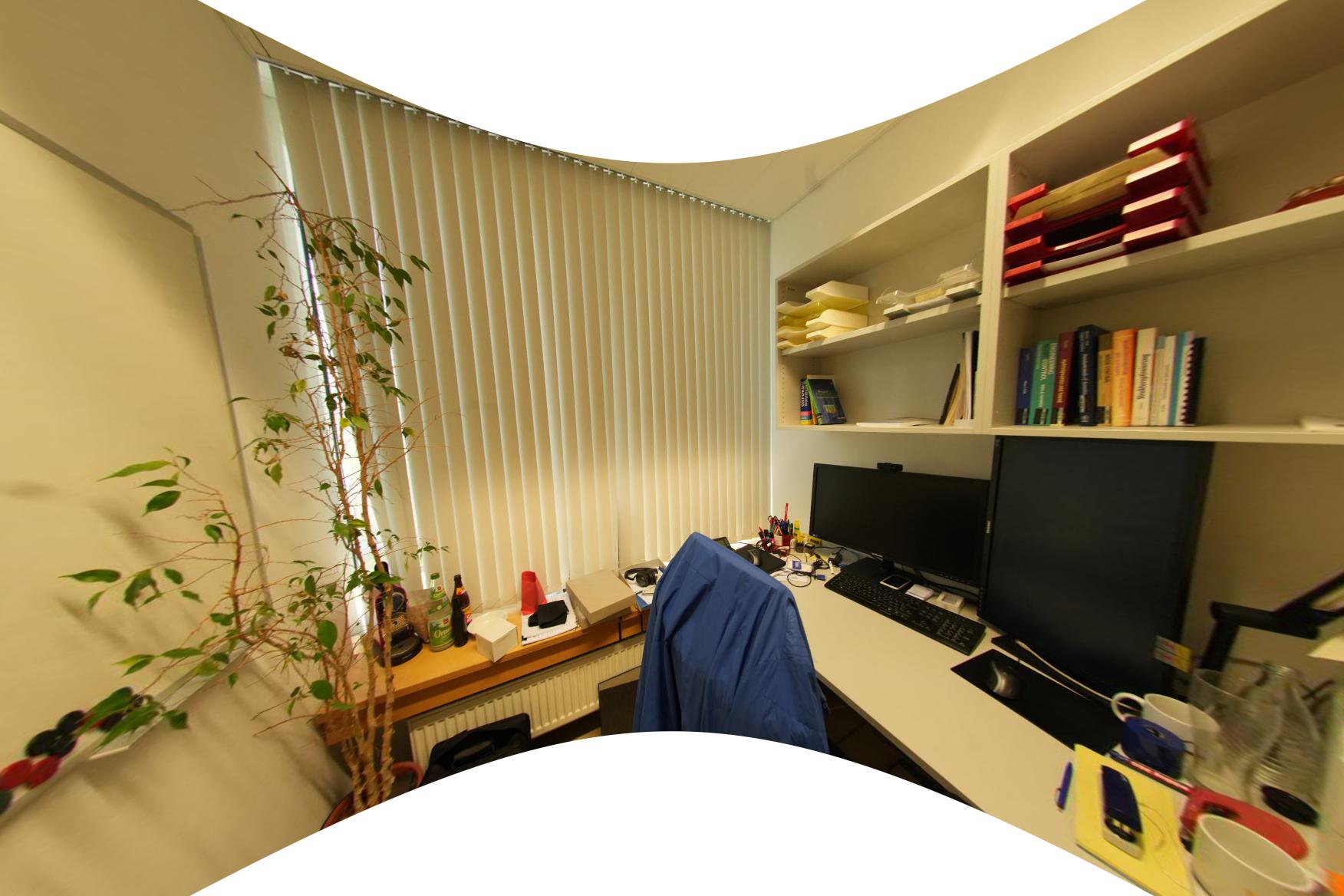} \\
        \includegraphics[width=\figw,valign=m]{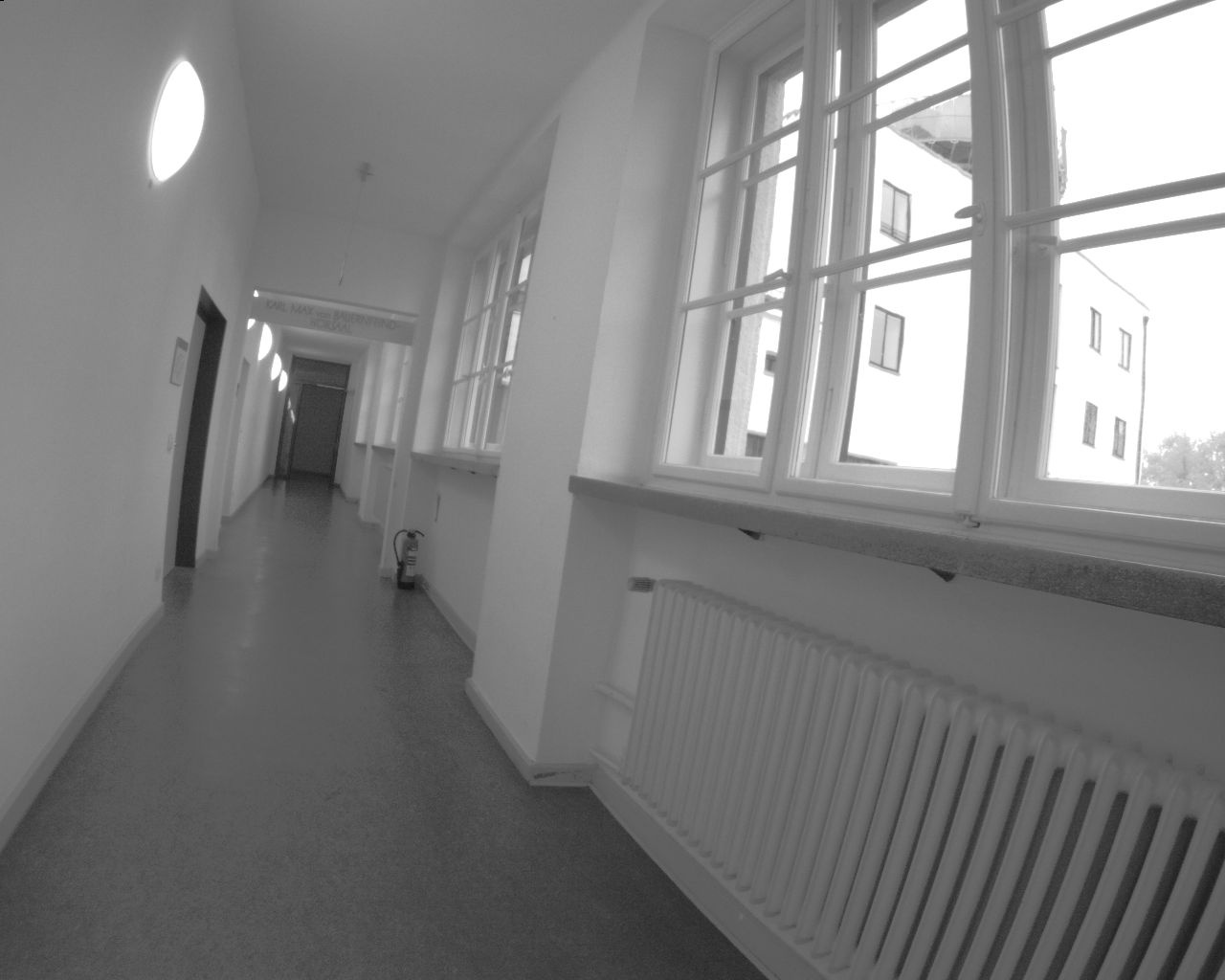} & 
        \includegraphics[width=\figw,valign=m]{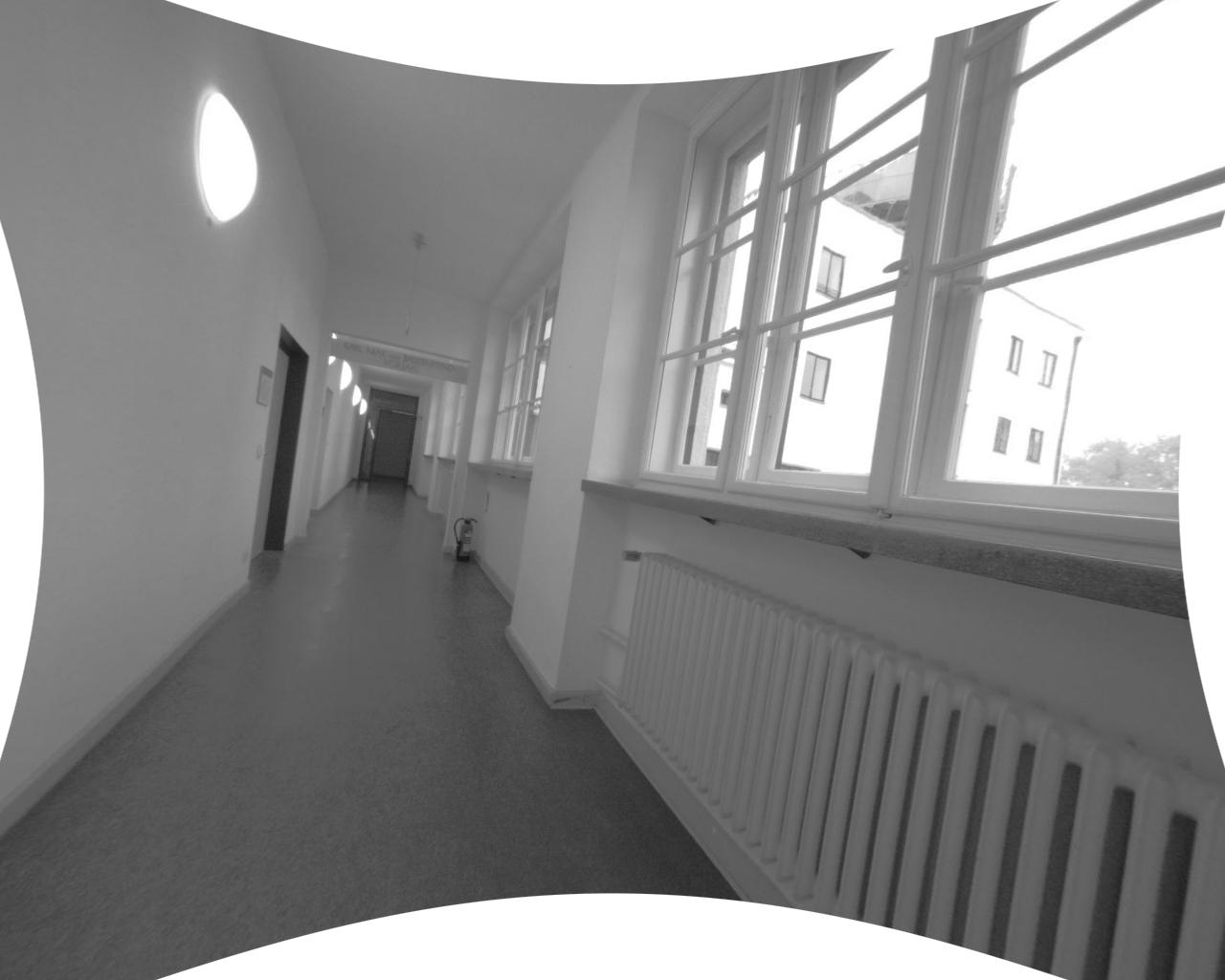} &
        \includegraphics[width=\figw,valign=m]{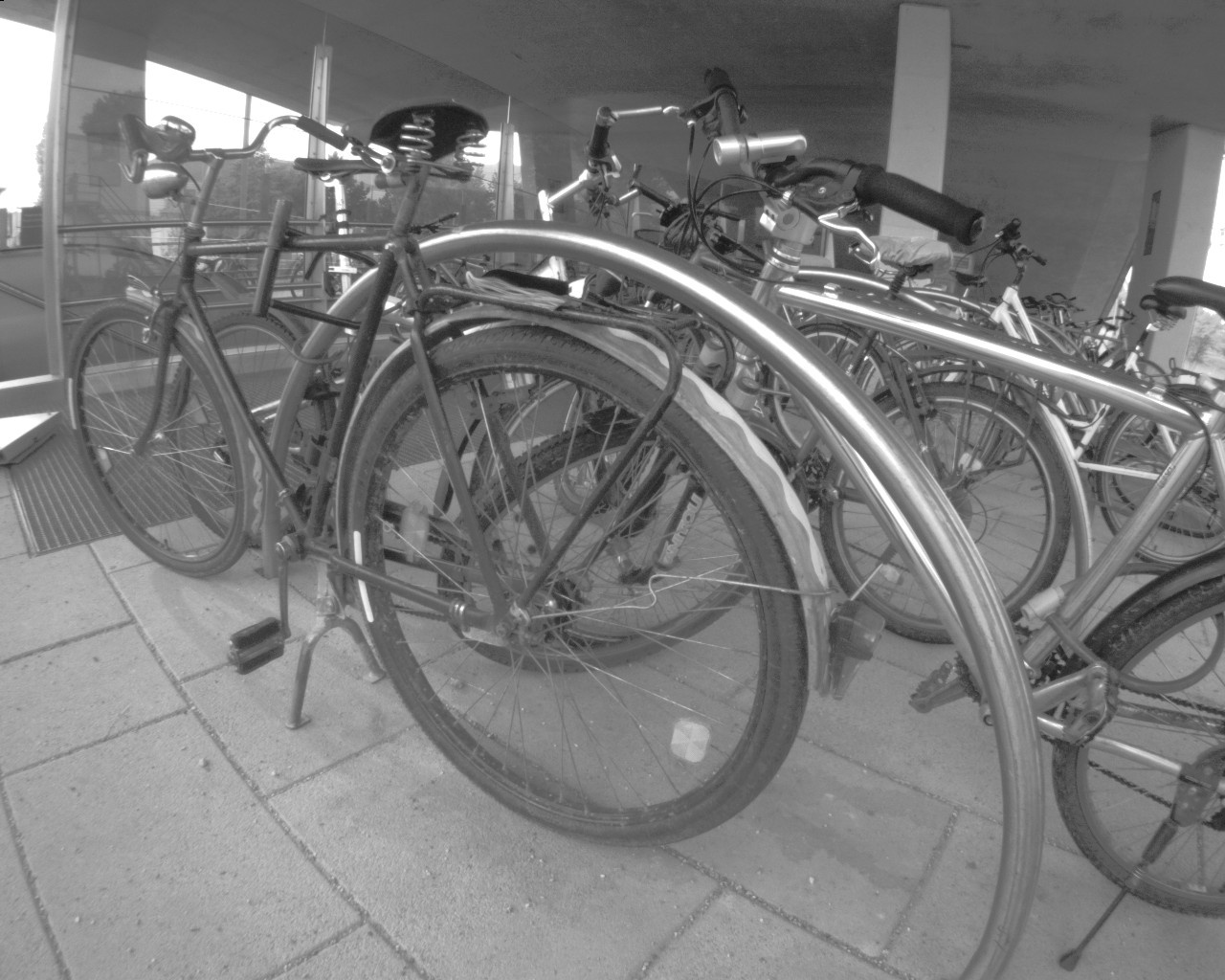} & 
        \includegraphics[width=\figw,valign=m]{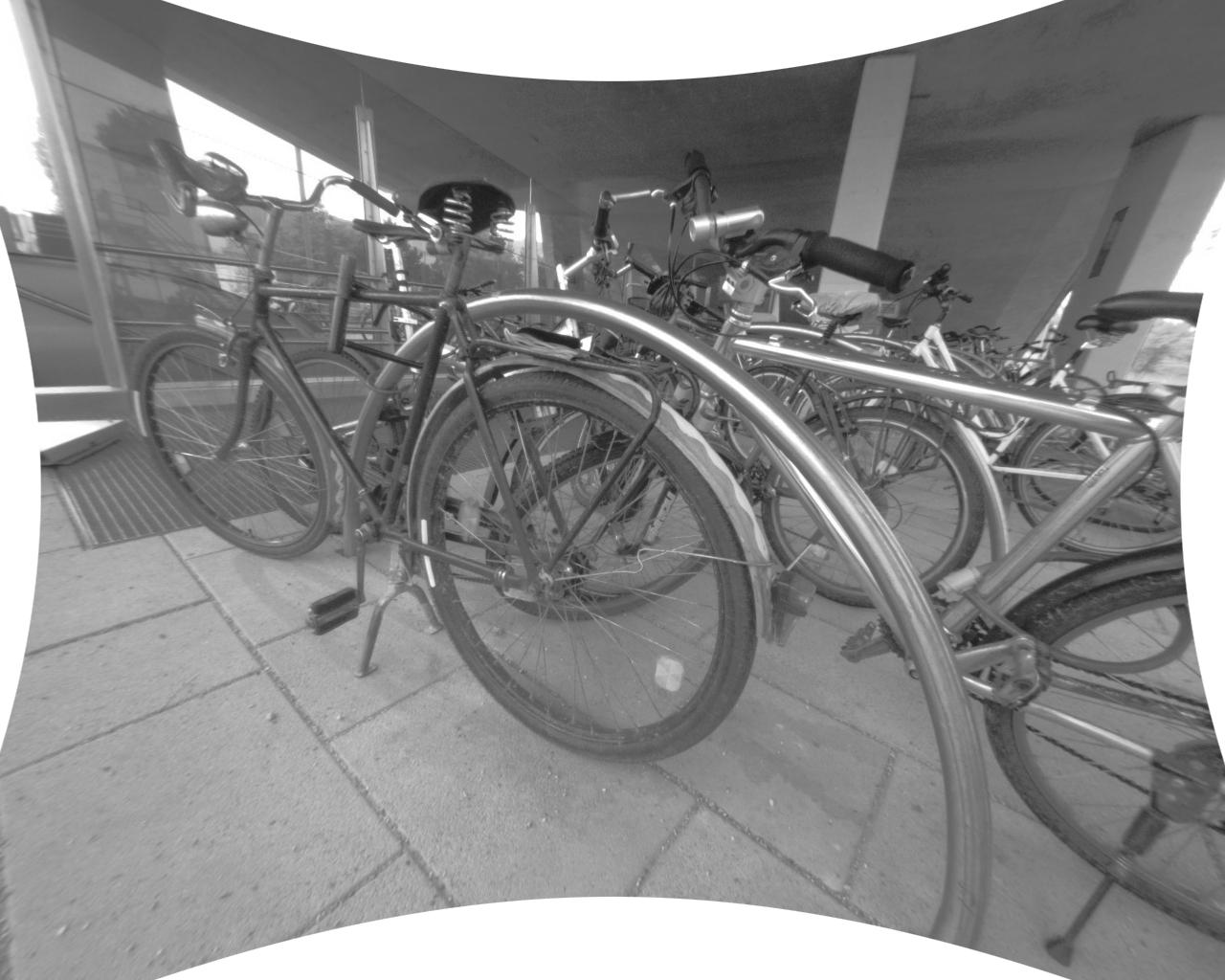} \\
        \includegraphics[width=\figw,valign=m]{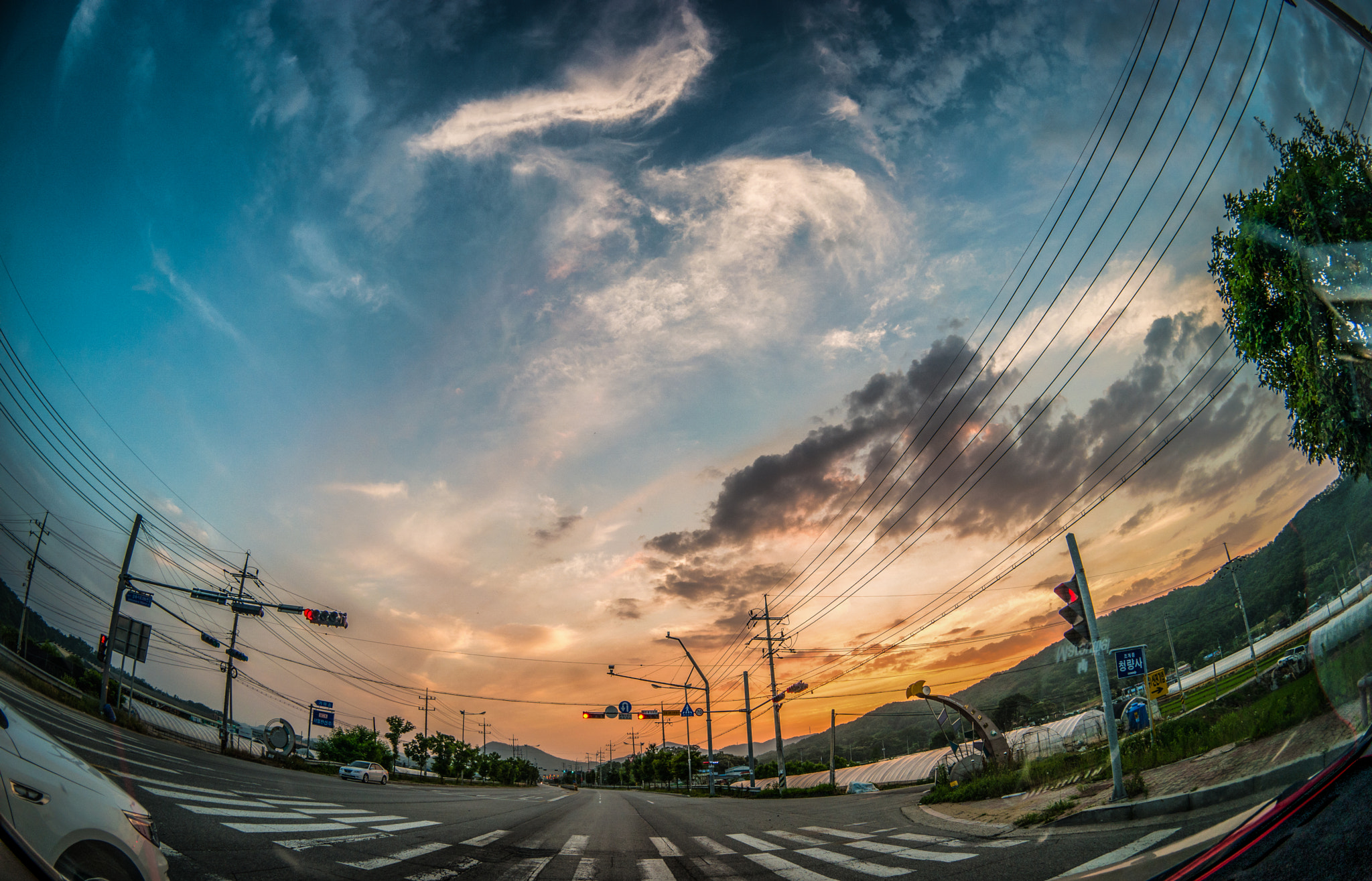} & 
        \includegraphics[width=\figw,valign=m]{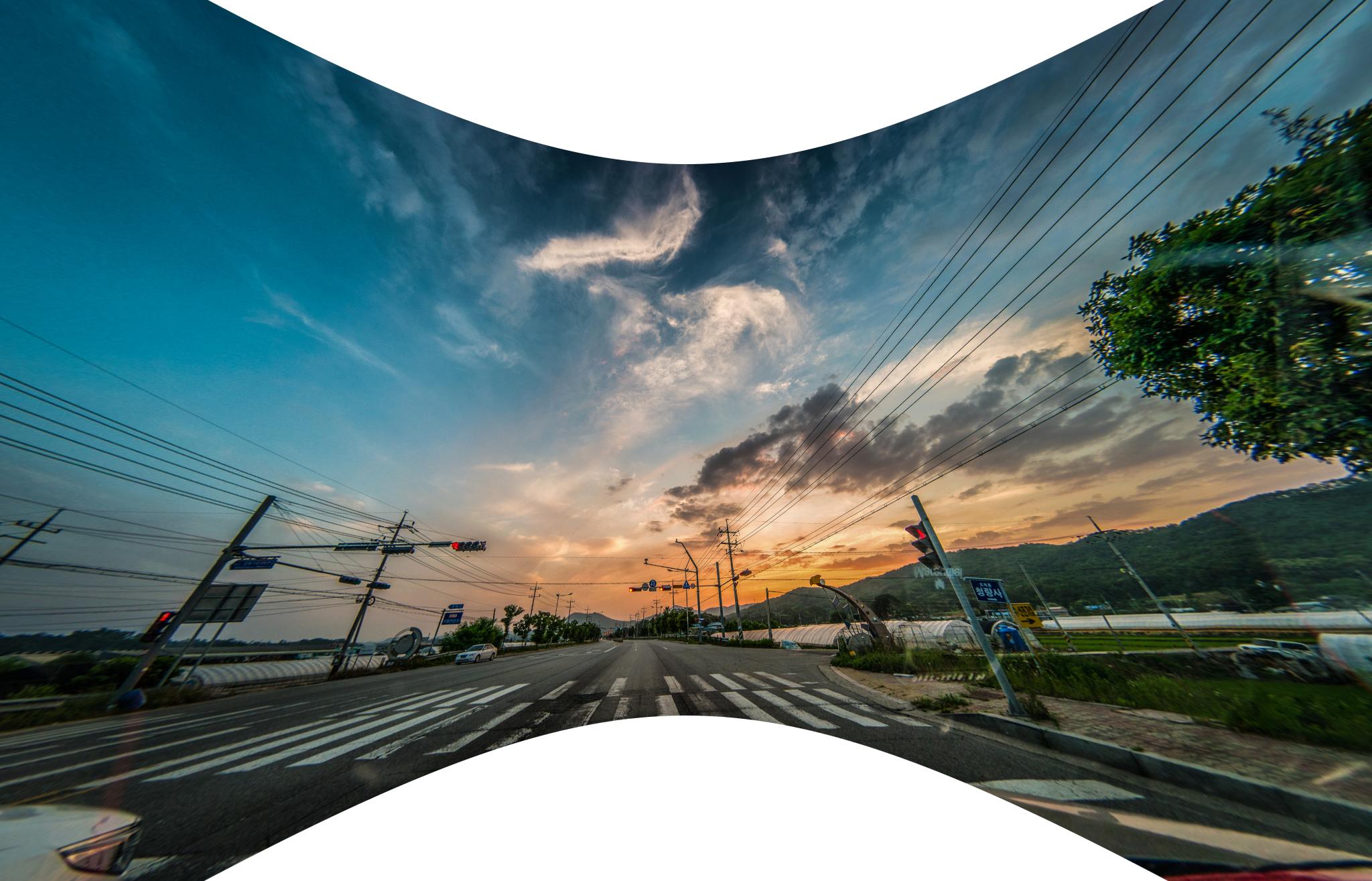} & 
        \includegraphics[width=\figw,valign=m]{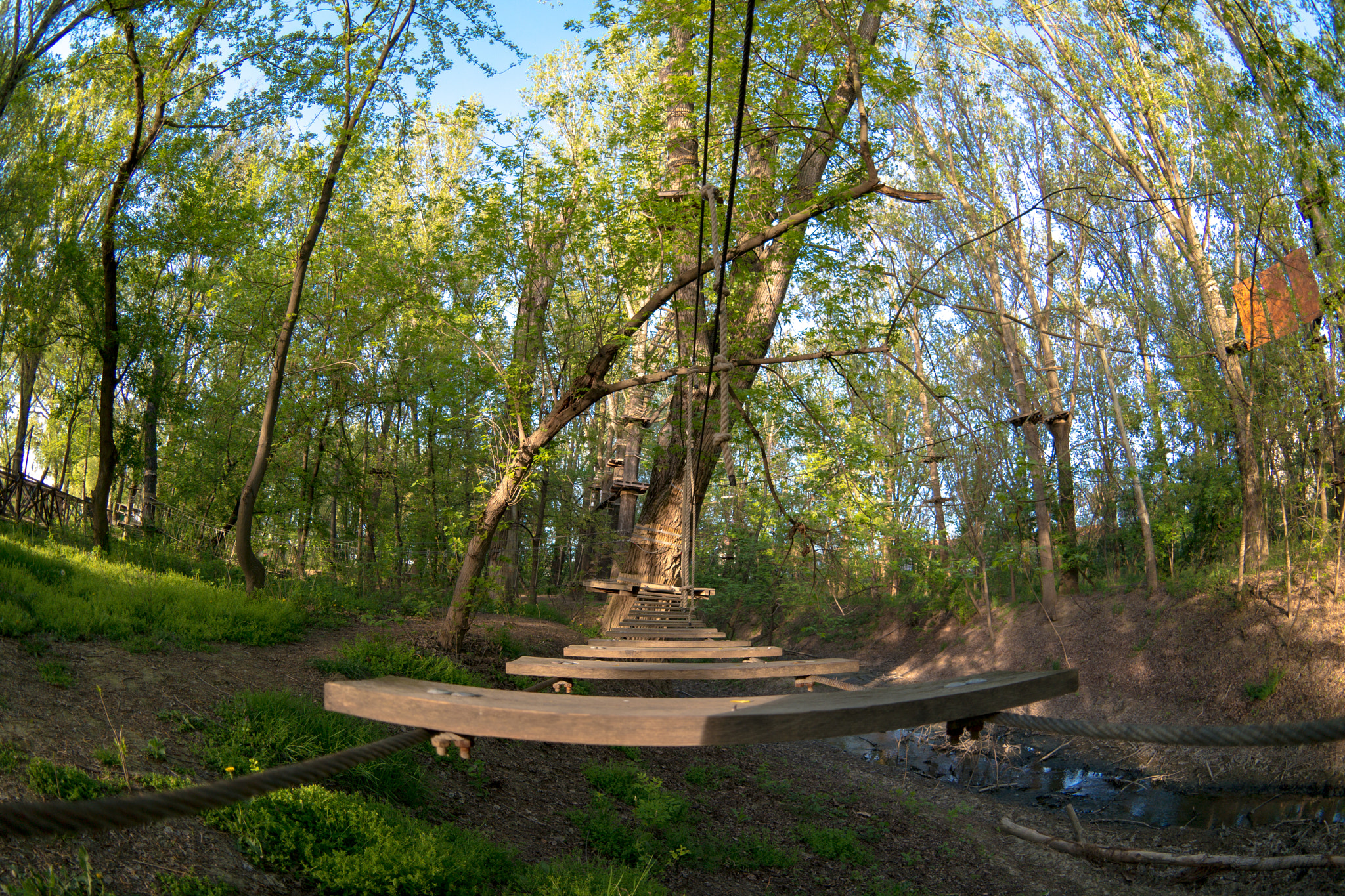} & 
        \includegraphics[width=\figw,valign=m]{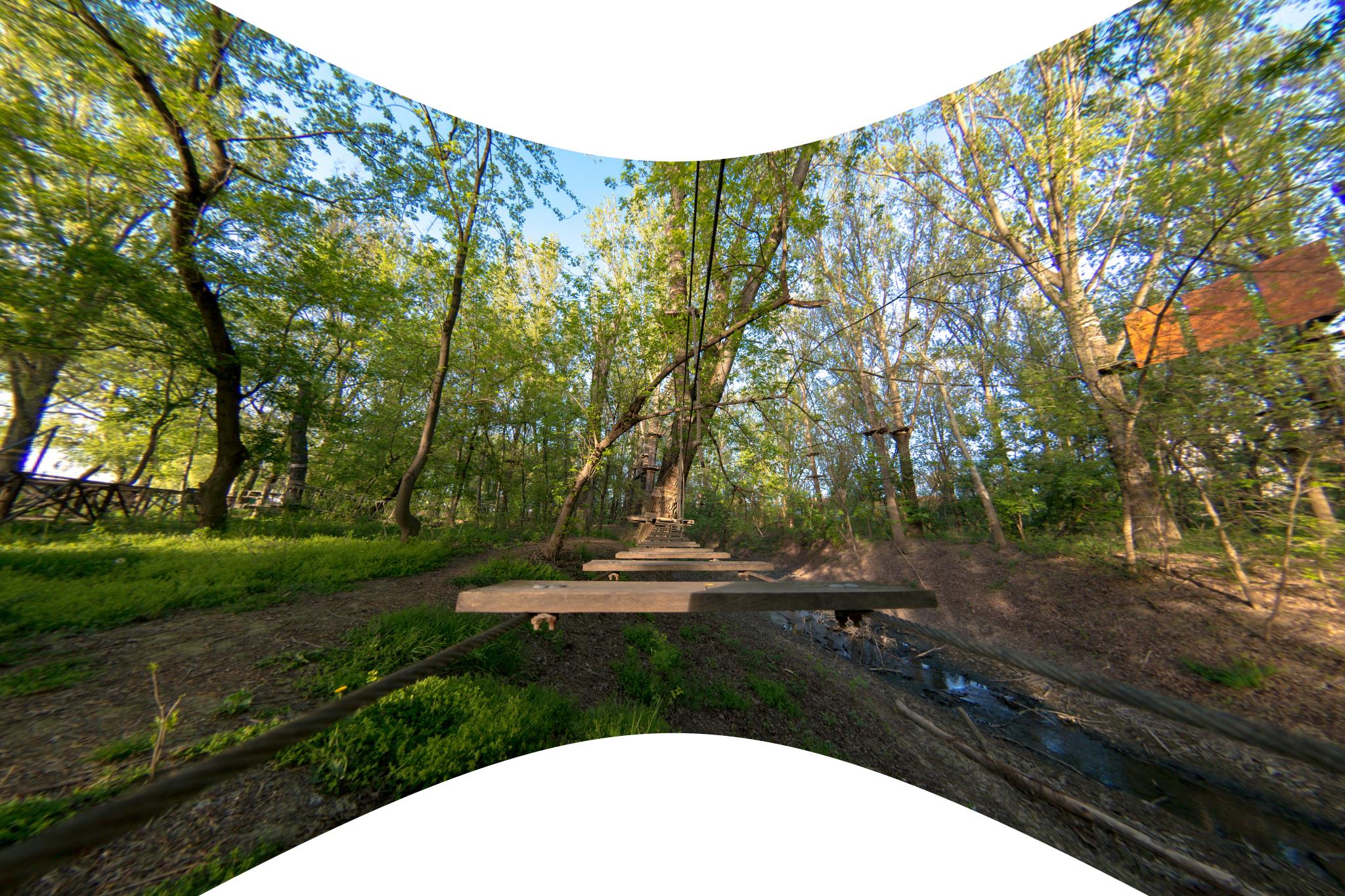} \\
    \end{tblr}
    \caption{\textbf{Qualitative undistortion results} with AnyCalib$_\mathrm{gen}$ (trained on \datg), on images from ScanNet++ \cite{Yeshwanth2023scannetpp} (top), Mono \cite{engel2016monodataset} (middle) and captured with a Samsung NX 10mm F3.5 Fisheye lens (bottom), provided by \href{https://explorecams.com/photos/lens/samsung-nx-10mm-f3-5-fisheye}{ExploreCams}---authors: \href{https://explorecams.com/photos/epidmHVSwD?lens=samsung-nx-10mm-f3-5-fisheye}{crystal Yang} (left) and \href{https://explorecams.com/photos/dYXHUevDXw?lens=samsung-nx-10mm-f3-5-fisheye}{Imre Farago} (right).}
    \label{fig:supp_qual_undist}
\end{figure*}

We show qualitative results, using AnyCalib$_{\mathrm{gen}}$ (trained on \datg) on perspective images in \cref{fig:supp_qual_tartan,fig:supp_qual_stan,fig:supp_qual_lamar,fig:supp_qual_mega} and on distorted images in \cref{fig:supp_qual_scan,fig:supp_qual_monovo}. We also show undistortion results using the same model in \cref{fig:supp_qual_undist}. Additional qualitative results on edited (stretched and cropped) images are shown in \cref{fig:supp_qual_edit}, with AnyCalib being trained following \cite{he2025diffcalib,zhu2023wildcam} (\cref{sec:imp}).

\begin{figure*}[t]

    \setlength{\figh}{3cm}

    \definecolor{codecolor}{RGB}{246, 248, 250}
    \newcommand{\code}[1]{\colorbox{codecolor}{\texttt{#1}}}

    \newcommand{\lfnamebase}{figures/supp_qualitative_perspective/area_1-camera_ffd2cca53bfe4dc580469bc1fffb077e_office_22_frame_equirectangular_domain_rgb_r32_p2_y-168_f53}
    \newcommand{\rfnamebase}{figures/supp_qualitative_perspective/area_3-camera_4a7bfe0577f74a1a891683cf5b435f93_lounge_1_frame_equirectangular_domain_rgb_r-16_p2_y56_f37}

    \centering
    \begin{tblr}{
        width=1.0\linewidth,
        vspan=even,
        colspec={*{6}c},
        colsep=1pt,
        rowsep=1pt,
        cell{2}{1}={r=2}{c, cmd=\rotatebox{90}},
        cell{4}{1}={c, cmd={\rotatebox[origin=c]{90}}},
        cell{5,6,7}{2}={c, cmd={\rotatebox[origin=c]{90}}},
        cell{5}{1}={r=3}{c, cmd=\rotatebox{90}},
    }
        && ground-truth & prediction & ground-truth & prediction \\
        FoV field & $\theta_x$ & 
        \includegraphics[height=\figh,valign=m]{\lfnamebase/theta_x_gt.pdf} & 
        \includegraphics[height=\figh,valign=m]{\lfnamebase/theta_x.pdf} & 
        \includegraphics[height=\figh,valign=m]{\rfnamebase/theta_x_gt.pdf} & 
        \includegraphics[height=\figh,valign=m]{\rfnamebase/theta_x.pdf} \\
        -         & $\theta_y$ & 
        \includegraphics[height=\figh,valign=m]{\lfnamebase/theta_y_gt.pdf} & 
        \includegraphics[height=\figh,valign=m]{\lfnamebase/theta_y.pdf} &
        \includegraphics[height=\figh,valign=m]{\rfnamebase/theta_y_gt.pdf} & 
        \includegraphics[height=\figh,valign=m]{\rfnamebase/theta_y.pdf} \\
        polar angles $\lVert\boldsymbol{\theta}\rVert$ & & 
        \includegraphics[height=\figh,valign=m]{\lfnamebase/theta_norm_gt.pdf} & 
        \includegraphics[height=\figh,valign=m]{\lfnamebase/theta_norm.pdf} & 
        \includegraphics[height=\figh,valign=m]{\rfnamebase/theta_norm_gt.pdf} & 
        \includegraphics[height=\figh,valign=m]{\rfnamebase/theta_norm.pdf} \\
        fits & \code{pinhole} & 
        \includegraphics[height=\figh,valign=m]{\lfnamebase/theta_norm_cam_gt.pdf} & 
        \includegraphics[height=\figh,valign=m]{\lfnamebase/theta_norm_cam_pinhole.pdf} &
        \includegraphics[height=\figh,valign=m]{\rfnamebase/theta_norm_cam_gt.pdf} & 
        \includegraphics[height=\figh,valign=m]{\rfnamebase/theta_norm_cam_pinhole.pdf} \\
        - & \code{division:1} && 
        \includegraphics[height=\figh,valign=m]{\lfnamebase/theta_norm_cam_division_1.pdf} && 
        \includegraphics[height=\figh,valign=m]{\rfnamebase/theta_norm_cam_division_1.pdf} \\
        - & \code{division:2} && 
        \includegraphics[height=\figh,valign=m]{\lfnamebase/theta_norm_cam_division_2.pdf} &&
        \includegraphics[height=\figh,valign=m]{\rfnamebase/theta_norm_cam_division_2.pdf} \\
    \end{tblr}
    \caption{\textbf{Qualitative results on perspective images} in Stanford2D3D \cite{armeni2017stanford2d3d} with AnyCalib$_{\mathrm{gen}}$---trained on \datg. The FoV field ($\theta_x$ and $\theta_y$) is regressed by the network and $\lVert\boldsymbol\theta\rVert$ represents both its norm and the polar angle of the ray corresponding to each pixel. The predicted FoV field is used to fit the camera model of choice. \code{division:i} corresponds to the division model with \code{i} distortion coefficients.}
    \label{fig:supp_qual_stan}
\end{figure*}

\newpage

\end{document}